\pdfoutput=1
\documentclass[3p,preprint,number]{elsarticle}
\DeclareGraphicsExtensions{.pdf,.gif,.jpg,.pgf}

\usepackage[utf8]{inputenc}
\usepackage[english]{babel}
\usepackage{ragged2e}
\usepackage{blindtext}
\usepackage{subcaption}
\usepackage{mathtools}
\usepackage{lipsum}
\usepackage{array}
\usepackage{colortbl}
\usepackage{chemfig}
\usepackage{tikz}
\tikzstyle{startstop} = [ellipse, draw, fill=red!30, text width=5em, text centered, minimum height=2em]
\tikzstyle{process} = [rectangle, draw, fill=blue!30, text width=7em, text centered, rounded corners, minimum height=3em]
\tikzstyle{arrow} = [thick,->,>=stealth]
\usetikzlibrary{patterns} 
\usetikzlibrary{shapes.geometric}
\usepackage{pgfplots}
\usepackage{amsmath}
\usepackage{physics}
\pgfplotsset{compat=1.18}
\usepackage{amssymb}
\usepackage{subcaption}
\usepackage{siunitx}
\usepackage{enumitem}
\usepackage{xcolor}
\usepackage{multirow}
\usepackage{adjustbox}
\usepackage{geometry}
\usepackage{tabularx}
\usepackage{csquotes}
\usetikzlibrary{patterns}
\usepackage[english]{babel}

\newtheorem{definition}{Definition}[section]
\makeatletter
\DeclareMathOperator*{\argmax}{arg\,max}
\DeclareMathOperator*{\argmin}{arg\,min}
\newcommand*{\rom}[1]{\expandafter\@slowromancap\romannumeral #1@}
\makeatother
\usepackage{tikz}
\usetikzlibrary{matrix}
\usepackage{booktabs}
\graphicspath{{Figures/}}

\usepackage{blindtext}
\usepackage{titlesec}
\titleformat{\chapter}[display]   
{\normalfont\huge\bfseries}{\chaptertitlename\ \thechapter}{16pt}{\Huge}   
\titlespacing*{\chapter}{0pt}{-10pt}{5pt}
\usepackage{algorithmic}
\usepackage[linesnumbered,ruled,vlined]{algorithm2e}
\usepackage{layouts}

\usepackage{setspace}

\begin{document}
	\bibliographystyle{elsarticle-num.bst}
	{\setlength{\parindent}{0cm} 
		\begin{frontmatter}
			\title{Optimising Urban Flood Resilience}
			\author{James Mckenna}
			\ead{James.Mckenna@newcastle.ac.uk}
			\author{Christos Iliadis}
			\ead{Christos.Iliadis@newcastle.ac.uk}
			\author{Vassilis Glenis\corref{cor1}}
			\ead{Vassilis.Glenis@newcastle.ac.uk}
			
			\cortext[cor1]{Corresponding author}
			\address{School of Engineering, Newcastle University, Newcastle upon Tyne, United Kingdom}
			\date{March 2025}
			\begin{abstract}			
				Due to the increasing frequency and severity of storm events, driven by the escalation of anthropogenic climate change and urban expansion, there is a requirement for increasingly efficient flood risk management strategies. While Blue-Green Infrastructure (BGI) offers a sustainable solution for managing flood risk, optimal implementation is challenging. To help overcome this challenge, this study presents a novel multi-objective optimisation tool that couples a state-of-the-art hydrodynamic model with a bespoke evolutionary algorithm.
				
				The use of a fully dynamic hydrodynamic model enables the tool to accurately evaluate the effectiveness of proposed BGI features with respect to property scale flood vulnerability and hazard analysis. This contrasts with alternative approaches which utilise simplified models, which can only reliably predict inundation extents, thus the proposed optimisation tool provides greater certainty regarding the optimality of the solutions. As a hydrodynamic simulation is required to evaluate each candidate solution, the bespoke evolutionary algorithm is specifically designed to minimise the number of simulations required, ensuring the tool is computationally practical. The effectiveness of the tool in this regard is validated via the derivation of exact convergence measures, for a tractable search space, and via comparisons with benchmark algorithms, for an intractable search space.
				
				Compared with traditional design practices, the proposed tool offers an automated approach capable of efficiently exploring a wide range of solutions, providing decision-makers with a set of optimal solutions from which they can make informed investment decisions. The presented methods provide a robust framework for optimising a variety of BGI features in complex urban environments.
			\end{abstract}
			\begin{keyword}
				Multi-objective optimisation, evolutionary algorithms, blue-green infrastructure, flood exposure, CityCAT.
			\end{keyword}
		\end{frontmatter}
		\section{Introduction}
		Storm events are a significant concern in hydrological risk analysis due to the potentially devastating impacts induced by flooding. This necessitates the estimation of the impact of extreme rainfall for specific time frames and selected return periods in order to adequately understand and mitigate the associated risks \cite{Vasiliades2024}. The emergence of anthropogenic climate change has brought more frequent and intense storm events, escalating the vulnerability and exposure of urban environments to flooding. Consequently, there is an acknowledged necessity for a paradigm shift in urban water management strategies to address the increasing frequency and intensity of storm events, prolonged dry seasons, and the effects of urban expansion, which reduce permeable spaces and further disrupt the natural hydrological cycle in dense urban areas \cite{Mignot2022, O'Donnell2020}. Due to the increasing frequency and intensity of extreme storm events, surface water flooding poses a critical hazard in urban environments, with its prevalence projected to rise globally. Consequently, this issue has become a critical focus in flood risk analysis.
			
		The design and implementation of effective adaptation strategies is essential to mitigate, and in some instances, manage the flow and volume of floodwaters within the urban landscape \cite{Galiatsatou2022}. One of the most effective tools which aids this process is the use of hydrodynamic models to simulate and predict flood behaviour in complex urban settings \cite{Rosenzweig2021}. These models enable the simulation of both surface and subsurface water dynamics, helping to predict the extent and behaviour of flood flows under different rainfall scenarios \cite{Glenis2018, Guo2021, Sanders2017, Teng2017, TUFLOW2018}. By combining geographic information system data, rainfall data, and hydrological inputs, these models allow practitioners to identify flood-prone areas and assess the impact of different intervention strategies \cite{Alves2016, Morgan2019, Singh2021}. Hydrodynamic models therefore not only provide a detailed understanding of the movement of flood flows but also allow for the analysis and optimisation of various flood management interventions \cite{Ferrans2023, Xie2024}.
		
		State-of-the-art hydrodynamic models typically employ a dual drainage model, in which the simulation of surface water is handled by a two-dimensional (2D) hydrodynamic model and the sub-surface domain is modelled by a one-dimensional (1D) hydrodynamic model \cite{Djordjevic1999, Guo2021, Mignot2022, Rosenzweig2021}. 1D overland flow modelling remains popular despite the emergence of 2D hydrodynamic models due to the reduced computational burden and data requirements \cite{Costabile2015}. This is especially true within industry practice, where there exists vast institutional knowledge and experience with 1D models. However, where flows are two-dimensional in nature, as is expected in urban environments, the accuracy of 1D models is restricted by their inability to capture transverse variations in the flow regime \cite{Costabile2012, Petaccia2012}.
			
		Simplified overland flow models, such as the simple inertia formulation \cite{Bates2010}, diffusive-wave formulation \cite{Costabile2017} or the kinematic-wave formulation \cite{Miller1984}, have also gained popularity as a reduced complexity alternative, suitable for modelling flood inundation extents only where the underlying assumptions are satisfied \cite{Costabile2017, Costabile2020}. However, where locally accurate hydrodynamic modelling of urban areas is required, as is the case for performing detailed analyses of the effectiveness of proposed intervention strategies, fully dynamic models solving the classical shallow water equations represent the minimum complexity of model required \cite{Costabile2020}. Locally accurate hydrodynamic modelling is required, since a high modelling resolution is necessary to adequately resolve complex urban flow dynamics such that property scale flood vulnerability and hazard analyses can be performed \cite{Dewals2023, Liu2018, Mignot2020, Bazin2017, Paquier2015, Bazin2012}. For a rigorous mathematical analysis of the limitations of the simple inertia formulation, where the advection term is neglected from the momentum conservation equation, see Cozzolino et al. \cite{Cozzolino2019}.
			
		Following the shift towards nature-based solutions in the late 20$^{\mathrm{th}}$ century, emphasised by policy implementation such as the European Union’s Water Framework Directive \cite{EU2000}, the concept of blue-green infrastructure (BGI) has emerged as a key aspect of modern flood risk management best practice. The concept of BGI emphasises water retention, infiltration and controlled conveyance via natural (green) and engineered (blue) solutions to mitigate flood risk whilst providing ecological and social benefits. Examples of BGI features include sustainable urban drainage systems (SuDS), such as green and blue roofs, permeable pavements, swales and detention/retention basins \cite{WoodsBallard2015}. In order to understand the effectiveness of urban flood risk management strategies, it is therefore crucial for practitioners to be able to explicitly model BGI features within high resolution hydrodynamic models.
		
		The increasing complexity of urban flood management has necessitated the adoption of more sophisticated methods of decision-making and design. Traditional approaches often rely on trial-and-error methods and/or engineering expertise to identify optimal flood mitigation strategies. However, this process may be time-consuming, expensive and prone to human error. Machine learning algorithms present an alternative automated approach capable of overcoming these limitations. Intelligent automated design processes promise practitioners the opportunity to more efficiently explore a broader range of solutions and achieve better outcomes. Recent research on this topic has primarily centred around the use of evolutionary algorithms, due to their suitability for solving such problems.
		
		For example, Luo et al. \cite{Luo2022} employed a popular evolutionary algorithm, the Non-Dominated Sorting Genetic Algorithm II (NSGA-II) \cite{Deb2000}, to optimise the design of Sponge City projects using the Storm Water Management Model (SWMM) \cite{Metcalf1971} as the hydrodynamic model. Similar optimisation strategies, using the same evolutionary algorithm and hydrodynamic model, have been extensively explored within the literature (see \cite{Wan2022, Li2024, Zhu2023, Yang2023, Zhou2024, DuarteLopes2021, Hassani2023, Hassani2024, MousaviJanbehsarayi2023, Tansar2023}). Likewise, alternative optimisation algorithms have also been coupled with SWMM \cite{Li2022, Yu2022, Wu2025, Taghizadeh2021, Marco2019, Shojaeizadeh2021, Talebi2024, Chen2024, Eckart2018}. It is clear therefore that, for the optimisation of the placement of BGI, SWMM is the predominant model for which optimisation tools are being developed. This is likely due to SWMM’s open-source nature and its relative computational simplicity.
			
		However, although the latest version of SWMM has three options for overland flow routing, including a steady-state approximation, kinematic wave approximation and a dynamic wave routing algorithm \cite{Singh2010, EPA2017}, none of the these options represent a suitable 2D overland flow model for the purpose of determining property scale vulnerability and hazard analysis. Furthermore, SWMM is not a spatially distributed model, meaning surface water interactions are represented in a highly simplified manner using storage nodes and predefined diversion structures \cite{Singh2010, EPA2017}. It is therefore not possible to explicitly model many BGI features or accurately determine their effectiveness with respect to localised flood exposure. Consequently, the aim of this research is to develop a novel optimisation tool utilising a state-of-the-art fully dynamic hydrodynamic model such that BGI placement can be accurately optimised with respect to localised property scale vulnerability and hazard analyses.
		
		Examples of optimisation tools utilising fully dynamic hydrodynamic models are rare; however, the work of Ur Rehman et al. \cite{UrRehman2024}, provides an example of such an approach. The proposed approach couples CityCAT \cite{Glenis2018}, a fully dynamic hydrodynamic model, with the NSGA-II to optimise the placement of BGI. However, the proposed optimisation framework is restricted to the optimisation of the placement of permeable pavements and the chosen evolutionary algorithm is inefficient with respect to more modern optimisation algorithms. It is therefore the intention of this research to build upon this work and develop a more flexible tool, capable of optimising a variety of BGI features, in a more efficient manner, utilising more recent developments in evolutionary computing.
		
		Through the development of a robust, accurate and efficient optimisation tool, the aim is to streamline the design process and accelerate the adoption of nature-based solutions for flood risk management. By automating this procedure, the study seeks to improve decision-making in urban planning, ensuring the most effective placement and design of interventions for improved flood resilience and environmental benefits. As the complexity of urban environments continues to grow, the need for advanced, automated, multi-objective optimisation approaches will only become more critical in ensuring cities are resilient to future flood events.
		
		The remainder of this paper is structured as follows: Section 2 details the methodological framework, including the principles of evolutionary algorithms, the choice of genetic representation for BGI features, the use of hydrodynamic simulations within the objective function and the development of the proposed evolutionary algorithm. Section 3 presents the validation of the optimisation tool, firstly through comparisons with an exhaustive search of a simple but tractable search space and secondly through inter-algorithm comparisons for a search of an intractable, but more realistic, search space. Finally, Section 4 provides concluding remarks and potential avenues for future research and development.
		
		\section{Methods}
		\subsection{Evolutionary Algorithms}
		The optimisation tool employs an evolutionary algorithm to provide an automated process for determining the optimal implementation of blue-green infrastructure within urban environments. Evolutionary algorithms are inspired by the principles of natural selection and evolution, where the principle of survival of the fittest is used to guide an iterative search towards high quality solutions. By mimicking biological evolution, a population of candidate solutions are evolved over successive generations, with the aim of producing efficient solutions. Consequently, it is important to define key terms related to the evolutionary process:
		\begin{description}[style=nextline]
			\item \textbf{Phenotype}: The set of observable characteristics or features of a candidate solution in a real world context. The phenotype is the real-world manifestation of the genotype.
			\item \textbf{Genotype}: The genetic representation of a candidate solution. 
			\item \textbf{Chromosome}: A structured set of genes within the genotype, representing a single candidate solution. Each chromosome contains the information necessary to encode a complete solution.
			\item \textbf{Gene}: An individual unit within the chromosome that encodes a specific aspect or component of a candidate solution. Genes collectively form the genotype. 
			\item \textbf{Allele}: The specific value or variant a gene can take. In the context of blue-green infrastructure, an allele indicates the value of an attribute of a feature.
			\item \textbf{Fitness}: A measure of the quality of a candidate solution with respect to an objective. The fitness landscape represents the relationship between all genotypes and their respective fitness values and can be conceptualised as a topographical map where minima/maxima signify optimal solutions.
			\item \textbf{Mutation}: A genetic operator that introduces small random changes to the genotype to promote genetic diversity. Mutation can alter alleles within a chromosome to aid exploration of the search space.
			\item \textbf{Recombination (Crossover)}: A genetic operator that combines portions of two parent chromosomes to produce offspring which inherit characteristics from both. Recombination, also referred to as crossover, aids the exploration of the search space via exploitation of existing high-quality solutions.
		\end{description}
		To explore the fitness landscape efficiently, a balance between exploring new possibilities and exploiting high-performing solutions is essential.
		
		In the context of BGI, the genotype may represent a binary encoding of intervention placements, with the corresponding phenotype being the spatial layout of the interventions in the urban environment. The fitness of each candidate solution can then subsequently be calculated with respect to suitable objectives, such as the financial cost of implementation and reduction of flood risk. Consequently, the optimisation of the placement of blue-green infrastructure within urban environments is classified as a multi-objective optimisation problem (MOOP), whereby conflicting objectives must be simultaneously minimised. Evolutionary algorithms are a robust and effective choice for solving such problems due to their ability to maintain a diverse set of solutions when efficiently searching large and discontinuous search spaces \cite{Eiben2015}. 
		
		\subsubsection{Feature Representation}\label{Section: Feature Representation}
		To optimise the placement of BGI, a suitable genetic representation must be determined to map a phenotypical flood intervention to a genotype which can be manipulated by an evolutionary algorithm. As such, two styles of feature representation, \textit{zonal features} and \textit{local features}, are proposed and outlined. Through the use of the proposed feature representation styles, a flexible and efficient representation of a range of BGI is enabled. Regardless of the choice of feature representation style, the resultant genotype consists of a single bit-string formed by concatenating the individual bit-strings representing each BGI feature under consideration (Figure \ref{Fig: Genotype}). In the terminology of evolutionary algorithms, a candidate solution is genetically represented as a chromosome containing genes, with each gene representing a proposed flood intervention. The choice of feature representation therefore determines the genotype-phenotype mapping for each gene, enabling a phenotypic interpretation of the bit-coded alleles of each gene. 
		\begin{figure}[hbt!]
			\centering
			\begin{tikzpicture}
				\draw[thick, fill=gray, opacity=0.5] (-7.0, 0) rectangle (-7.5, 0.75); 
				\draw[thick, fill=gray, opacity=0.5] (-6.5, 0) rectangle (-7.0, 0.75); 
				\draw[thick, fill=gray, opacity=0.5] (-6.0, 0) rectangle (-6.5, 0.75); 
				\draw[thick, fill=gray, opacity=0.5] (-5.5, 0) rectangle (-6.0, 0.75); 
				\draw[thick] (-5.0, 0) rectangle (-5.5, 0.75); 
				\draw[thick] (-4.5, 0) rectangle (-5.0, 0.75); 
				\draw[thick] (-4.0, 0) rectangle (-4.5, 0.75); 
				\draw[thick] (-3.5, 0) rectangle (-4.0, 0.75); 
				\draw[thick] (-3.0, 0) rectangle (-3.5, 0.75); 
				\draw[thick] (-2.5, 0) rectangle (-3.0, 0.75); 
				\draw[thick, fill=gray, opacity=0.5] (-2.0, 0) rectangle (-2.5, 0.75); 
				\draw[thick, fill=gray, opacity=0.5] (-1.5, 0) rectangle (-2.0, 0.75); 
				\draw[thick, fill=gray, opacity=0.5] (-1.0, 0) rectangle (-1.5, 0.75); 
				\draw[thick, fill=gray, opacity=0.5] (-0.5, 0) rectangle (-1.0, 0.75);
				\draw[thick] (-0.0, 0) rectangle (-0.5, 0.75);
				\draw[thick] (0.0, 0) rectangle (0.5, 0.75);
				\draw[thick] (0.5, 0) rectangle (1.0, 0.75); 
				\draw[thick] (1.0, 0) rectangle (1.5, 0.75); 
				\draw[thick, fill=gray, opacity=0.5] (1.5, 0) rectangle (2.0, 0.75); 
				\draw[thick, fill=gray, opacity=0.5] (2.0, 0) rectangle (2.5, 0.75); 
				\draw[thick, fill=gray, opacity=0.5] (2.5, 0) rectangle (3.0, 0.75);
				\draw[thick, fill=gray, opacity=0.5] (3.0, 0) rectangle (3.5, 0.75);
				\draw[thick, fill=gray, opacity=0.5] (3.5, 0) rectangle (4.0, 0.75);
				\draw[thick, fill=gray, opacity=0.5] (4.0, 0) rectangle (4.5, 0.75);
				\draw[thick, fill=gray, opacity=0.5] (4.5, 0) rectangle (5.0, 0.75);
				\draw[thick, fill=gray, opacity=0.5] (5.0, 0) rectangle (5.5, 0.75);
				\draw[thick] (5.5, 0) rectangle (6.0, 0.75);
				\draw[thick] (6.0, 0) rectangle (6.5, 0.75);
				\draw[thick] (6.5, 0) rectangle (7.0, 0.75);
				\draw[thick] (7.0, 0) rectangle (7.5, 0.75);
				\node at (-7.25, 0.3725) {\large 1};
				\node at (-6.75, 0.3725) {\large 0};
				\node at (-6.25, 0.3725) {\large 0};
				\node at (-5.75, 0.3725) {\large 1};
				\node at (-5.25, 0.3725) {\large 1};
				\node at (-4.75, 0.3725) {\large 1};
				\node at (-4.25, 0.3725) {\large 0};
				\node at (-3.75, 0.3725) {\large 0};
				\node at (-3.25, 0.3725) {\large 1};
				\node at (-2.75, 0.3725) {\large 0};
				\node at (-2.25, 0.3725) {\large 1};
				\node at (-1.75, 0.3725) {\large 0};
				\node at (-1.25, 0.3725) {\large 1};
				\node at (-0.75, 0.3725) {\large 0};
				\node at (-0.25, 0.3725) {\large 1};
				\node at (0.25, 0.3725) {\large 0};
				\node at (0.75, 0.3725) {\large 0};
				\node at (1.25, 0.3725) {\large 0};
				\node at (1.75, 0.3725) {\large 1};
				\node at (2.25, 0.3725) {\large 1};
				\node at (2.75, 0.3725) {\large 0};
				\node at (3.25, 0.3725) {\large 0};
				\node at (3.75, 0.3725) {\large 1};
				\node at (4.25, 0.3725) {\large 0};
				\node at (4.75, 0.3725) {\large 1};
				\node at (5.25, 0.3725) {\large 1};
				\node at (5.75, 0.3725) {\large 0};
				\node at (6.25, 0.3725) {\large 0};
				\node at (6.75, 0.3725) {\large 1};
				\node at (7.25, 0.3725) {\large 0};
				\draw[decorate,decoration={brace,amplitude=10pt}] (7.5, 0) --  (-7.5, 0)
				node[midway,below=10pt] {\large Candidate Solution};
				\draw[decorate,decoration={brace,amplitude=10pt}] (-7.5, 0.75) --  (-5.5, 0.75)
				node[midway,above=10pt] {\large Feature 1};
				\draw[decorate,decoration={brace,amplitude=10pt}] (-5.5, 0.75) --  (-2.5, 0.75)
				node[midway,above=10pt] {\large Feature 2};
				\draw[decorate,decoration={brace,amplitude=10pt}] (-2.5, 0.75) --  (-0.5, 0.75)
				node[midway,above=10pt] {\large Feature 3};
				\draw[decorate,decoration={brace,amplitude=10pt}] (-0.5, 0.75) --  (1.5, 0.75)
				node[midway,above=10pt] {\large Feature 4};
				\draw[decorate,decoration={brace,amplitude=10pt}] (1.5, 0.75) --  (5.5, 0.75)
				node[midway,above=10pt] {\large Feature 5};
				\draw[decorate,decoration={brace,amplitude=10pt}] (5.5, 0.75) --  (7.5, 0.75)
				node[midway,above=10pt] {\large Feature 6};
			\end{tikzpicture}
			\caption{Example genetic representation of the placement of a range of BGI within a spatial domain. The representation of the candidate solution for the domain is composed of a concatenation of the bit-string representations for each of the features under consideration.}
			\label{Fig: Genotype}
		\end{figure}
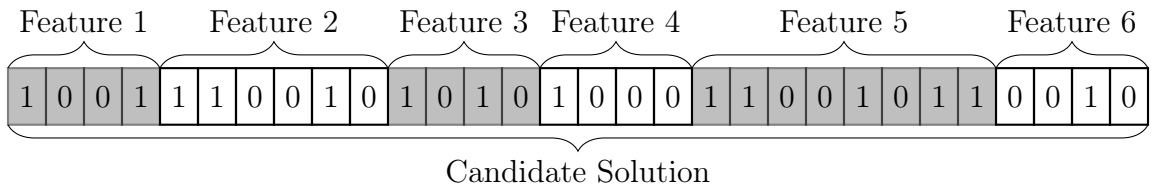
		
		Using bit-strings to encode non-binary information can be erroneous \cite{Eiben2015} however, in the context of the optioneering of BGI, it is argued that the pursuit of real-valued or floating-point precision amidst considerable uncertainty is unwarranted and inefficient. A discrete representation is therefore considered to be justified since the aim of the optimisation tool is to inform the optioneering process, requiring further refinement via detailed design prior to real-world implementation. Furthermore, a binary representation is preferred as fundamental theory states that binary alphabets enable optimal schema processing compared to alternative high-cardinality alphabets such as floating-point or real-valued codings \cite{Goldberg1989, Holland1975}. Although practitioners have experienced success using high-cardinality alphabets such as real-valued codings, schema theory demonstrates that algorithms utilising such codings may be blocked from converging to global optima in some circumstances \cite{Goldberg1991a}.
		
		\subsubsection*{Zonal Features}\label{Section: Zonal}
		The first style of feature representation is referred to as a \textit{zonal feature}. These are features which can be modelled as a subset of the spatial domain with a Boolean state, in which the entirety of the zone either includes or excludes the specified flood intervention. BGI which is highly suitable for treatment as a zonal feature includes permeable paving, and green roofing and rain gardens. However, it is worth noting that any individual BGI feature or suitable combination of BGI features may be represented as zonal features provided their characteristics are fixed such that only their state of inclusion is of concern. As a Boolean decision variable, a straightforward genotype-phenotype mapping is implemented in which the $n^{\textit{th}}$ bit encodes the state of the $n^{\textit{th}}$ zone, whereby a `$1$' denotes inclusion of the flood intervention within the zone and `$0$' denotes the exclusion (Figure \ref{Fig: Zonal Feature}).
		\begin{figure}[hbt!]
			\centering
			\resizebox{0.65\textwidth}{!}{
			\begin{tikzpicture}
				\fill[gray!50]
				(-10,0)
				.. controls (-10.0,0.0) and (-9.0,-0.5) .. (-7.5,-1.0)  
				.. controls (-7.3,-0.9) and (-7.1,-0.85) .. (-7.0,-0.7)  
				.. controls (-6.9,2.6) and (-7.1,1.0) .. (-7.4,0.6)
				.. controls (-7.8,0.55) and (-8.0,0.5) .. (-8.9,0.4) 
				.. controls (-9.1,0.3) and (-9.4,0.1) .. (-10.0,0.0);
				\fill[gray!50] (-7.0,0.6) -- (-5.0,0.6) -- (-5.0,-0.9) -- (-7.0,-0.7) -- (-7.0,0.6) -- cycle;
				\fill[gray!50] (-5.0,0.9) -- (-3.0,0.9) -- (-3.0,-1.2) -- (-5.0,-0.9) -- (-5.0,-0.9) -- cycle;
				\fill[gray!50] (-3.0,2.6) -- (2.8,2.6) -- (2.8,-2.0) -- (-0.2,-2.0) -- (-0.2,0.25) -- (-3,0.25) -- (-3.0,2.6) -- cycle;
				\draw[thick] 
				(-10,0.0) 
				.. controls (-10.0,0.0) and (-9.0,-0.5) .. (-7.5,-1.0)  
				.. controls (-7.3,-0.9) and (-7.1,-0.85) .. (-7.0,-0.7)  
				.. controls (-6.5,-0.75) and (-5.0,-0.85) .. (-2.5,-1.3)  
				.. controls (-2.45,-1.35) and (-2.3,-1.7) .. (-2.25,-2.2)  
				.. controls (-2.2,-2.0) and (-1.0,-3.0) .. (-0.25,-4.7)  
				.. controls (0.0,-4.6) and (0.6,-4.3) .. (1.9,-4.0)  
				.. controls (2.3,-3.6) and (2.5,-3.0) .. (3.0,-2.5)  
				.. controls (3.5,-2.25) and (5.0,-2.0) .. (7.2,-1.9)  
				.. controls (7.0,-1.7) and (6.4,-1.2) .. (6.0,-0.8)  
				.. controls (5.8,-0.9) and (5.5,-0.85) .. (5.0,-0.9)  
				.. controls (4.75,-0.7) and (4.5,-0.5) .. (4.0,0.0)  
				.. controls (3.8,0.5) and (3.25,1.5) .. (2.8,2.6)  
				.. controls (2.9,2.5) and (3.2,3.4) .. (3.5,4.0)  
				.. controls (3.25,4.2) and (3.0,4.5) .. (2.5,5.0)  
				.. controls (2.25,4.9) and (0.5,4.8) .. (0.2,4.7)  
				.. controls (-0.2,4.2) and (-0.7,4.0) .. (-1.25,3.0)  
				.. controls(-1.5,2.8) and (-2.7,2.75) .. (-3.0,2.6)  
				.. controls (-3.6,2.7) and (-4.4,2.55) .. (-5.2,2.5)  
				.. controls (-5.4,2.35) and (-5.9,2.25) .. (-6.8,2.0)  
				.. controls (-6.95,1.7) and (-7.1,1.0) .. (-7.4,0.6)  
				.. controls (-7.8,0.55) and (-8.0,0.5) .. (-8.9,0.4)  
				.. controls (-9.1,0.3) and (-9.4,0.1) .. (-10.0,0.0); 
				\draw[thick] (-7.0,-0.7) -- (-7.0,1.5); 
				\draw[thick] (-7.0,0.6) -- (-5.0,0.6); 
				\draw[thick] (-5.0,-0.9) -- (-5.0,2.5); 
				\draw[thick] (-5.0,0.9) -- (-3.0,0.9); 
				\draw[thick] (-3.0,-1.2) -- (-3.0,2.6); 
				\draw[thick] (-3.0,2.6) -- (2.8,2.6); 
				\draw[thick] (2.8,2.6) -- (2.8,-2.7); 
				\draw[thick] (-2.3,-2.0) -- (2.8,-2.0); 
				\draw[thick] (-0.2,-2) -- (-0.2,2.6); 
				\draw[thick] (-3.0,0.25) -- (2.8,0.25); 
				\node at (-8,-0.1) {\large Zone 1};
				\node at (-6,1.35) {\large Zone 2};
				\node at (-6,0) {\large Zone 3};
				\node at (-4,1.65) {\large Zone 4};
				\node at (-4,-0.1) {\large Zone 5};
				\node at (1,3.5) {\large Zone 6};
				\node at (-1.6,1.4) {\large Zone 7};
				\node at (1.3,1.4) {\large Zone 8};
				\node at (-1.6,-0.8) {\large Zone 9};
				\node at (1.3,-0.8) {\large Zone 10};
				\node at (4,-1.2) {\large Zone 11};
				\node at (0.5,-3) {\large Zone 12};
				\draw[thick, fill=gray, opacity=0.5] (-10.0, 3.5) rectangle (-9.5, 4.25);
				\node at (-9.75, 3.8725) {\large 1};
				\draw[thick, fill=white, opacity=0.5] (-9.5, 3.5) rectangle (-9.0, 4.25);
				\node at (-9.25, 3.8725) {\large 0};
				\draw[thick, fill=gray, opacity=0.5] (-9.0, 3.5) rectangle (-8.5, 4.25);
				\node at (-8.75, 3.8725) {\large 1};
				\draw[thick, fill=white, opacity=0.5] (-8.5, 3.5) rectangle (-8.0, 4.25);
				\node at (-8.25, 3.8725) {\large 0};
				\draw[thick, fill=gray, opacity=0.5] (-8.0, 3.5) rectangle (-7.5, 4.25);
				\node at (-7.75, 3.8725) {\large 1};
				\draw[thick, fill=white, opacity=0.5] (-7.5, 3.5) rectangle (-7.0, 4.25);
				\node at (-7.25, 3.8725) {\large 0};
				\draw[thick, fill=gray, opacity=0.5] (-7.0, 3.5) rectangle (-6.5, 4.25);
				\node at (-6.75, 3.8725) {\large 1};
				\draw[thick, fill=gray, opacity=0.5] (-6.5, 3.5) rectangle (-6.0, 4.25);
				\node at (-6.25, 3.8725) {\large 1};
				\draw[thick, fill=white, opacity=0.5] (-6.0, 3.5) rectangle (-5.5, 4.25);
				\node at (-5.75, 3.8725) {\large 0};
				\draw[thick, fill=gray, opacity=0.5] (-5.5, 3.5) rectangle (-5.0, 4.25);
				\node at (-5.25, 3.8725) {\large 1};
				\draw[thick, fill=white, opacity=0.5] (-5.0, 3.5) rectangle (-4.5, 4.25);
				\node at (-4.75, 3.8725) {\large 0};
				\draw[thick, fill=white, opacity=0.5] (-4.5, 3.5) rectangle (-4.0, 4.25);
				\node at (-4.25, 3.8725) {\large 0};
				\node at (-7, 3.2) {\large Genotype};
				\node at (-4, -2) {\large Phenotype};
			\end{tikzpicture}
			}
			\caption{Genotype-phenotype mapping for a \textit{zonal feature}. The example genotype provides a genetic representation of a candidate solution in which zones 1,3,5,7,8 and 10 of the spatial domain include the specified zonal flood intervention.}
			\label{Fig: Zonal Feature}
		\end{figure}
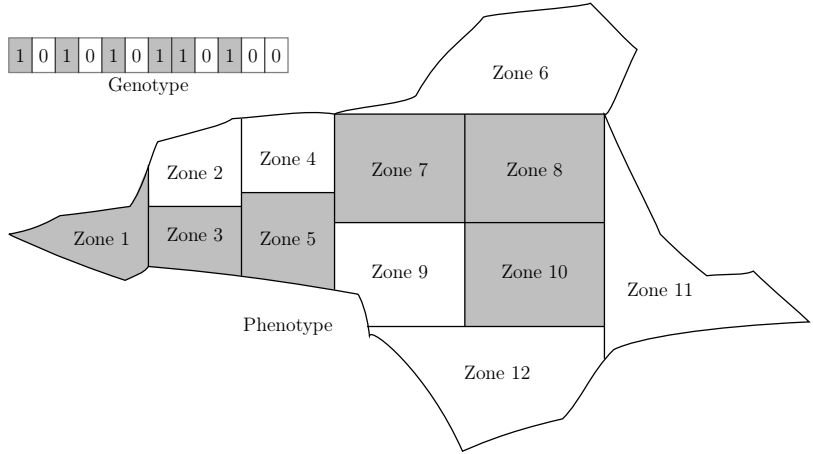
		
		Zones must be delineated in a disjoint manner, as shown in Figure \ref{Fig: Zonal Feature}, to prevent redundancy from hindering the effectiveness of evolutionary search \cite{Rothlauf2003, Rothlauf2006}. Additionally, the zones do not need to cover the entire spatial domain. However, it is essential to ensure that each zone has a distinct state of inclusion and exclusion. Specifically, if a zone contains no suitable locations for a feature, it should be omitted to prevent redundancy. Zones may be arbitrarily delineated via an automated process or manually delineated to incorporate domain knowledge. A chromosome of length $L$, representing $L$ zones, results in a genotypic search space of cardinality $2^L$ and therefore a \textit{quality-time} balance must be sought when determining the number of the zones included in the genetic representation. This is the case as, although there does not exist a definitive proof of time to convergence for evolutionary algorithms applied to MOOPs, analyses suggest that the time to convergence grows super-linearly with the size of the search space \cite{Laumanns2004, Giel2010, Osuna2020, Doerr2022, Zheng2022, Doerr2023}.
		
		\subsubsection*{Local Features}\label{Section: Local}
		The second style of feature representation is referred to as a \textit{local feature}. This style of representation is intended for features which are incompatible with a zonal representation, where a simple Boolean decision variable provides inadequate capacity to represent and optimise the component characteristics of the feature. To achieve this, a suitable discrete-binary mapping is implemented to represent a discretised value of each component characteristic. A component characteristic can be any state variable that modifies the effectiveness of the feature such as the location, material properties or sizing. It is worth emphasising that only characteristics which modify the effectiveness of the feature should be encoded, as otherwise unnecessary redundancy will be introduced, diminishing the effectiveness of the evolutionary search process. For characteristics which are continuous in nature, this must be complemented with a suitable continuous-discrete mapping. 
		
		The overall mapping from phenotype to genotype for a continuous component characteristic is therefore described in a similar manner to Lim et al. \cite{Lim2011}:
		\begin{equation}\label{eq: overall mapping}
			x^{(g)}_n = \mathcal{G}_n\left(x^{(d)}_n\right) = \mathcal{G}_n\left(\mathcal{F}_n\left(x^{(c)}_n\right)\right)
		\end{equation}
		where $x^{(g)}_n$ is the binary encoded genotype for the $n^\mathrm{th}$ component characteristic, $x^{(c)}_n$ is the $n^\mathrm{th}$ continuous component characteristic and $x^{(d)}_n$ is the corresponding discretised value of the characteristic with $\mathcal{F}_n : \mathbb{R} \to D \subset \mathbb{R}$ being the continuous-discrete map and $\mathcal{G}_n : D \to \{0,1\}^L$ being the discrete-binary map. For further clarification, the overall mapping is illustrated in Figure \ref{Fig: Mapping}.
		\begin{figure}[hbt!]
			\centering
			\begin{tikzpicture}
				
				\newcommand{\ovalDistance}{3.25cm}
				\newcommand{\ovalHeight}{3.50cm}
				\newcommand{\ovalWidth}{2.15cm}
				
				\node (oval1) at (0,0) [draw, ellipse, minimum width=\ovalWidth, minimum height=\ovalHeight] {$x^{(c)}_n \in \mathbb{R}$};
				\node (oval2) at (\ovalDistance,0) [draw, dashed, ellipse, minimum width=0.60*\ovalWidth, minimum height=\ovalHeight] {$i_n \in \mathbb{Z}$};
				\node (oval3) at (2*\ovalDistance,0) [draw, ellipse, minimum width=\ovalWidth, minimum height=\ovalHeight] {};
				\node (oval3a) at (2*\ovalDistance, 0.25) {$x^{(d)}_n \in D$};
				\node (oval3b) at (2*\ovalDistance, -0.25) {\small$(D \subset \mathbb{R})$};
				\node (oval4) at (3*\ovalDistance,0) [draw, dashed, ellipse, minimum width=0.60*\ovalWidth, minimum height=\ovalHeight] {$i_n \in \mathbb{Z}$};
				\node (oval5) at (4*\ovalDistance,0) [draw, ellipse, minimum width=\ovalWidth, minimum height=\ovalHeight] {$x^{(g)}_n \in \{0,1\}^L$};
				
				\draw[->, thick] (oval1) -- node[above] {$f_{n,1}$} (oval2);
				\draw[->, thick] (oval2) -- node[above] {$f_{n,2}$} (oval3);
				\draw[->, thick] (oval3) -- node[above] {$g_{n,1}$} (oval4);
				\draw[->, thick] (oval4) -- node[above] {$g_{n,2}$} (oval5);
				
				\newcommand{\textModifier}{0.90}
				\node[below] at (0, -\textModifier*\ovalWidth) {Continuous Phenotype} (oval1);
				\node[below] at (\ovalDistance, -\textModifier*\ovalWidth) {Discrete Index} (oval2);
				\node[below] at (2*\ovalDistance, -\textModifier*\ovalWidth) {Discrete Phenotype} (oval3);
				\node[below] at (3*\ovalDistance, -\textModifier*\ovalWidth) {Discrete Index} (oval4);
				\node[below] at (4*\ovalDistance, -\textModifier*\ovalWidth) {Binary Genotype} (oval5);
				
				\draw[->, thick] (oval1.north east) -- node[above] {$\mathcal{F}_n$} (oval3.north west);
				\draw[->, thick] (oval3.north east) -- node[above] {$\mathcal{G}_n$} (oval5.north west);
				
			\end{tikzpicture}
			\caption{Overall mapping from phenotype to genotype for a continuous component characteristic as described in equation (\ref{eq: overall mapping}). The continuous-discrete map, $\mathcal{F} = f_{n,2} \circ f_{n,1}$, maps a continuous phenotype, $x^{(c)}_n$, to a discrete phenotype, $x^{(d)}_n$. The discrete-binary map, $\mathcal{G} = g_{n,2} \circ g_{n,1}$, maps a discrete phenotype, $x^{(d)}_n$, to a binary encoded genotype, $x^{(g)}_n$.}
			\label{Fig: Mapping}
		\end{figure}
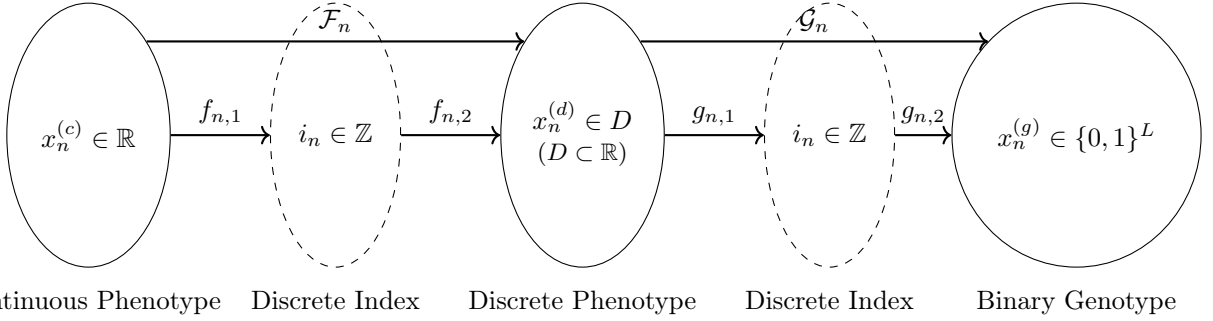
		
		The continuous-discrete map, $\mathcal{F}$, is defined as the composite function $f_{n,2} \circ f_{n,1}$, where $f_{n,1} : \mathbb{R} \to \mathbb{Z}$ maps the continuous variable to a discrete index, $i_n$, and $f_{n,2} : \mathbb{Z} \to D \subset \mathbb{R}$ maps the discrete index, $i_n$, to a discrete phenotype, $x^{(d)}_{n, i_n} \in D$. The proposed composite continuous-discrete map is defined as:
		\begin{gather}
			i_n = f_{n,1}\left(x^{(c)}_n\right) = \begin{dcases}
				0 \quad \textrm{for} \quad x^{(c)}_n \leq x^{(c)}_{n,\mathrm{min}} + \frac{\epsilon_n}{2} \\
				\left\lfloor \frac{x^{(c)}_n - \left(x^{(c)}_{n,\mathrm{min}} + \frac{\epsilon_n}{2}\right)}{\epsilon_n} \right\rfloor + 1 \quad \mathrm{for} \quad x^{(c)}_{n,\mathrm{min}} + \frac{\epsilon_n}{2}  < x^{(c)}_n < x^{(c)}_{n,\mathrm{max}} - \frac{\epsilon_n}{2} \\
				N_n-1 \quad \mathrm{for} \quad x^{(c)}_n \geq x^{(c)}_{n,\mathrm{max}} - \frac{\epsilon_n}{2} 
			\end{dcases} \\ \notag \\
			x^{(d)}_{n, i_n} = f_{n,2}(i_n) = x^{(c)}_{n,\mathrm{min}} + i_n\epsilon_n
		\end{gather}
		where:
		\begin{equation}
			\epsilon_n = \frac{x^{(c)}_{n,\mathrm{max}}-x^{(c)}_{n,\mathrm{min}}}{N_n-1}
		\end{equation}
		where $x^{(c)}_{n,\mathrm{min}}$ and $x^{(c)}_{n,\mathrm{max}}$ are user defined bounds upon the continuous characteristic and $N_n$ denotes the number of unique phenotypes encoded by the encoding scheme. 
		
		As $D = \{x^{(c)}_{n,\mathrm{min}} + i_n\epsilon_n \,|\, i_n = 0, 1, \dots, N_n-1 \}$, the granularity of the discretisation is dependent upon $\epsilon_n$, equal to the spacing between the elements of $D$, which is in turn dependent upon the selected bounds, $x^{(c)}_{n,\mathrm{min}}$ and $x^{(c)}_{n,\mathrm{max}}$, and $N_n$. As the user defined bounds should ideally be selected based upon physical principles, such as spatial or design constraints for the characteristic, the primary control upon the granularity of the representation is $N_n$. $N_n$ is modified by changing the length of the bit-string, dependent upon the scaling of the chosen representation (Figure \ref{Fig: Binary Encoding}). However, it should be noted that a finer granularity enlarges the search space and diminishes the capacity of the mutation operator in aiding convergence \cite{Hinterding1995}. As a consequence, coarser representations are recommended where appropriate. This does however compromise the quality of the representation of the search space for continuous characteristics \cite{Veldhuizen1999b} meaning optimisation of the characteristics is compatible with an initial optioneering process rather than detailed design.
		
		In the scenario where a characteristic is inherently discrete, consider for example water butts which are manufactured to retain specific volumes of rain water, the continuous-discrete map is superfluous and $D$ can be straightforwardly constructed as an ordered set containing an appropriate selection of the permissible discrete values. It is preferable, if possible, for the set of discrete values to be equally spaced, as constructed by the proposed continuous-discrete map, in order to promote high locality by ensuring consistent genotypic-phenotypic distances \cite{Rothlauf2006}. Furthermore, the length of the binary representation, scaling of the representation and the number of discrete values included in $D$ should be selected such that the cardinality of $D$ is equal to $N_n$, ensuring a bijective mapping with no phenotypic redundancy, which is where multiple genotypes map to the same phenotype.
		
		The discrete-binary map, $\mathcal{G}$, is defined as the composite function $g_{n,2} \circ g_{n,1}$, where $g_{n,1} : D \to \mathbb{Z}$ maps the discrete value $x^{(d)}_n$ to an index, $i_n$, which is mapped to a bit-string of length $L$ via a suitable integer-binary encoding scheme, $g_{n,2} : \mathbb{Z} \to \{0,1\}^L$. Where $D$ is constructed via a continuous-discrete map, $\mathcal{F}_n$:
		\begin{equation}
			i_n = g_{n,1}\left(x^{(d)}_n\right) = \frac{x^{(d)}_n - x^{(c)}_{n,\mathrm{min}}}{\epsilon_n} = f_{n,2}^{-1}\left(x^{(d)}_n\right)
		\end{equation}
		Alternatively, where $D$ is constructed from a inherently discrete phenotype:
		\begin{equation}
			i_n = g_{n,1}\left(x^{(d)}_n\right) = \argmin_{i_n \in [0, N_n-1]} \left\vert x^{(d)}_n - x^{(d)}_{n, i_n} \right\vert
		\end{equation}
		The integer-binary map, $g_{n,1}$, is referred to as a encoding scheme, $g_{n,1}(i_n) = \mathcal{C}_n(i_n)$, where $\mathcal{C}_n(i_n)$ denotes the $i_n^{\mathrm{th}}$ element in the Hamming space for the $n^{\mathrm{th}}$ component characteristic, referred to as a codeword \cite{Lim2011}. 
		
		Careful consideration is required when selecting a suitable encoding scheme, as the corresponding redundancy, scaling and locality properties impact the quality of solutions and the time to convergence for the search. In this regard, the representation theory presented by Rothlauf \cite{Rothlauf2006} provides a theoretical basis upon which genetic representations can be assessed in the context of a \textit{quality-time} framework. Implemented within the optimisation tool are the three most common methods for encoding discrete phenotypes: standard binary, Gray binary and unary representations (Figure \ref{Fig: Binary Encoding}). 
		
		For standard binary representations, $(x^{(g)}_n)^{\mathrm{bin}} \in \{0,1\}^L = \{b_0^{\mathrm{bin}}, b_1^{\mathrm{bin}}, \dots , b_{L-1}^{\mathrm{bin}}\}$, integers are converted directly into their binary equivalent \cite{Rothlauf2006}:
		\begin{gather}
			x^{(g)}_n = g_{n,2}(i_n) = \{b_0^{\mathrm{bin}}, b_1^{\mathrm{bin}}, \dots b_{L-1}^{\mathrm{bin}}\} \quad \textrm{ where } \quad b_i^{\mathrm{bin}} = \left\lfloor\frac{i_n}{2^i}\right\rfloor\bmod{2} \\
			i_n = g_{n,2}^{-1}\left(x^{(g)}_n\right) = \sum_{i=0}^{L-1} 2^i b_{i}
		\end{gather}
		where $b_i$ represents the $i^{\mathrm{th}}$ bit of $x^{(g)}_n$. A standard binary representation encodes $N = \log_2(L)$ phenotypes for a bit-string of length $L$. 
		
		There are $2^L(L!)$ Gray codes for a binary space represented by bit-strings of length $L$, however, the focus is placed upon binary reflected Gray code \cite{Gray1953}, which is the most commonly implemented Gray encoding for evolutionary algorithms \cite{Hinterding1995, Chakraborty2003, Mathias1994, Greiner2005, Whitley1999}. Gray encoding can be produced from a standard binary bit-string via \cite{Rothlauf2006, Chakraborty2003, Mathias1994, Chiam2006}:
		\begin{equation}
			\left(x^{(g)}_n\right)^{\mathrm{Gray}} = \begin{dcases}
				b_i^{\mathrm{bin}} \quad \mathrm{if} \quad i = 0, \\
				b_{i-1}^{\mathrm{bin}} \oplus b_{i}^{\mathrm{bin}} \quad \mathrm{otherwise}
			\end{dcases}
		\end{equation}
		where $\oplus$ denotes exclusive OR (XOR) operation, which is equivalent to the modulo-2 sum. This converts the standard binary encoded bit-string to the Gray-encoded bit-string $(x^{(g)}_n)^{\mathrm{Gray}} \in \{0,1\}^L = \{b_0^{\mathrm{Gray}}, b_1^{\mathrm{Gray}}, \dots , b_{L-1}^{\mathrm{Gray}}\}$. Gray code can be decoded to standard binary via \cite{Rothlauf2006}:
		\begin{equation}
			b_i^{\mathrm{bin}} = \underset{j=0}{\overset{i}{\bigoplus}} \, b_j^{\mathrm{Gray}}
		\end{equation}
		where $\bigoplus$ represents the cumulative XOR operation over all Gray code bits from $j=0$ to $j=i$. As for standard binary representations, Gray codes enable the compact encoding of $N = \log_2(L)$ phenotypes for a bit-string of length $L$.
		
		Unary representations encode integers through bit summation \cite{Rothlauf2006}:
		\begin{equation}\label{eq: unary encoding}
			i_n = g_n^{-1}(x^{(g)}_n) = \sum^{L-1}_{i=0} b_i
		\end{equation}
		where $b_i$ represents the $i^{\mathrm{th}}$ bit of $x^{(g)}_n$. Unary representations encode $N = L+1$ unique phenotypes for a bit-string of length $L$. Unlike standard and Gray binary representations, which are one-to-one mappings, unary representations are a many-to-one mapping, resulting in redundancy within the encoding scheme (Figure \ref{Fig: Binary Encoding}a). 
		\begin{figure}[hbt!]
			\centering
			\begin{tikzpicture}
				\draw[thick] (-1.0,-1.5) -- (-1.0,3.0); 
				\draw[thick] (-8.0,-1.5) -- (-8.0,3.0); 
				\draw[thick] (-8.0,3.0) -- (-1.0,3.0); 
				\draw[thick] (-7.0,2.6) -- (-1.0,2.6); 
				\draw[thick] (-8.0,1.8) -- (-1.0,1.8); 
				\draw[thick] (-8.0,-1.5) -- (-1.0,-1.5); 
				\draw[thick] (-7.0,-1.5) -- (-7.0,3.0); 
				\draw[thick] (-5.5,-1.5) -- (-5.5,2.6); 
				\draw[thick] (-4.0,-1.5) -- (-4.0,2.6); 
				\node at (-7.5,2.3) {\large $i_n$};
				\node at (-4.25,2.8) {\small $x^{(g)}_n$};
				\node at (-6.25,2.4) {\normalsize Standard};
				\node at (-6.25,2.0) {\normalsize  Binary};
				\node at (-4.75,2.4) {\normalsize Gray};
				\node at (-4.75,2.0) {\normalsize  Binary};
				\node at (-2.5,2.2) {\normalsize Unary};
				\node at (-7.5,1.55) {\large $0$}; 
				\node at (-7.5+1.25,1.55) {\large `000'}; 
				\node at (-7.5+2.75,1.55) {\large `000'}; 
				\node at (-7.5+5,1.55) {\large `000'}; 
				\node at (-7.5,1.55-1*0.4) {\large $1$}; 
				\node at (-7.5+1.25,1.55-1*0.4) {\large `001'}; 
				\node at (-7.5+2.75,1.55-1*0.4) {\large `001'}; 
				\node at (-7.5+5,1.55-1*0.4) {\large `100' `010' `001'}; 
				\node at (-7.5,1.55-2*0.4) {\large $2$}; 
				\node at (-7.5+1.25,1.55-2*0.4) {\large `010'}; 
				\node at (-7.5+2.75,1.55-2*0.4) {\large `011'}; 
				\node at (-7.5+5,1.55-2*0.4) {\large `110' `011' `101'}; 
				\node at (-7.5,1.55-3*0.4) {\large $3$}; 
				\node at (-7.5+1.25,1.55-3*0.4) {\large `011'}; 
				\node at (-7.5+2.75,1.55-3*0.4) {\large `010'}; 
				\node at (-7.5+5,1.55-3*0.4) {\large `111'}; 
				\node at (-7.5,1.55-4*0.4) {\large $4$}; 
				\node at (-7.5+1.25,1.55-4*0.4) {\large `100'}; 
				\node at (-7.5+2.75,1.55-4*0.4) {\large `110'}; 
				\node at (-7.5+5,1.55-4*0.4) {\large -}; 
				\node at (-7.5,1.55-5*0.4) {\large $5$}; 
				\node at (-7.5+1.25,1.55-5*0.4) {\large `101'}; 
				\node at (-7.5+2.75,1.55-5*0.4) {\large `111'}; 
				\node at (-7.5+5,1.55-5*0.4) {\large -}; 
				\node at (-7.5,1.55-6*0.4) {\large $6$}; 
				\node at (-7.5+1.25,1.55-6*0.4) {\large `110'}; 
				\node at (-7.5+2.75,1.55-6*0.4) {\large `101'}; 
				\node at (-7.5+5,1.55-6*0.4) {\large -}; 
				\node at (-7.5,1.55-7*0.4) {\large $7$}; 
				\node at (-7.5+1.25,1.55-7*0.4) {\large `111'}; 
				\node at (-7.5+2.75,1.55-7*0.4) {\large `100'}; 
				\node at (-7.5+5,1.55-7*0.4) {\large -}; 
				\node at (-4.5,-1.9) {\small (a): Binary encoding, $\mathbb{C} : i_n \mapsto x^{(g)}_n$, for a};
				\node at (-4.5,-2.3) {\small bit-string of length $3$.};
				\draw[thick] (0.0,-1.3) -- (0.0,2.0); 
				\draw[thick] (8.2,-1.3) -- (8.2,3.0); 
				\draw[thick] (2.0,-1.3) -- (2.0,3.0); 
				\draw[thick] (4.0,-1.3) -- (4.0,3.0); 
				\draw[thick] (6.0,-1.3) -- (6.0,3.0); 
				\draw[thick] (2.0,3.0) -- (8.2,3.0); 
				\draw[thick] (0.0,2.0) -- (8.2,2.0); 
				\draw[thick] (0.0,2.0-1.1) -- (8.2,2.0-1.1); 
				\draw[thick] (0.0,2.0-2*1.1) -- (8.2,2.0-2*1.1); 
				\draw[thick] (0.0,2.0-3*1.1) -- (8.2,2.0-3*1.1); 
				\node at (3.0,2.6) {\normalsize Standard};
				\node at (3.0,2.2) {\normalsize  Binary};
				\node at (5.0,2.6) {\normalsize Gray};
				\node at (5.0,2.2) {\normalsize  Binary};
				\node at (7.1,2.4) {\normalsize Unary};
				
				\node at (1.0,1.65) {\normalsize Redundancy};
				\node at (3.0,1.70) {\small Redundancy};
				\node at (3.0,1.70-0.3) {\small free};
				\node at (3.0+2,1.70) {\small Redundancy};
				\node at (3.0+2,1.70-0.3) {\small free};
				\node at (3.1+2*2,1.75) {\small Non-};
				\node at (3.1+2*2,1.75-0.3) {\small synonymously};
				\node at (3.1+2*2,1.75-0.6) {\small redundant};
				\node at (1.0,1.65-1*1.1) {\normalsize Scaling};
				\node at (3.0,1.65-1*1.1) {\small exponential};
				\node at (3.0,1.65-1*1.1-0.4) {\small $N = \log_2(L)$};
				\node at (5.0,1.65-1*1.1) {\small exponential};
				\node at (5.0,1.65-1*1.1-0.4) {\small $N = \log_2(L)$};
				\node at (7.1,1.65-1*1.1) {\small linear};
				\node at (7.1,1.65-1*1.1-0.4) {\small $N = L+1$};
				\node at (1.0,1.65-2*1.1) {\normalsize Locality};
				\node at (3.0,1.65-2*1.1) {\small low};
				\node at (5.0,1.65-2*1.1) {\small low};
				\node at (7.1,1.65-2*1.1) {\small low};
				\node at (4.1,-1.9+0.4) {\small (b): Representation properties as per Rothlauf \cite{Rothlauf2006}. $N$ };
				\node at (4.1,-1.9) {\small denotes the number of unique phenotypes encoded };
				\node at (4.1,-1.9-0.4) {\small by a bit-string of length $L$.};
			\end{tikzpicture}
			\caption{Comparison between the three genetic representations used to encode a discretised component characteristic as a bit-string.}
			\label{Fig: Binary Encoding}
		\end{figure}
		
		As a non-synonymously redundant representation, where genotypes that represent the same phenotype are not similar and not neighbours in the mutational space, the unary representation disrupts genetic operators resulting in an inefficient evolutionary search \cite{Rothlauf2003}. Redundancy is however not inherently detrimental to evolutionary search, with Rothlauf and Goldberg \cite{Rothlauf2003} demonstrating that synonymously redundant representations, such as the trivial voting mapping \cite{Shackleton2000}, can enhance the efficiency of evolutionary search. However, this requires \textit{a priori} knowledge of the optimal genotypic building blocks, which are challenging to identify for the MOOPs under consideration. In this regard, the redundancy free nature of standard and Gray binary encodings is advantageous. 
		
		The locality of a representation describes the similarity of the neighbourhood structure across the phenotypic and genotypic search spaces. Rothlauf \cite{Rothlauf2006} demonstrates that high locality, corresponding to the scenario in which the genotypic and phenotypic neighbours are consistent, is required to ensure that problem difficulty is not modified, meaning problems of bounded difficulty can be solved reliably. Since the presented representations are all of low locality (Figure \ref{Fig: Binary Encoding}), the problem difficulty is modified. Consequently, it is not easily verifiable \textit{a priori} whether the evolutionary search will outperform a random search for a given problem. 
		
		Whitley \cite{Whitley1999} contributed to the resolution of this challenge by presenting a Free Lunch theorem for Gray binary representations versus standard binary representations indicating that Gray codes are expected to reduce problem difficulty for realistic problems. Whitley's proof demonstrates, via the No Free Lunch theorem \cite{Wolpert1997}, that since Gray codes preserve more optima on average over a set of \textit{worst case} functions, Gray codes will on average induce fewer optima than standard binary codes for a set of functions which are more likely to be encountered in practice. In contrast, Rothlauf \cite{Rothlauf2006} reports that the converse is true for recombination-based search approaches, juxtaposing the result proved for mutation-based search by Whitley. Furthermore, the work of Chakraborty and Janikow \cite{Chakraborty2003} indicates that the presence of fewer local optima for Gray representations does not necessarily result in improved performance. Shastri and Frachtenberg's analysis \cite{Shastri2020} has also cast further doubt upon the role of locality in explaining the difference in performance of standard versus Gray binary encoding. As such, the topic of locality remains unresolved pending further research, with high locality representations providing greater reliability on average. There is however, empirical evidence which suggests that Gray representations are superior for practical optimisation problems as demonstrated for a number of single objective test functions \cite{Hinterding1995, Mathias1994, Caruana1988} and for a MOOP \cite{Greiner2005}.
		
		Consequently, the primary advantage for Gray representations as opposed to standard binary representations is the removal of Hamming cliffs \cite{Schaffer1989}. Hamming cliffs refer to the large genetic distances which can occur between neighbours in the genotypic search space, which disrupt mutation and crossover operations. Gray representations eliminate Hamming cliffs as they are designed such that genotypic neighbours are separated by a Hamming distance equal to one. Alternative approaches, which modify the mutation operator, such as the introduction of a creep mutation operator to enable the evolutionary search to traverse Hamming cliffs \cite{Charbonneau1995}, are also viable. In fact, it is demonstrable that due to the isomorphism of fitness landscapes \cite{Reeves1999}, different metric-representation combinations can result in the same fitness landscape. Hence, the approach taken is to focus on the choice of an optimal representation for standard operators.
		
		Consequently, the use of Gray representations is recommended however, standard binary representations are sufficient and not necessarily inferior dependent on the nature of the optimisation problem. The use of unary representations is strongly discouraged. Synonymously redundant representations such as trivial voting mapping, although not included in the tool, may prove useful provided \textit{a priori} knowledge of sufficient optimal genotypic building blocks is available. However, the identification of optimal building blocks is a non-trivial task, especially for MOOPs whereby clear genetic patterns that contribute to optimality across \textbf{all} objectives may not even exist.
		
		The genetic representation for a \textit{local feature} therefore comprises of a gene containing multiple sequences of alleles, each representing a discretised component characteristic (Figure \ref{Fig: Local Genotype}). \textit{Zonal features} are therefore distinct from \textit{local features} due to the number of encoded states: whereas \textit{zonal features} can only be included or excluded, \textit{local features} can have numerous states of inclusion.
		\begin{figure}[hbt!]
			\centering
			\begin{tikzpicture}
				\draw[thick, fill=gray, opacity=0.5] (-7.0, 0) rectangle (-7.5, 0.75); 
				\draw[thick, fill=gray, opacity=0.5] (-6.5, 0) rectangle (-7.0, 0.75); 
				\draw[thick, fill=gray, opacity=0.5] (-6.0, 0) rectangle (-6.5, 0.75); 
				\draw[thick, fill=gray, opacity=0.5] (-5.5, 0) rectangle (-6.0, 0.75); 
				\draw[thick] (-5.0, 0) rectangle (-5.5, 0.75); 
				\draw[thick] (-4.5, 0) rectangle (-5.0, 0.75); 
				\draw[thick] (-4.0, 0) rectangle (-4.5, 0.75); 
				\draw[thick] (-3.5, 0) rectangle (-4.0, 0.75); 
				\draw[thick] (-3.0, 0) rectangle (-3.5, 0.75); 
				\draw[thick] (-2.5, 0) rectangle (-3.0, 0.75); 
				\draw[thick, fill=gray, opacity=0.5] (-2.0, 0) rectangle (-2.5, 0.75); 
				\draw[thick, fill=gray, opacity=0.5] (-1.5, 0) rectangle (-2.0, 0.75); 
				\draw[thick, fill=gray, opacity=0.5] (-1.0, 0) rectangle (-1.5, 0.75); 
				\draw[thick, fill=gray, opacity=0.5] (-0.5, 0) rectangle (-1.0, 0.75);
				\draw[thick] (-0.0, 0) rectangle (-0.5, 0.75);
				\draw[thick] (0.0, 0) rectangle (0.5, 0.75);
				\draw[thick] (0.5, 0) rectangle (1.0, 0.75); 
				\draw[thick] (1.0, 0) rectangle (1.5, 0.75); 
				\draw[thick] (1.5, 0) rectangle (2.0, 0.75); 
				\draw[thick] (2.0, 0) rectangle (2.5, 0.75); 
				\draw[thick] (2.5, 0) rectangle (3.0, 0.75);
				\draw[thick] (3.0, 0) rectangle (3.5, 0.75);
				\draw[thick] (3.5, 0) rectangle (4.0, 0.75);
				\node at (-7.25, 0.3725) {\large 0};
				\node at (-6.75, 0.3725) {\large 0};
				\node at (-6.25, 0.3725) {\large 1};
				\node at (-5.75, 0.3725) {\large 1};
				\node at (-5.25, 0.3725) {\large 0};
				\node at (-4.75, 0.3725) {\large 1};
				\node at (-4.25, 0.3725) {\large 0};
				\node at (-3.75, 0.3725) {\large 1};
				\node at (-3.25, 0.3725) {\large 1};
				\node at (-2.75, 0.3725) {\large 0};
				\node at (-2.25, 0.3725) {\large 1};
				\node at (-1.75, 0.3725) {\large 1};
				\node at (-1.25, 0.3725) {\large 0};
				\node at (-0.75, 0.3725) {\large 1};
				\node at (-0.25, 0.3725) {\large 0};
				\node at (0.25, 0.3725) {\large 0};
				\node at (0.75, 0.3725) {\large 1};
				\node at (1.25, 0.3725) {\large 0};
				\node at (1.75, 0.3725) {\large 1};
				\node at (2.25, 0.3725) {\large 0};
				\node at (2.75, 0.3725) {\large 0};
				\node at (3.25, 0.3725) {\large 1};
				\node at (3.75, 0.3725) {\large 1};
				\draw[decorate,decoration={brace,amplitude=10pt}] (4.0, 0) --  (-7.5, 0)
				node[midway,below=10pt] {\large Local Feature};
				\draw[decorate,decoration={brace,amplitude=10pt}] (-7.5, 0.75) --  (-5.5, 0.75)
				node[midway,above=10pt] {\small Characteristic 1};
				\draw[decorate,decoration={brace,amplitude=10pt}] (-5.5, 0.75) --  (-2.5, 0.75)
				node[midway,above=10pt] {\small Characteristic 2};
				\draw[decorate,decoration={brace,amplitude=10pt}] (-2.5, 0.75) --  (-0.5, 0.75)
				node[midway,above=10pt] {\small Characteristic 3};
				\draw[decorate,decoration={brace,amplitude=10pt}] (-0.5, 0.75) --  (4.0, 0.75)
				node[midway,above=10pt] {\small Characteristic 4};
			\end{tikzpicture}
			\caption{Example genetic representation of a local feature. The representation of the candidate solution for the local feature is composed of a concatenation of the bit-string representations for each of the binary encoded component characteristics under consideration.}
			\label{Fig: Local Genotype}
		\end{figure}
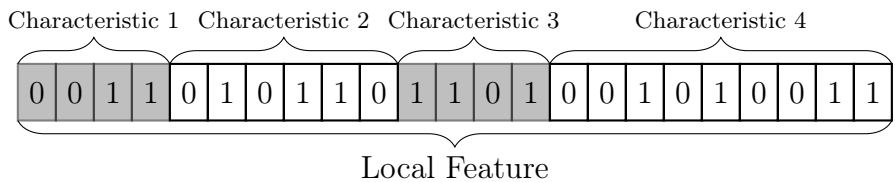
		
		An example is presented in Figure \ref{Fig: Local Feature}, demonstrating the representation of a detention basin as a local feature. Local features such as detention basins or water butts can be represented as zonal features if their component characteristics (e.g., size, location) are determined \textit{a priori}. However, this likely requires significant human input, which may be impractical and the quality of the optimised solutions can be limited by the initial zone designs. In cases where there are severe physical constraints limiting the design of such features, this may be a viable alternative. Alternatively, subsequent optimisations may enable the implementation of the optimised characteristics of a group of local features as a zonal feature when exploring in further detail interesting areas of the search space.
		\begin{figure}[hbt!]
			\centering
			\resizebox{0.8\textwidth}{!}{
			\begin{tikzpicture}
				
				\draw[thick, fill=gray, opacity=0.5] (-6.5, 0.5) rectangle (-6.0, 1.25);
				\node at (-6.25, 0.8725) {\large 1};
				\draw[thick, fill=white, opacity=0.5] (-6.0, 0.5) rectangle (-5.5, 1.25);
				\node at (-5.75, 0.8725) {\large 0};
				\draw[thick, fill=gray, opacity=0.5] (-5.5, 0.5) rectangle (-5.0, 1.25);
				\node at (-5.25, 0.8725) {\large 1};
				\draw[thick, fill=white, opacity=0.5] (-5.0, 0.5) rectangle (-4.5, 1.25);
				\node at (-4.75, 0.8725) {\large 0};
				\draw[thick, fill=gray, opacity=0.5] (-4.5, 0.5) rectangle (-4.0, 1.25);
				\node at (-4.25, 0.8725) {\large 1};
				\draw[thick, fill=gray, opacity=0.5] (-4.0, 0.5) rectangle (-3.5, 1.25);
				\node at (-3.75, 0.8725) {\large 1};
				\draw[decorate,decoration={brace,amplitude=10pt}] (-3.5, 0.5) --  (-6.5, 0.5)
				node[midway,below=10pt] {\large Genotype};
				\draw[decorate,decoration={brace,amplitude=10pt}] (-6.5, 1.25) --  (-5.0, 1.25)
				node[midway,above=10pt] {\normalsize Depth};
				\draw[decorate,decoration={brace,amplitude=10pt}] (-5.0, 1.25) --  (-3.5, 1.25)
				node[midway,above=10pt] {\normalsize Area};
				
				\draw[thick] 
				(-2,1.0) 
				.. controls (-2.0,1.0) and (-1.0,0.35) .. (0.0,0.5)
				.. controls (1.0,1.0) and (1.3,0.35) .. (1.5,0.5)
				.. controls (2.25,0.75) and (2.45,1.05) .. (2.1,1.5)
				.. controls (1.25,1.75) and (0.85,2.05) .. (0.5,1.7)
				.. controls (0.25,1.25) and (-0.85,1.95) .. (-1.0,2.0)
				.. controls (-2.25,2.25) and (-2.15,1.25) .. (-2.0,1.0);
				\draw[dashed] 
				(-2,1.0-1) 
				.. controls (-2.0,1.0-1) and (-1.0,0.35-1) .. (0.0,0.5-1)
				.. controls (1.0,1.0-1) and (1.3,0.35-1) .. (1.5,0.5-1)
				.. controls (2.25,0.75-1) and (2.45,1.05-1) .. (2.1,1.5-1)
				.. controls (1.25,1.75-1) and (0.85,2.05-1) .. (0.5,1.7-1)
				.. controls (0.25,1.25-1) and (-0.85,1.95-1) .. (-1.0,2.0-1)
				.. controls (-2.25,2.25-1) and (-2.15,1.25-1) .. (-2.0,1.0-1);
				\fill[gray!25]
				(-2,1.0) 
				.. controls (-2.0,1.0) and (-1.0,0.35) .. (0.0,0.5)
				.. controls (1.0,1.0) and (1.3,0.35) .. (1.5,0.5)
				.. controls (2.25,0.75) and (2.45,1.05) .. (2.1,1.5)
				.. controls (1.25,1.75) and (0.85,2.05) .. (0.5,1.7)
				.. controls (0.25,1.25) and (-0.85,1.95) .. (-1.0,2.0)
				.. controls (-2.25,2.25) and (-2.15,1.25) .. (-2.0,1.0);
				\draw[thick] (-2.08,0.2) -- (-2.08,1.2); 
				\draw[thick] (2.25,0) -- (2.25,1); 
				\draw[thick] (0,-0.5) -- (0,0.5); 
				\draw[thick] (1.4,-0.53) -- (1.4,0.47); 
				\draw[thick] (-2.5,0.2) -- (-2.5,1.2); 
				\draw[thick] (-2.25,1.2) -- (-2.75,1.2); 
				\draw[thick] (-2.25,0.2) -- (-2.75,0.2); 
				\node[left] at (-2.5, 0.7) {\normalsize $d$};
				\node[left] at (0.25, 1.1) {\normalsize $a$};
				\draw[decorate,decoration={brace,amplitude=10pt}] (2.5,-0.5) --  (-2.5,-0.5)
				node[midway,below=10pt] {\large Phenotype};
				\node at (0, 2.5) {\normalsize Detention Basin};
				
				\def\xshift{0.5}
				\node at (3.4+\xshift, 2.75) {\normalsize Bit-string};
				\node at (3.4+\xshift, 2.25) {\normalsize $000$};
				\node at (3.4+\xshift, 1.75) {\normalsize $001$};
				\node at (3.4+\xshift, 1.25) {\normalsize $011$};
				\node at (3.4+\xshift, 0.75) {\normalsize $010$};
				\node at (3.4+\xshift, 0.25) {\normalsize $110$};
				\node at (3.4+\xshift, -0.25) {\normalsize $111$};
				\node at (3.4+\xshift, -0.75) {\normalsize $101$};
				\node at (3.4+\xshift, -1.25) {\normalsize $100$};
				\node at (5.05+\xshift, 2.75) {\normalsize Depth};
				\node at (5.05+\xshift, 2.25) {\normalsize $d_0$};
				\node at (5.05+\xshift, 1.75) {\normalsize $d_0 + \epsilon_d$};
				\node at (5.05+\xshift, 1.25) {\normalsize $d_0 + 2\epsilon_d$};
				\node at (5.05+\xshift, 0.75) {\normalsize $d_0 + 3\epsilon_d$};
				\node at (5.05+\xshift, 0.25) {\normalsize $d_0 + 4\epsilon_d$};
				\node at (5.05+\xshift, -0.25) {\normalsize $d_0 + 5\epsilon_d$};
				\node at (5.05+\xshift, -0.75) {\normalsize $d_0 + 6\epsilon_d$};
				\node at (5.05+\xshift, -1.25) {\normalsize $d_{\textrm{max}}$};
				\node at (6.65+\xshift, 2.75) {\normalsize Area};
				\node at (6.7+\xshift, 2.25) {\normalsize $a_0$};
				\node at (6.7+\xshift, 1.75) {\normalsize $a_0 + \epsilon_a$};
				\node at (6.7+\xshift, 1.25) {\normalsize $a_0 + 2\epsilon_a$};
				\node at (6.7+\xshift, 0.75) {\normalsize $a_0 + 3\epsilon_a$};
				\node at (6.7+\xshift, 0.25) {\normalsize $a_0 + 4\epsilon_a$};
				\node at (6.7+\xshift, -0.25) {\normalsize $a_0 + 5\epsilon_a$};
				\node at (6.7+\xshift, -0.75) {\normalsize $a_0 + 6\epsilon_a$};
				\node at (6.7+\xshift, -1.25) {\normalsize $a_{\textrm{max}}$};
				\draw[thick] (2.5+\xshift,2.5) -- (7.5+\xshift,2.5);
				\draw[thick] (4.25+\xshift,3.0) -- (4.25+\xshift,-1.6);
				\draw[thick] (5.875+\xshift,3.0) -- (5.875+\xshift,-1.6);
				\draw[decorate,decoration={brace,amplitude=10pt}] (7.5+\xshift,-1.6) --  (2.5+\xshift,-1.6)
				node[midway,below=10pt] {\large Binary-Discrete Mapping};
			\end{tikzpicture}
			} 
			\caption{Genotype-discrete phenotype mapping for a \textit{local feature}. The genetic representation is a concatenation of the bit-strings used to encode each component characteristic (depth and area in the presented example). A suitable binary encoding (Gray coding in the provided example) can then be used to vary the characteristics between a minimum and maximum value as shown in the table (see also equations (\ref{eq: overall mapping}-\ref{eq: unary encoding})).}
			\label{Fig: Local Feature}
		\end{figure}
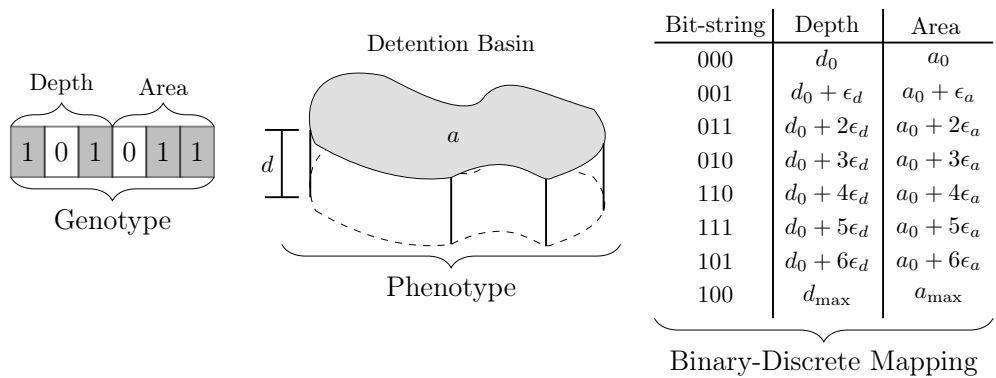
		
		\subsubsection{Objective Function}
		The role of the objective function is to provide a fitness evaluation of a candidate solution. In the context of the optimal placement and design of blue-green infrastructure, a fitness evaluation refers to an assessment of the simulated performance of a candidate solution against the chosen objectives, facilitated by analysing the results of a hydrodynamic flood model. As such, the proposed optimisation tool is designed to perform bi-objective optimisations of either the expected cost of damages due to flooding or the number of buildings exposed to \textit{high} flood risk versus the estimated whole life cost of implementation.
		
		Formally a MOOP seeks to simultaneously optimise, via minimisation in the presented case, a set of $m \geq 2$, likely competing, objective functions:
		\begin{equation}
			\min_{\boldsymbol{x}_p \in \Omega} \boldsymbol{F}(\boldsymbol{x}_p) = \left[f_1(\boldsymbol{x}_p), f_2(\boldsymbol{x}_p), \dots, f_m(\boldsymbol{x}_p) \right]^T
		\end{equation}
		where $\boldsymbol{x}_p = (x_1, \dots, x_n) \in \{0,1\}^{n_b} \times D^{n_d}$ is a decision vector in a feasible solution space, referred to as the decision space, $\Omega$, and $n = n_b + n_d$ is the number of binary and discrete decision variables. The set of attainable objective values is called the feasible objective set, $\mathcal{Z}$, defined as the image of the decision space under the objective functions: $\mathcal{Z} = \{\boldsymbol{F}(\boldsymbol{x}_p) \in \mathbb{R}^m : \boldsymbol{x}_p \in \Omega\}$. Here, each objective function, $f_i : \Omega \to \mathbb{R}$, maps the performance of a decision vector from the decision space to the objective space.
		
		Since the objective functions are required to evaluate measures of flood risk, accurate hydrodynamic modelling of the candidate solution is therefore crucial to the solution of the proposed MOOPs. As Costabile et al. \cite{Costabile2020} emphasise, `\textit{fully-dynamic modeling should be the unavoidable reference tool when the goal of urban flood mapping activity is not limited to the evaluation of the flood-prone areas extent but also involves the local estimation of flood hazard/vulnerability}'. Fully-dynamic modeling is therefore critical to facilitate an accurate fitness evaluation and consequently perform an effective optimisation procedure. Therefore, the minimum requirements for the chosen hydrodynamic model are that it be a spatially distributed, physically based, fully dynamic model, solving the classical shallow water equations in conservative form with the capacity to explicitly model the desired BGI features. The presented optimisation tool utilises CityCAT, a state-of-the-art hydrodynamic model that meets these criteria \cite{Glenis2018}; however, other suitable hydrodynamic models with comparable capabilities can also be used.
		
		Since a hydrodynamic simulation must be performed for each unique fitness evaluation (see Figure \ref{Fig: Flow Chart}), strategies for solving MOOPs involving expensive objective functions are of paramount importance. A popular solution is the use of metamodels or surrogate models to approximate the fitness evaluation process in a computationally efficient manner \cite{Chugh2019, DeWinter2024, Akhtar2016, Peri2006, Tenne2010}. However, for the proposed application, fully dynamic modelling represents the minimum complexity required to accurately analyse flood hazard and vulnerability, rendering metamodels unsuitable. Consequently, efforts must focus on designing evolutionary search strategies that efficiently approximate optimal solutions while minimising the number of unique fitness evaluations. 
		
		\begin{figure}[hbt!]
			\centering
			\resizebox{0.75\textwidth}{!}{
			\begin{tikzpicture}[node distance=1.6cm]
				
				\node (input1) [startstop] {\small Buildings Data};
				\node (input2) [startstop, below of=input1] {\small Topographic Data};
				\node (input3) [startstop, below of=input2] {\small Green Area Data};
				\node (input4) [startstop, below of=input3] {\small Rainfall Data};
				\node (input5) [startstop, below of=input4] {\small Genotype};
				
				\node (Decoder) [process, right of=input3, xshift=2cm, yshift=0cm] {\small Decoder};
				\node (CityCAT) [process, right of=Decoder, xshift=0.5cm, yshift=2cm] {\small Hydrodynamic \\ Simulation};
				\node (Exposure) [process, right of=CityCAT, xshift=2cm] {\small Exposure \\ Analysis};
				\node (Damages) [process, right of=Exposure, xshift=2cm] {\small Risk Metric\\ Calculation};
				\node (Cost) [process, right of=Decoder, xshift=0.5cm, yshift=-2cm] {\small Implementation Cost Calculation};
				
				\node (Fitness) [startstop, below of=Damages, yshift=-2.5cm] {\small Update \\ Fitness};
				
				\draw [arrow] (input1) -- (Decoder);
				\draw [arrow] (input2) -- (Decoder);
				\draw [arrow] (input3) -- (Decoder);
				\draw [arrow] (input4) -- (Decoder);
				\draw [arrow] (input5) -- (Decoder);
				\draw [arrow] (input5) -- (Decoder);
				\draw [arrow] (Decoder) -- (CityCAT);
				\draw [arrow] (Decoder) -- (Cost);
				\draw [arrow] (CityCAT) -- (Exposure);
				\draw [arrow] (Exposure) -- (Damages);
				\draw [arrow] (Damages) -- (Fitness);
				\draw [arrow] (Cost) -- (Fitness);
			\end{tikzpicture}
			}
			\caption{A workflow illustrating the logic used by the objective function when evaluating the chosen flood risk metric (number of \textit{high} risk buildings or the expected cost of damages).}
			\label{Fig: Flow Chart}
		\end{figure}
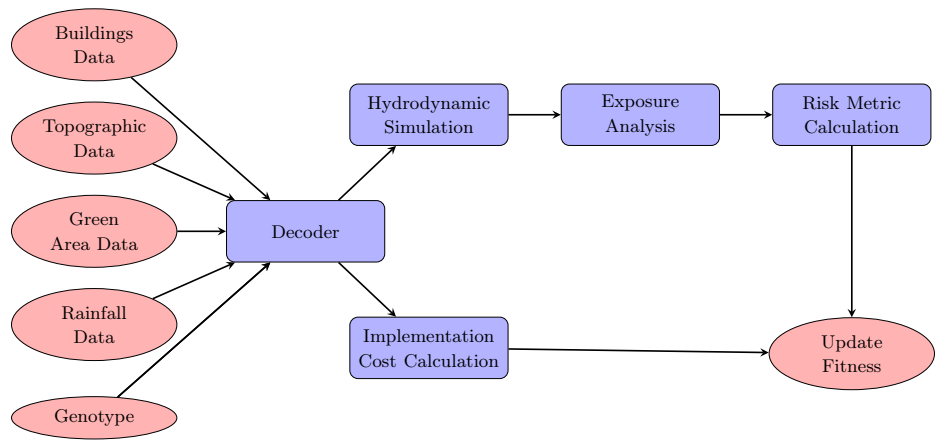
		
		Efficient solutions lie within the Pareto set, which comprises optimal decision vectors — each representing an infrastructure configuration that best balances objectives in the objective space. The optimal decision vectors belonging to the Pareto set are referred to as Pareto optimal and the image of the Pareto optimal solutions, via the objective functions, is called the Pareto front. Knowledge of the Pareto set allows decision-makers to assess trade-offs between objectives and make informed choices regarding flood risk mitigation and investment. The Pareto set, formally defined as follows, includes all non-dominated solutions within the objective space:
		\begin{definition}[Dominance Relation]\label{Def: Dominance}
			Let \( \boldsymbol{z}, \boldsymbol{w} \in \mathbb{R}^m \) be feasible objective vectors, \( \boldsymbol{z} \) dominates \( \boldsymbol{w} \) (denoted as \( \boldsymbol{z} \succ \boldsymbol{w} \)) if and only if:
			\begin{enumerate}[topsep=0pt, itemsep=0pt]
				\item \( \forall i \in \{1, \dots, m\} : z_{i} \leq w_{i} \)
				\item \( \exists j \in \{1, \dots, m\} : z_{j} < w_{j} \)
			\end{enumerate}
		\end{definition}
		\begin{definition}[Pareto Set]\label{Def: Pareto Set}
			Let \( \mathcal{Z} \subseteq \mathbb{R}^m \) be the feasible objective set. The Pareto set, \(\mathcal{P} \subseteq \mathcal{Z}\), consists of all vectors \(\boldsymbol{p} \in \mathcal{Z}\) that are not dominated by any vector \(\boldsymbol{z} \in \mathcal{Z}\):
			\begin{equation*}
				\mathcal{P} := \{\boldsymbol{p} \in \mathcal{Z} \mid \nexists \boldsymbol{z} \in \mathcal{Z} : \boldsymbol{z} \succ \boldsymbol{p}\}
			\end{equation*}
		\end{definition}
		\begin{figure}[hbt!]
			\centering
			\resizebox{0.55\textwidth}{!}{
			\begin{tikzpicture}
				\draw[->] (-0.25,0) -- (3.5,0) node[anchor=north] {$f_1$};
				\draw[->] (0,-0.25) -- (0,3.5) node[anchor=east] {$f_2$};
				
				\fill[pattern=north east lines, pattern color=gray!50] (0.5,3.5) .. controls (1.25,1.2) and (2.75,0.7) .. (3.5,0.6) -- (3.5,0) -- (0,0) -- (0,3.5) -- cycle;
				\node[below] at (0.8,0.8) {$R^m \setminus \mathcal{Z}$};
				
				\draw[dashed, red!50] (0.5,3.5) .. controls (1.25,1.2) and (2.75,0.7) .. (3.5,0.6);
				\foreach \x/\y [count=\n] in {0.5/3.5, 0.75/2.85, 1.25/2, 1.5/1.68, 2/1.24, 2.75/0.79, 3.5/0.6} {
					\node[red] at (\x,\y) {$\times$};
					\node[anchor=north east, red] at (\x,\y) {$p_{\n}$};
				}
				\draw[thick, red!75] (0.5,3.5) -- (0.75,3.5) -- (0.75,2.85) -- (1.25,2.85) -- (1.25,2) -- (1.5,2) -- (1.5,1.68) -- (2,1.68) -- (2,1.24) -- (2.75,1.24) -- (2.75,0.79) -- (3.5,0.79) -- (3.5,0.6);
				\draw[dashed, red!75] (0.5, 3.5) -- (3.5, 3.5) -- (3.5,0.6);
				\foreach \x/\y [count=\n] in {0.85/3.5, 1.3/3.3, 1.8/3.2, 2.5/3.4, 1.4/2.5, 1.2/2.9, 1.9/2.6, 2.2/2.7, 2.8/2.75, 3.0/2.5, 1.7/2.1, 1.8/2.4, 1.9/2.2, 2.4/2.15, 2.6/2.05, 3.1/2, 1.7/1.85, 2.1/1.6, 2.3/1.8, 2.4/1.55, 2.9/1.9, 3.3/1.75, 2.5/1.35, 2.9/1, 3/1.4, 3.2/1.3, 3.3/1.2, 3.5/1.2, 3.1/0.95, 3.35/0.85, 2.5/2.5, 2/2, 2.25/2.4, 1.75/3, 1.5/2.75, 1/3.2, 2.75/1.6, 2.25/3.25, 2.1/2.9, 2.4/3, 1.6/3.4} {
					\node[gray] at (\x,\y) {$\times$};
				}
				\node[below] at (2.75,2.75) {\large $\mathcal{Z}$};
				\node at (1.75, -0.6) {\small (a)};
				
				\draw[->] (5.75,0) -- (9.5,0) node[anchor=north] {$f_1$};
				\draw[->] (6,-0.25) -- (6,3.5) node[anchor=east] {$f_2$};
				\fill[pattern=north east lines, pattern color=gray!80, draw=gray!50] (7,1) -- (7,3.5) -- (9.5,3.5) -- (9.5,1) -- (7,1) -- cycle;
				\node[black] at (7,1) {$\times$};
				\node[anchor=north east, black] at (7,1) {$\mathbf{z}$};
				\node at (7.75, -0.6) {\small (b)};
			\end{tikzpicture}
			}
			\caption{(a) Illustrates a Pareto set, $\mathcal{P} = \{p_1, p_2, \dots, p_7\}$, whose points are denoted by an annotated red $\times$. The region dominated by the Pareto front is outlined in red and dominated solutions contained within the feasible objective space, $\mathcal{Z}$, are illustrated in grey. (b) illustrates the dominance relation defined in Definition \ref{Def: Dominance}, whereby, for a minimisation problem, $\mathbf{z}$ dominates the shaded region.}
			\label{Fig: Pareto}
		\end{figure}
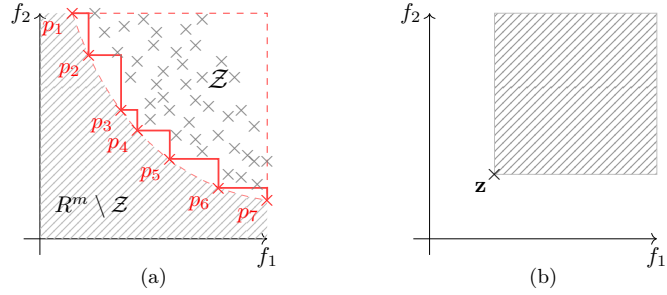
		
		\subsubsection{Exposure Analysis}\label{Section: Exposure Analysis}
		\begin{figure}[hbt!]
			\centering
			\includegraphics[width=0.65\textwidth]{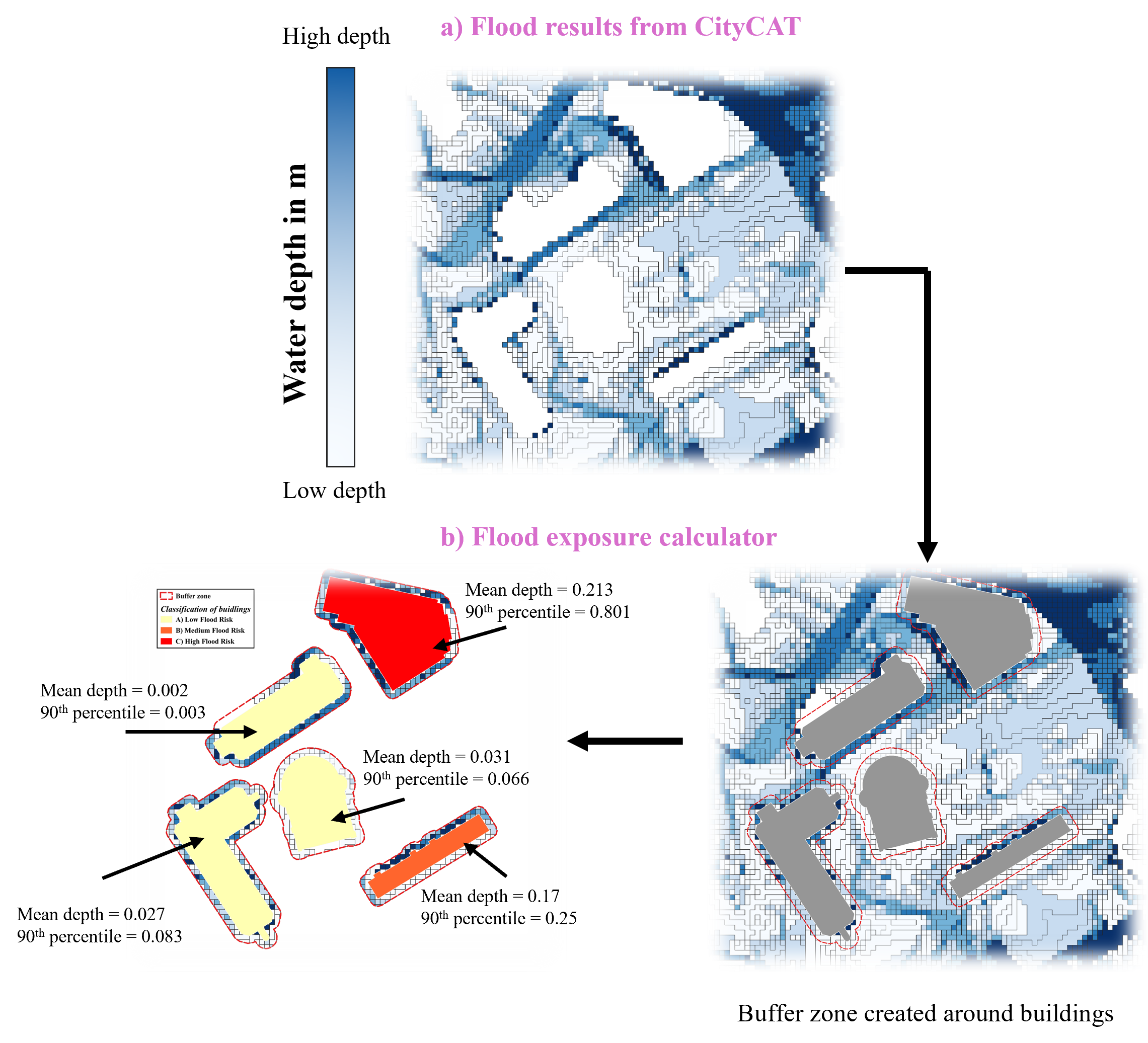}
			\caption{Schematic workflow of the flood exposure analysis tool for the classification of buildings according to the water depth in the buffer zone.}
			\label{Fig: Exposure Tool}
		\end{figure}
		
		The presented optimisation tool utilises a flood exposure calculator \cite{Bertsch2022} to quantify the flood exposure metric for each candidate solution. The tool is capable of estimating either the number of buildings exposed to \textit{high} flood risk or the expected cost of damages. These metrics are estimated by assessing the water depths within the immediate vicinity of each structure, specifically the flow depth within a buffer zone surrounding each building. The minimum buffer zone extent is a single computational cell, which can be extended to some multiple of a single computational cell. The buffer zone extent is therefore a function of the spatial resolution and the extent of the buffer zone should be adjusted accordingly. From experience, for high resolution spatial domains a buffer zone of a single cell is sufficient, for coarser meshes this may be extended to one-and-a-half cells.
			
		The water depths are then converted into a mean depth and a 90$^{\mathrm{th}}$ percentile depth. Buildings are then classified as being at \textit{high}, \textit{medium}, or \textit{low} risk (see Table \ref{Table: Classification}) based on whether the water depths exceed a threshold of 30cm (according to the UK standards) \cite{Bertsch2022}. Buildings deemed to be at low and medium flood risk are excluded from this analysis, as their potential flood-related damages are considered minor relative to those at higher risk levels. 
		
		\begin{table}[h!]
			\centering
			\begin{tabular}{ccc}
				\toprule
				\textbf{Exposure Class} & \textbf{Mean Depth (m)} & \textbf{90th Percentile (m)} \\ \midrule
				Low    & $<$ 0.10                     & $<$ 0.30                     \\
				Medium (I) & $<$0.10 & $\geq$ 0.30                    \\
				Medium (II) & $\geq$ 0.10 - $<$ 0.30     & $<$ 0.30                    \\
				High   & $\geq$ 0.10                  & $\geq$ 0.30                  \\ \bottomrule
			\end{tabular}
			\caption{The criteria for calculating flood exposure likelihood for buildings.}
			\label{Table: Classification}
		\end{table}
		
		Where the expected cost of damages is selected as the flood risk metric, the cost is estimated as a function of the flood risk and building classification (residential or commercial) following the method outlined in `\textit{Flood and Coastal Erosion Risk Management: A Manual for Economic Appraisal}' \cite{Priest2022}. This source provides standardised damage costs, offering a reliable basis for economic evaluation in flood risk assessments. Figure \ref{Fig: Flood Damages} illustrates some typical values for short duration storms without warning in the Northeast of England.

		\begin{figure}[hbt!]\centering
			\begin{subfigure}{.45\linewidth}
				\centering
				\includegraphics[width=\linewidth]{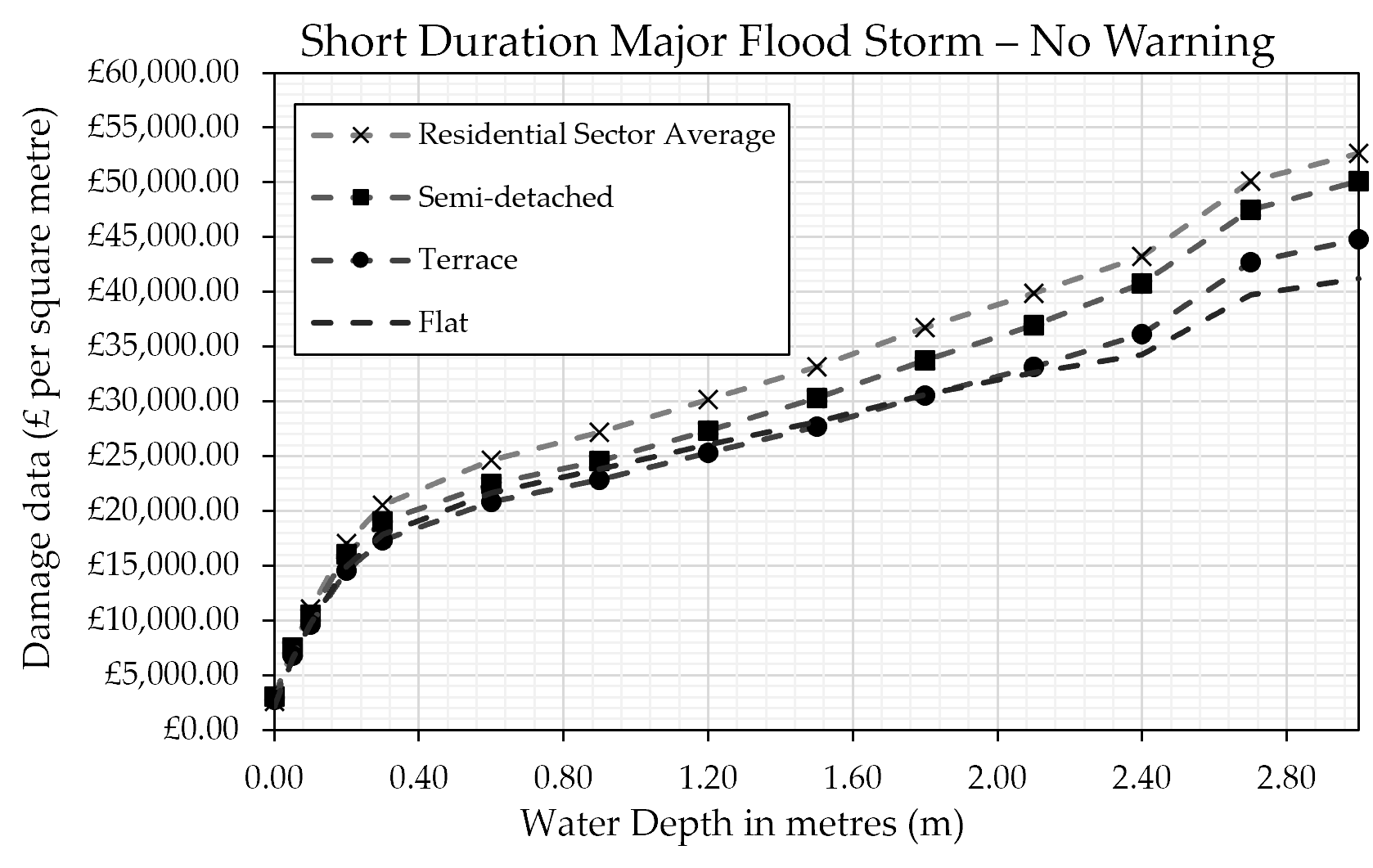}
				\caption{Residential Buildings}\label{Fig: Residential}
			\end{subfigure}
			\hspace{0.05\linewidth}
			\begin{subfigure}{.45\linewidth}
				\centering
				\includegraphics[width=\textwidth]{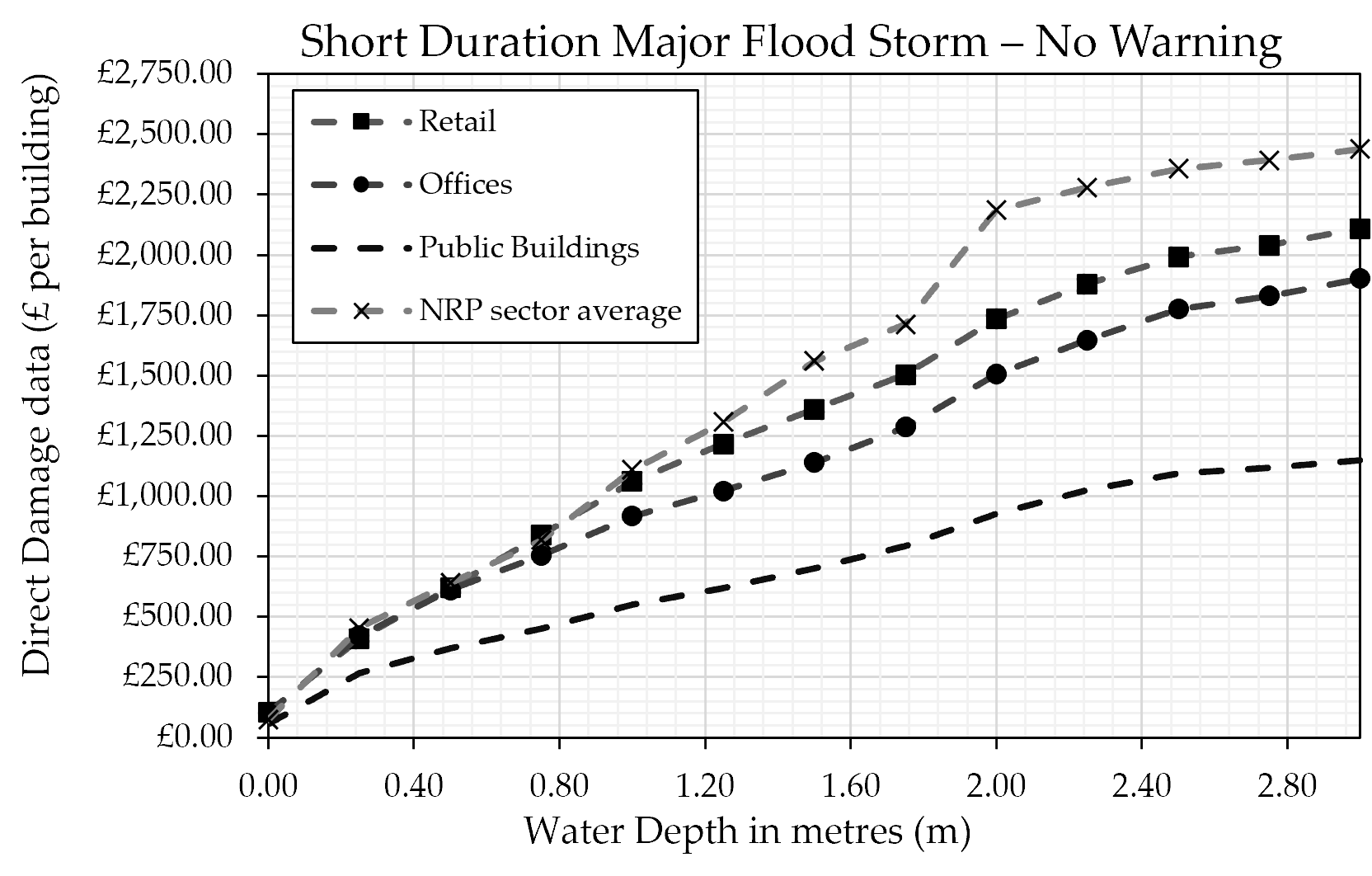}
				\caption{Commercial Buildings}\label{Fig: Commercial}
			\end{subfigure}
			\caption{Flood damages for different water depths in the North East of England for short duration storms without warning \cite{Priest2022}.}
			\label{Fig: Flood Damages}
		\end{figure}
		
		\subsubsection*{Hydrodynamic Model}
		As previously referenced, a fully-dynamic model is required where the aim extends beyond the simple mapping of flood inundation extents. Consequently, the optimisation tool utilises CityCAT which is a fully coupled 1D/2D hydrodynamic City Catchment Analysis Tool – CityCAT \cite{Glenis2018}. CityCAT is an advanced urban flood modelling tool capable of simulating both surface and pipe network flows. It explicitly represents buildings, surface and sub-surface drainage systems, as well as a variety of BGI features.
		
		For the representation of buildings, the model uses the ‘Building Hole’ approach (for additional approaches and validation see Iliadis et al. \cite{Iliadis2023b}). By excluding buildings from the computational grid and generating a no-flow boundary around them, the rainfall from roof of the building is redistributed to the nearest grid square, enabling the modelling of green roofing. This method improves the accuracy of flow path representation around buildings as well as reducing computational burden, improving the overall efficiency of the model.
		
		CityCAT generates time series data for water depth and flow velocity, along with flood maps and volume calculations for various infrastructure components such as manholes, gully drains, and buildings. The essential inputs for the model include the DTM, building footprints, permeable areas, and rainfall intensity data. The model is recognised as one of the most advanced and comprehensive hydrodynamic models available, making it highly effective for flood risk assessment and evaluating flood mitigation strategies within urban environments \cite{Guerreiro2017, Herreros2024, Iliadis2023a, Iliadis2023b, Jenkins2018, Kilsby2020, Kutija2014, Pregnolato2016, Vercruysse2019, Wilkinson2014, Bertsch2022}. For further details and validation of CityCAT refer to \cite{Glenis2018}.
		
		\subsection{Multi-Objective Optimisation Problems}
		Since for a MOOP the goal is to determine a diverse set of Pareto-optimal solutions, as opposed to a singular optimal solution for a single-objective problem, diversity preservation is a key consideration. In fact, the field of multi-objective optimisation using evolutionary algorithms (MOEAs) traces its origins to Schaffer's development of the vector-evaluated genetic algorithm (VEGA) in 1984 \cite{Schaffer1984} and the suggestion of Goldberg in 1989 \cite{Goldberg1989} to use non-dominated sorting (see definition \ref{Def: Dominance}), as opposed to sorting based on absolute objective scores, combined with diversity preservation tactics such as niching and speciation. 
		
		Early algorithms such as Fonseca and Fleming's multi-objective genetic algorithm (MOGA) \cite{Fonseca1993}, Srinivas and Deb's non-dominated sorting genetic algorithm (NSGA) \cite{Srinivas1994} and Horn et al.'s niched Pareto genetic algorithm (NPGA) \cite{Horn1994}, demonstrated that MOEAs based upon these principles could perform promisingly across a range of test problems. However, the algorithms crucially suffered from a strong dependence on appropriate parameter choices and a lack of elitism \cite{Eiben2015}. The lack of elite preservation strategies, corresponding to a failure to preserve the fittest members of the population, often led to the loss of good solutions making convergence to the Pareto set inconsistent.
		
		Consequently, the second generation of MOEAs developed in the 1990's and early 2000's focussed on the introduction of elitist strategies. Of this second generation of MOEAs, Deb et al.'s NSGA-II \cite{Deb2000} and the strength Pareto evolutionary algorithm (SPEA-2) \cite{Zitzler2001} emerged as prominent algorithms proficient at solving MOOPs. These algorithms incorporated crowding distance metrics and archive-based elitism, providing explicit diversity maintenance and reduced dependency on parameters, addressing the primary limitations of the first generation MOEAs. Despite being introduced over two decades ago, the NSGA-II and, to a lesser extent, the SPEA-2 maintain relevancy (see for example \cite{Soleimani2022, Rafati2023, Lin2021, Wu2022, Wang2022a, Wang2022b, Bailey2023, Kumardey2022, Cheng2023}) and consequently, the NSGA-II and SPEA-2 are used to benchmark the performance of the optimisation tool. The main loops for the NSGA-II and SPEA-2 are summarised in Algorithm \ref{Alg: NSGA-II} and Algorithm \ref{Alg: SPEA-2} respectively.
		
		\begin{algorithm}[H]\label{Alg: SPEA-2}
			\caption{SPEA-2 main loop, see \cite{Zitzler2001} for further details.}
			\KwIn{Population size $N$, archive size $\bar{N}$, length of binary representation $n$, number of objectives $m$, objective functions $\mathbf{F} = \left[f_1, \dots, f_m \right]^T$, and $T_{\mathrm{end}}$ (maximum number of generations).}
			\KwOut{Updated archive population $\bar{P}_{t+1}$.}
			
			\SetKw{KwSet}{Set}
			\SetKwComment{Comment}{$\triangleright$\ }{} 
			
			$P_0 \gets \{\mathbf{x}_1, \dots, \mathbf{x}_N \} \, , \, \mathbf{x}_i \in \{0,1\}^n \, , \, i \in \{1, \dots, N\}$ \, \Comment{Random initial population}
			$\bar{P}_0 \gets \emptyset \, , \, t \gets 0$ \, \Comment{Initialise archive population and generation counter}
			
			\While{$t < T_{\mathrm{end}}$}{
				$\mathcal{P} \gets P_t \cup \bar{P}_t$ \, \Comment{Combined population}
				
				$\forall \mathbf{x}_i \in \mathcal{P} : F_i \gets \mathbf{F}(\mathbf{x_i})$ \, \Comment{Evaluate objective function}
								
				$\forall \mathbf{x}_i \in \mathcal{P} : S_i \gets \left\vert \{\mathbf{x}_j \in P_t \cup \bar{P}_t \mid \mathbf{x}_i \succ \mathbf{x}_j\} \right\vert$ \, \Comment{Strength: count of dominated solutions}
				
				$\forall \mathbf{x}_i \in \mathcal{P} : R_i \gets \sum_{\mathbf{x}_j \in \mathcal{P}, \, \mathbf{x}_j \succ \mathbf{x}_i} S_j$ \, \Comment{Raw fitness: sum of strength of dominators}
				
				$\forall \mathbf{x}_i \in \mathcal{P} : \mathcal{D}_i \gets \left\{ {\sum_{k=1}^{m}\frac{\left|f_k(\mathbf{x}_i) - f_k(\mathbf{x}_j)\right|}{f_k^{\mathrm{max}} - f_k^{\mathrm{min}}}} \, , \, \mathbf{x}_j \in \mathcal{P} \setminus \mathbf{x}_i \right\}$ \, \Comment{$\mathcal{D}_i$ is the set of all normalised distances from $\mathbf{x}_i$ to all other $\mathbf{x}_j$ in the objective space}
				
				$\forall \mathbf{x}_i \in \mathcal{P} : D_i \gets \frac{1}{\sigma_i^k + 2} \; \mathrm{where} \; k = \lfloor \sqrt{N+\bar{N}} \rfloor \, , \, \sigma_i^k = \min \left\{ d \in \mathcal{D}_i \;\bigg\vert\; \Big\vert \{ \delta \in \mathcal{D}_i : \delta \le d \} \Big\vert \ge k \right\}$ \, \Comment{Crowding distance: $\sigma_i^k$ is the $k$-th smallest element of $D_i$}
				
				$\forall \mathbf{x}_i \in \mathcal{P} : \mathbb{F}_i \gets R_i + D_i$ \, \Comment{Combined fitness}
				
				$\bar{P}_{t+1} \gets \{\mathbf{x}_i \in \mathcal{P} \mid \mathbb{F}_i < 1 \}$ \, \Comment{Add non-dominated solutions to archive}
				
				\While{$|\bar{P}_{t+1}| < \bar{N}$}{
					$\bar{P}_{t+1} \gets \bar{P}_{t+1} \cup \Big\{ \arg\min_{\mathbf{x}_i \in \mathcal{P} \setminus \bar{P}_{t+1}} \mathbb{F}_i \Big\}$ \, \Comment{Fill with the best dominated solutions}
				}
				
				\While{$|\bar{P}_{t+1}| > \bar{N}$}{
					$\bar{P}_{t+1} \gets \bar{P}_{t+1} \setminus \Big\{ \mathbf{x}_i \in \bar{P}_{t+1} \mid \forall \mathbf{x}_j \in \bar{P}_{t+1} \setminus \mathbf{x}_i: \left( \forall 0 < k < |\bar{P}_{t+1}| : \sigma_i^k = \sigma_j^k \right) \, \vee \, \exists \, 0 < k < |\bar{P}_{t+1}| : \Big[ \left( \forall 0 < l < k : \sigma_i^l = \sigma_j^l \right) \land \sigma_i^k < \sigma_j^k \Big] \Big\}$ \, \Comment{Truncate archive iteratively by removing the individual with minimum distance to their nearest neighbour, comparing distances up to the $k$-th nearest neighbour to break ties}
				}
				
				\If{$t < T_{\mathrm{end}}$}{
					$P_{t+1} \gets \mathrm{Variation}\left(\mathrm{BinaryTournament}(\bar{P}_{t+1}, N)\right)$ \, \Comment{Apply binary tournament selection with replacement, crossover and mutation}
				}
				
				$t \gets t + 1$ \, \Comment{Increment generation counter}
			}
			\Return $\bar{P}_{t+1}$\;
		\end{algorithm}
		
		\begin{algorithm}[H]\label{Alg: NSGA-II}
			\caption{NSGA-II main loop, see \cite{Deb2000} for further details.}
			\KwIn{Population size $N$, length of binary representation $n$, number of objectives $m$, objective functions $\mathbf{F} = \left[f_1, \dots, f_m \right]^T$ and a maximum number of generations $T_{\textrm{end}}$.}
			\KwOut{Updated population $P_{t+1}$.}
			
			\SetKw{KwSet}{Set}
			\SetKwComment{Comment}{$\triangleright$\ }{} 
			
			$P_0 \gets \{\mathbf{x}_1, \dots, \mathbf{x}_N \} \, , \, \mathbf{x}_i \in \{0,1\}^n \, , \, i \in \{1, \dots, N\}$ \, \Comment{Random initial population}
			$Q_0 \gets \emptyset \, , \, t \gets 0$ \, \Comment{Initialise offspring population and generation counter}
			
			\While{$t < T_{\mathrm{end}}$}{
				$\mathcal{F} \gets \mathrm{FastNonDominatedSort}(P_t \cup Q_t)$ \, \Comment{Use a fast non-dominated sort \cite{Deb2000} to divide into fronts of equal dominance}
				$P_{t+1} \gets \emptyset \, , \, i \gets 0$ \, \Comment{Initialise next generation and front counter}
				\While{$|P_{t+1}| + |F_i| \leq N$}{
					$\mathrm{CrowdingDistance}(F_i)$\;
					$P_{t+1} \gets P_{t+1} \cup F_i$ \, \Comment{Add the whole front to the next generation}
					$i \gets i + 1$ \, \Comment{Move to next front}
				}
					
				\If{$|P_{t+1}|<N$}{
					$\mathrm{CrowdingDistance}(F_i)$\;
					\While{$|P_{t+1}| < N$}{
						$P_{t+1} \gets P_{t+1} \cup \{ \argmax_{\mathbf{x}_i \in F_i} d[\mathbf{x}_i] \}$ \, \Comment{Add least crowded front member}
						$F_i \gets F_i \setminus \mathbf{x}_i$\;
					}
				}
				$Q_{t+1} \gets \mathrm{Variation}\left(\mathrm{BinaryTournament}({P}_{t+1}, N)\right)$ \, \Comment{Create offspring}
				$t \gets t + 1$ \, \Comment{Increment generation counter}
			}
			\Return $P_{t+1}$\;
			
   			\SetKwProg{Fn}{Function}{:}{}
			\Fn{CrowdingDistance($I$)}{
				$\forall \mathbf{x} \in I: d[\mathbf{x}] \gets 0$ \;
				\For{$k \gets 1$ \KwTo $m$}{
					$S \gets \mathrm{sort}(I, f_k)$ \, \Comment{Sort members of $I$ by objective $k$}
					$d[S_1] \gets d[S_{|I|}] \gets \infty$ \, \Comment{Boundary distances are infinite}
					\For{$j \gets 2$ \KwTo $|I|-1$}{
						$d[S_j] \gets d[S_j] + \frac{f_k(S_{j+1}) - f_k(S_{j-1})}{f_k^{\max} - f_k^{\min}}$ \, \Comment{Accumulate normalised distance}
					}
				}
			}
		\end{algorithm}
		
		\subsection{$\epsilon$-MOEA}
		The proposed optimisation tool utilises a MOEA, hereafter referred to as a $\epsilon$-MOEA due to its use of $\epsilon-$dominance concepts, which are explained in detail in Section \ref{Section: Epsilon Dominance}. The proposed $\epsilon-$MOEA incorporates several features aimed at enhancing usability and improving evolutionary search performance relative to the benchmark second generation algorithms previously discussed. Specifically, these features include the use of $\epsilon-$dominance concepts, adaptive population sizing, seeding of the initial archive population and a self-termination criterion. In the following subsections, the key components of the proposed $\epsilon-$MOEA, as referenced in Algorithm \ref{Alg: epsilon-MOEA}, are outlined.
		
		\begin{algorithm}[H]\label{Alg: epsilon-MOEA}
			\caption{$\epsilon-$MOEA main loop.}
			\KwIn{Maximum archive size $\bar{N}$, length of binary representation $n$, number of objectives $m$, objective functions $\mathbf{F} = \left[f_1, \dots, f_m \right]^T$, a convergence threshold $\Delta p$, a lag window of $K_w$ generations and $T_{\textrm{end}}$ (maximum number of generations).}
			\KwOut{Updated archive population $\bar{P}_{t+1}$.}
			
			\SetKw{KwSet}{Set}
			\SetKwComment{Comment}{$\triangleright$\ }{} 
			
			$t \gets 0$ \, \Comment{Initialise generation counter}
			$P_0 \gets \{\mathbf{x}_1, \dots, \mathbf{x}_{5} \} \, , \, \mathbf{x}_i \in \{0,1\}^n \, , \, i \in \{1, \dots, 5\}$ \, \Comment{Random initial population of size 5}
			$\bar{P}_0 \gets \{\boldsymbol{x}_{p, \min}, \boldsymbol{x}_{p,\max} \}$ \, \Comment{Initialise archive population using extreme decision vectors}
			$\epsilon \gets [\epsilon_1, \epsilon_2]^T$ \, \Comment{Calculate $\epsilon$ for each objective using equation (\ref{Eq: epsilon})}
			update $\bar{P}_0$ \, \Comment{Update the archive, using algorithm \ref{Alg: Update}}
			\While{Termination condition is False}{
				update $P_t$ \, \Comment{Generate offspring via selection, recombination, and mutation}
				update $\bar{P}_t$ \, \Comment{Update the archive using Algorithm \ref{Alg: Update}}
				
				recalculate $\epsilon$ and prune $\bar{P}_t$ if necessary \, \Comment{Dynamic grid adaptation, see Section \ref{Section: Adaptive Epsilon}}
				
				\If{$K_w$ generations have passed since last check}{
					check inter-run criterion \, \Comment{Self-termination criterion, see Section \ref{Section: Self-Termination}}
				}
				
				\If{Termination condition is False}{
					check intra-run criterion \, \Comment{Adaptive population sizing using Algorithm \ref{Alg: Intra-Run}}
				}
				
				$t \gets t + 1$ \, \Comment{Update generation counter}
			}
			
			\Return $\bar{P}_{t+1}$\;
		\end{algorithm}
		
		\subsubsection{$\epsilon-$Dominance}\label{Section: Epsilon Dominance}
		Using the standard dominance relation, as defined in definition \ref{Def: Dominance}, the resulting Pareto set, as defined in definition \ref{Def: Pareto Set}, may be intractable for most realistic problems. Furthermore, even if determination of the full Pareto set proved tractable, the density of the Pareto-optimal points may be overwhelming and impractical for a decision maker regardless. The concept of $\epsilon-$dominance enables for a more manageable approximation of the Pareto set to be sought. Epsilon dominance, using the additive approximation for a minimisation problem, is defined as \cite{Laumanns2002}:
		\begin{definition}[\(\epsilon\)-Dominance]
			Let \(\mathbf{z},\mathbf{w} \in {\mathbb{R}^{+}}^m\) be feasible objective vectors, \(\mathbf{z}\) is then said to \(\epsilon\)-dominate \(\mathbf{w}\) for some \(\epsilon > 0\), denoted as \(\mathbf{z} \succ_{\epsilon} \mathbf{w}\), if and only if for all \(i \in \{1,\dots,m\}\): 
			\begin{equation}
				z_i - \epsilon_i \leq w_i
			\end{equation}
		\end{definition}
		An $\epsilon-$approximate Pareto set, denoted with a superscript $^*$ to indicate approximation, can therefore be defined based upon the $\epsilon-$dominance relation as \cite{Laumanns2002}:
		\begin{definition}[\(\epsilon\)-Approximate Pareto Set]\label{def: eps-approx set}
			Let \( \mathcal{Z} \subseteq {\mathbb{R}^{+}}^m \) be the feasible objective set and \(\epsilon > 0\). Then a set \(\mathcal{P}^*_{\epsilon}\) is called an \(\epsilon\)-approximate Pareto set of \(\mathcal{Z}\), if any vector \(\mathbf{z} \in \mathcal{Z}\) is \(\epsilon\)-dominated by at least one vector \(p^* \in \mathcal{P}^*_{\epsilon}\): 
			\begin{equation}
				\forall \mathbf{z} \in \mathcal{Z} : \exists p^* \in \mathcal{P}^*_{\epsilon} \, \textrm{ such that } \, \mathbf{p}^* \succ_{\epsilon} \mathbf{z}
			\end{equation}
			
			The set of all \(\epsilon\)-approximate Pareto sets of $\mathcal{Z}$ is denoted as \(\mathcal{P}^*_{\epsilon}(\mathcal{Z})\).
		\end{definition}
		Refinement of the concept of an $\epsilon$-approximate Pareto set, as detailed by Laumanns et al. \cite{Laumanns2002}, results in the concept of a $\epsilon$-Pareto set:
		\begin{definition}[\(\epsilon\)-Pareto Set]
			Let \( \mathcal{Z} \subseteq {\mathbb{R}^{+}}^m \) be the feasible objective set and \(\epsilon > 0\). Then a set \(\mathcal{P}_{\epsilon} \subseteq \mathcal{Z}\) is called an \(\epsilon\)-Pareto set of $\mathcal{Z}$ if:
			\begin{enumerate}
				\item \(\mathcal{P}_{\epsilon}\) is an \(\epsilon\)-approximate Pareto set of \(\mathcal{Z}\): \(\mathcal{P}_{\epsilon} \in \mathcal{P}^*_{\epsilon}(\mathcal{Z})\)
				\item \(\mathcal{P}_{\epsilon}\) contains Pareto points of \(\mathcal{Z}\) only: \(\mathcal{P}_{\epsilon} \subseteq \mathcal{P}\)
			\end{enumerate}
			Then the set of all \(\epsilon\)-Pareto sets of \(\mathcal{Z}\) is denoted as \(P_{\epsilon}(\mathcal{Z})\).
		\end{definition}
		An $\epsilon$-Pareto set, $\mathcal{P}_{\epsilon}$, therefore not only $\epsilon$-dominates all vectors in $\mathcal{Z}$, but also only consists of Pareto-optimal vectors of $\mathcal{Z}$, and therefore $\mathcal{P}^*_{\epsilon}(\mathcal{Z}) \subseteq \mathcal{P}_{\epsilon}(\mathcal{Z})$. The $\epsilon$-approximate Pareto set therefore offers a practical and attractive solution for decision makers as it allows for a simplified set of solutions that still captures the essential trade-offs among objectives. Whilst an $\epsilon$-Pareto set presents a more refined approximation, containing only Pareto optimal solutions, an $\epsilon-$approximate Pareto set is easier to compute due to its more relaxed inclusion criteria. Consequently, $\epsilon-$approximate Pareto solutions are appealing in real-world applications whereby tractability and efficiency may trump strict optimality concerns and particularly for expensive objective functions where minimising the number of fitness evaluations is a high priority.
		
		The parameter $\epsilon$ represents a permissible tolerance in objective values, and its choice depends on the specific needs of the application. In ideal circumstances, problem domain knowledge will be incorporated by a decision-maker to select an appropriate value of $\epsilon$ based on the desired granularity of the approximation of the Pareto set.
		
		\subsubsection{Environmental Selection}
		Following the recommendations of Laumanns et al. \cite{Laumanns2002}, $\epsilon-$dominance concepts are incorporated within the environmental selection process. The environmental selection process refers to the selection of individuals from a population for survival into the next generation. Effective environmental selection is key to ensure that the evolutionary search converges to a representative and diverse set of solutions.
		
		$\epsilon-$dominance is incorporated into the environmental selection process at a coarse level via the use of a \textit{box} function to divide the objective space into boxes (hypercubes for $m \geq 3$). At a fine level, environmental selection is further enforced by constraining the archive to a maximum of one element per box (hypercube). The \textit{box} function used to implement an additive $\epsilon$ is included in Algorithm \ref{Alg: Box Function}.
		
		\begin{algorithm}[H]\label{Alg: Box Function}
			\caption{A $\mathrm{box}$ function used to map the feasible objective vectors into discrete boxes (hypercubes) in the objective space.}
			\KwIn{A feasible objective vector $\mathbf{z} = \mathbf{F}(\mathbf{x}) = [z_1,\dots,z_m]^T$, $m \geq 2$ objectives, $\epsilon > 0$ and a lower bound of the fitness for each objective $\mathbf{z}^{\mathrm{min}} = [z_1^{\mathrm{min}}, \dots, z_m^{\mathrm{min}}]^T = \arg\min_{\boldsymbol{x}_p \in \Omega} \mathbf{F}(\boldsymbol{x}_p) $.}
			\KwOut{A $\mathrm{box}$ index vector $\mathbf{b} \in \mathbb{Z}^m$, where each $b_i \in \mathbf{b}$ represents the index of the box that $\mathbf{z}$ lies in for the $i$-th objective.}
			
			\SetKw{KwSet}{Set}
			\SetKwComment{Comment}{$\triangleright$\ }{} 
			
			\ForAll{$i \in \{1,\dots,m\}$}{
				$b_i \gets \lfloor \frac{z_i - z_{i}^{\mathrm{min}}}{\epsilon_i} \rfloor$
			}
			\Return $\mathbf{b}$\; 
		\end{algorithm}
		
		This leads to the generalised concept of box-dominance, which is defined as follows \cite{Laumanns2002}:
		\begin{definition}[Box Dominance]\label{Def: Box Dominance}
			Let \(\mathbf{z},\mathbf{w} \in {\mathbb{R}^{+}}^m \) be feasible objective vectors, with \textit{box} index vectors \(\mathbf{b}(\mathbf{z})\) and \(\mathbf{b}(\mathbf{w})\) respectively. \(\mathbf{z}\) is said to \(\textit{box}\)-dominate \(\mathbf{w}\) for some \(\epsilon > 0\), denoted as \(\mathbf{z} \succ_{\textit{box}} \mathbf{w}\), if and only if for all \(i \in \{1,\dots,m\}\): 
			\begin{enumerate}
				\item \( \forall i \in \{1,\dots, m\} : b_i(\mathbf{z}) \leq b_i(\mathbf{w}) \)
				\item \( \exists j \in \{1,\dots, m\} : b_j(\mathbf{z}) < b_j(\mathbf{w}) \)
			\end{enumerate}
		\end{definition}
		The generalised dominance relation ensures that the $\epsilon-$MOEA maintains a set of non-dominated boxes within the feasible objective space, ensuring an $\epsilon$-approximation of the Pareto set.
		\begin{figure}[hbt!]
			\centering
			\resizebox{0.55\textwidth}{!}{
			\begin{tikzpicture}
				\draw[->] (-0.25,0) -- (3.5,0) node[anchor=north] {$f_1$};
				\draw[->] (0,-0.25) -- (0,3.5) node[anchor=east] {$f_2$};
				
				\newcommand{\val}{0.7cm}
				\foreach \xy in {\val, 2*\val, 3*\val, 4*\val} {
					\draw[dashed] (0, \xy) -- (3.5, \xy); 
					\draw[dashed] (\xy, 0) -- (\xy, 3.5); 
				}
				
				\fill[pattern=north east lines, pattern color=gray!50] (0.5,3.5) .. controls (1.25,1.2) and (2.75,0.7) .. (3.5,0.6) -- (3.5,0) -- (0,0) -- (0,3.5) -- cycle;
				\node[below] at (0.8,0.8) {$R^m \setminus \mathcal{Z}$};
				\draw[<->] (0,-0.125) -- (0.7,-0.125);
				\node at (0.35,-0.25) {$\epsilon_1$};
				\draw[<->] (-0.125,0) -- (-0.125,0.7);
				\node at (-0.28, 0.35) {$\epsilon_2$};
				\draw[dashed, red!50] (0.5,3.5) .. controls (1.25,1.2) and (2.75,0.7) .. (3.5,0.6);
				\node[red] at (0.5, 3.5) {$\times$}; 
				\node[anchor=north east, red] at (0.5, 3.5) {$p_{1}$}; 
				\node[gray] at (0.75, 2.85) {$\times$}; 
				\node[anchor=north east, gray] at (0.75, 2.85) {$p_{2}$}; 
				\node[red] at (1.25, 2.00) {$\times$}; 
				\node[anchor=north east, red] at (1.25, 2.00) {$p_{3}$}; 
				\node[gray] at (1.50, 1.68) {$\times$}; 
				\node[anchor=north east, gray] at (1.50, 1.68) {$p_{4}$}; 
				\node[red] at (2.00, 1.24) {$\times$}; 
				\node[anchor=north east, red] at (2.00, 1.24) {$p_{5}$}; 
				\node[gray] at (2.75, 0.79) {$\times$}; 
				\node[anchor=north east, gray] at (2.75, 0.79) {$p_{6}$}; 
				\node[red] at (3.50, 0.60) {$\times$}; 
				\node[anchor=north east, red] at (3.50, 0.60) {$p_{7}$}; 
				\draw[thick, red!75] (0.7,3.5) -- (0.7,3.5-1.4) -- (1.4,3.5-1.4) -- (1.4,3.5-2.1) -- (2.1,3.5-2.1) -- (2.1,3.5-2.8) -- (3.5,3.5-2.8);
				\draw[dashed, red!75] (0.7,3.5) -- (3.5,3.5) -- (3.5,0.7);
				\foreach \x/\y [count=\n] in {0.85/3.5, 1.3/3.3, 1.8/3.2, 2.5/3.4, 1.4/2.5, 1.2/2.9, 1.9/2.6, 2.2/2.7, 2.8/2.75, 3.0/2.5, 1.7/2.1, 1.8/2.4, 1.9/2.2, 2.4/2.15, 2.6/2.05, 3.1/2, 1.7/1.85, 2.1/1.6, 2.3/1.8, 2.4/1.55, 2.9/1.9, 3.3/1.75, 2.5/1.35, 2.9/1, 3/1.4, 3.2/1.3, 3.3/1.2, 3.5/1.2, 3.1/0.95, 3.35/0.85, 2.5/2.5, 2/2, 2.25/2.4, 1.75/3, 1.5/2.75, 1/3.2, 2.75/1.6, 2.25/3.25, 2.1/2.9, 2.4/3, 1.6/3.4} {
					\node[gray] at (\x,\y) {$\times$};
				}
				\node[below] at (2.75,2.75) {\large $\mathcal{Z}$};
				\node at (1.75, -0.6) {\small (a)};
				
				\draw[->] (5.75,0) -- (9.5,0) node[anchor=north] {$f_1$};
				\draw[->] (6,-0.25) -- (6,3.5) node[anchor=east] {$f_2$};
				\draw[<->] (6,-0.125) -- (6.5,-0.125);
				\node at (6+0.25,-0.25) {$\epsilon_1$};
				\draw[<->] (6-0.125,0) -- (6-0.125,0.5);
				\node at (6-0.28, 0.25) {$\epsilon_2$};
				\def\grid{0.5}
				\foreach \num in {1, 2, 3, 4, 5, 6, 7} {
					\draw[dashed] (6, {\num*\grid}) -- (9.5, {\num*\grid}); 
					
					\draw[dashed] ({6+\num*\grid}, 0) -- ({6+\num*\grid}, 3.5); 
				}
				\fill[pattern=north east lines, pattern color=gray!80, draw=gray!50] (7.5,1) -- (7.5,1.5) -- (7,1.5) -- (7,3.5) -- (9.5,3.5) -- (9.5,1) -- (7.5,1) -- cycle;
				\node[black] at (7.3,1.25) {$\times$};
				\node[anchor=north east, black] at (7.3,1.25) {$\mathbf{z}$};
				\node at (7.75, -0.6) {\small (b)};
			\end{tikzpicture}
			}
			\caption{(a) Illustrates a $\epsilon-$Pareto set, $\mathcal{P}_{\epsilon} = \{p_1, p_3, p_5, p_7\}$, whose points are denoted by an annotated red crosses ($\times$). Note that $\mathcal{P}_{\epsilon} \subset \mathcal{P}$, as the Pareto optimal points $\{p_2, p_4, p_6\}$ are box-dominated. The box-dominated region is outlined in red and box dominated solutions contained within the feasible objective space, $\mathcal{Z}$, are denoted by grey crosses ($\times$). (b) illustrates the box dominance relation defined in Definition \ref{Def: Box Dominance}, whereby, for a minimisation problem, $\mathbf{z}$ box-dominates the shaded region.}
			\label{Fig: Hypercubes}
		\end{figure}
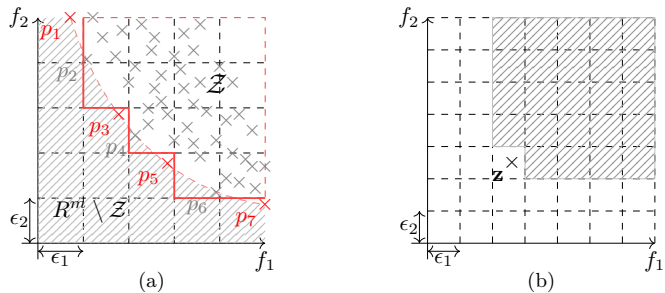
		
		Papadimitriou and Yannakakis \cite{Papadimitriou2000} and Erlebach et al. \cite{Erlebach2001} demonstrated that an approximate Pareto set of polynomial size in relation to the encoded input always exists. This was achieved through analysis of a multiplicative approximation, resulting in a hyper-grid with co-ordinates $1, (1+\epsilon_i), (1+\epsilon_i)^2, \dots$, whereby each hypercube contains at most one non-dominated feasible objective vector. The box dominance concept as per definition \ref{Def: Box Dominance} thus produces a bounded $\epsilon$-approximate Pareto set \cite{Laumanns2002}:
		\begin{equation}\label{Eq: Bound}
			|\mathcal{P}_{\epsilon}| \leq \prod_{j=1}^{m-1}\frac{f^{\mathrm{max}}_i - f^{\mathrm{min}}_i}{\epsilon_i}
		\end{equation}
		where $f^{\mathrm{max}}_i$ and $f^{\mathrm{min}}_i$ are the upper and lower bounds of the $i^{\mathrm{th}}$ objective for all $i \in \{1,\dots, m\}$ objectives. 
		
		The boundedness of the $\epsilon$-approximate Pareto set is of practical significance since Pareto sets are not necessarily bounded, and the boundedness of an arbitrary Pareto set is not easily determined \textit{a priori}. This presents challenges when selecting an appropriate population size for second-generation algorithms with fixed archive sizes, in particular the NSGA-II: in their mathematical runtime analysis of the NSGA-II, Zheng and Doerr \cite{Zheng2022} demonstrated that the efficiency of the search for the NSGA-II is dependent on the choice of the population size. Specifically, Zheng and Doerr demonstrated that for population sizes of less than or equal to the size of the Pareto set, the algorithm will always miss a constant fraction of the Pareto front. Furthermore, Doerr and Qu \cite{Doerr2023} showed that the NSGA-II does not profit from larger population sizes. Consequently, it is practically impossible to determine an optimal population size \textit{a priori}, which may lead to an inefficient evolutionary search and hinder convergence. The availability of an upper bound for the size of the $\epsilon$-approximate Pareto set is therefore advantageous.
		
		The use of box-domination, enforced at the coarse level within the objective space, ensures convergence to an $\epsilon-$approximate Pareto set, however, convergence to, and maintenance of, an $\epsilon-$Pareto set is more desirable as previously highlighted. This can be achieved by complementing the coarse level application of box dominance with a standard dominance definition applied within each box, used to ensure that only non-dominated solutions are preserved where multiple non-dominated solutions share the same box index vector. This is further complemented in this implementation by a tie-breaker condition for cases in which multiple non-dominated solutions share the same box index vector but none of the solutions dominate. The tie-breaker condition evaluates the normalised euclidean distance in the objective space from each objective vector to the bottom left corner of the box, selecting the solution with the minimal distance. This adds additional selection pressure at the fine scale which aids the convergence of the algorithm. The combination of these procedures is detailed in Algorithm \ref{Alg: Update}, which is an update function designed to maintain a $\epsilon-$Pareto set. 
		
		The update function produces an archive containing unique objective vectors, converging to a set of diverse $\epsilon-$Pareto optimal solutions. The rejection of individuals with identical objective vectors from the archive is beneficial, as their inclusion can hinder diversity and reduce search efficiency. While SPEA-2 includes a mechanism to prevent the admission of overlapping solutions, the NSGA-II lacks such a mechanism. Nojima et al. \cite{Nojima2005} demonstrated, via computational experiments, that for bi-objective optimisation problems the NSGA-II may produce a significant number of overlapping solutions, thereby diminishing the algorithm's performance. 
		
		The use of $\epsilon-$dominance concepts to enforce diversity preservation by limiting the archive to a single non-dominated solution per box also more strictly enforces diversity than the crowding metrics used by NSGA-II and SPEA-2. Although it cannot be guaranteed that solutions will be uniformly spaced, it is guaranteed that each box will contain at most one solution and that solutions within alternate boxes are separated by a distance $d_i \in [\epsilon_i, 2\epsilon_i]$ for each objective. However, solution vectors in neighbouring boxes may still be close in the objective space. In contrast, because NSGA-II and SPEA-2 use crowding metrics which simply discourage the archive to maintain solutions with similar objective vectors, there are no guarantees regarding the distribution of solutions, meaning clustering can still occur.
		
		\begin{algorithm}[H]\label{Alg: Update}
			\caption{Archive update function for the maintenance of a \(\epsilon\)-Pareto set.}
			\KwIn{A population $P_t = \{p_1, \dots, p_N\}$, an archive population $\bar{P}_t = \{\bar{p}_1, \dots, \bar{p}_N\}$, each with a corresponding set of feasible objective vectors $\{\mathbf{z}_1,\dots,\mathbf{z}_N\}$,  $\{\bar{\mathbf{z}}_1,\dots,\bar{\mathbf{z}}_N\}$ and box index vectors $\{\mathbf{b}(\mathbf{z}_i),\dots,\mathbf{b}(\mathbf{z}_N)\}$, $\{\mathbf{b}(\bar{\mathbf{z}}_i),\dots,\mathbf{b}(\bar{\mathbf{z}}_N)\}$.}
			\KwOut{Updated archive population $\bar{P}_{t+1}$.}
			
			\SetKw{KwSet}{Set}
			\SetKwComment{Comment}{$\triangleright$\ }{} 
			
			$\bar{P}_{t+1} \gets \bar{P}_t$ \quad \Comment{Initialise archive for next generation}
			\For{$p \in P_t$  \text{with objective vector} $\mathbf{z}$}{
				$D \gets \emptyset$, \, $accept \gets \mathrm{True}$ \quad \Comment{Initialise dominated set and archive acceptance flag}
				\For{$\bar{p} \in \bar{P}_t$ \text{with objective vector} $\bar{\mathbf{z}}$}{
					\If{$\mathbf{z} \succ_{\mathit{box}} \bar{\mathbf{z}}$}{
						$D \gets D \cup \{\bar{p}\}$ \quad \Comment{Add archive member to dominated set}
					}
					\ElseIf{$\bar{\mathbf{z}} \succ_{\textit{box}} \mathbf{z}$}{
						$accept \gets \mathrm{False}$; \textbf{break} \quad \Comment{Reject $p$ from archive}
					}
					\ElseIf{$\mathbf{b}(\mathbf{z}) = \mathbf{b}(\bar{\mathbf{z}})$}{
						\If{$\mathbf{z} \succ \bar{\mathbf{z}}$}{
							$D \gets D \cup \{\bar{p}\}$ \quad \Comment{Add archive member to dominated set}
						}
						\ElseIf{$\bar{\mathbf{z}} \succ \mathbf{z}$}{
							$accept \gets \mathrm{False}$; \textbf{break} \quad \Comment{Reject $p$ from archive}
						}
						\ElseIf{$TieBreak(\bar{\mathbf{z}}, \mathbf{z})$}{
							$accept \gets \mathrm{False}$; \textbf{break} \quad \Comment{$\bar{\mathbf{z}}$ wins tie-breaker (algorithm \ref{Alg: tie-break})}
						}
						\Else{
							$D \gets D \cup \{\bar{p}\}$ \quad \Comment{$\mathbf{z}$ wins tie-breaker (algorithm \ref{Alg: tie-break})}
						}
					}
				}
				\If{$accept$}{
					$\bar{P}_{t+1} \gets (\bar{P}_{t+1} \setminus D) \cup \{p\}$ \quad \Comment{Remove dominated members and add $p$ to archive}
				}
			}
			
			\Return $\bar{P}_{t+1}$\; 
		\end{algorithm}
		
		\begin{algorithm}[H]\label{Alg: tie-break}
			\caption{Tie-break function for individuals within the same $\epsilon$-box.}
			\KwIn{Archive objective vector $\bar{\mathbf{z}}$, candidate objective vector $\mathbf{z}$, minimum objective values $\mathbf{z}^{\min} = (z_1^{\min}, z_2^{\min})$, and box sizes $\boldsymbol{\epsilon} = (\epsilon_1, \epsilon_2)$.}
			\KwOut{Boolean indicating whether $\bar{\mathbf{z}}$ is preferred over $\mathbf{z}$.}
			\SetKwComment{Comment}{$\triangleright$\ }{}

			$\Delta\bar{z}_1 \gets (\bar{z}_1 - z_1^{\min}) \bmod \epsilon_1$\;
			$\Delta\bar{z}_2 \gets (\bar{z}_2 - z_2^{\min}) \bmod \epsilon_2$\;
			$dist_{\bar{\mathbf{z}}} \gets \sqrt{\Delta\bar{z}_1^2 + \Delta\bar{z}_2^2}$ \quad \Comment{Distance to the box origin for the archive member $\bar{\mathbf{z}}$} 
			
			$\Delta z_1 \gets (z_1 - z_1^{\min}) \bmod \epsilon_1$\;
			$\Delta z_2 \gets (z_2 - z_2^{\min}) \bmod \epsilon_2$\;
			$dist_{\mathbf{z}} \gets \sqrt{\Delta z_1^2 + \Delta z_2^2}$ \quad \Comment{Distance to the box origin for the candidate $\mathbf{z}$}
			
			\If{$dist_{\bar{\mathbf{z}}} < dist_{\mathbf{z}}$}{
				\Return \text{True} \quad \Comment{$\bar{\mathbf{z}}$ is closer to origin, keep archive member}
			}
			\ElseIf{$dist_{\mathbf{z}} < dist_{\bar{\mathbf{z}}}$}{
				\Return \text{False} \quad \Comment{$\mathbf{z}$ is closer to origin, prefer candidate}
			}
			\Else{
				\Return $RandomChoice(\text{True}, \text{False})$ \quad \Comment{Distances are equal, resolve randomly}
			}
		\end{algorithm}
		
		\subsubsection{Automated Design Methodology}\label{Section: Guidelines}
		Adaptive population sizing, introduced by Kollat and Reed \cite{Kollat2005}, enhances algorithmic efficiency and resolves the challenge of selecting population size \textit{a priori}. Additionally, domain-specific knowledge is used to seed the initial archive population, aiding convergence. This seeding also guides the selection of an appropriate $\epsilon$ value for each objective based on a specified maximum archive size. Together, the adaptive population sizing and the self-termination criteria form a robust automated design methodology, minimising the need for parameter calibration. This approach is particularly valuable for MOOPs with computationally expensive objective functions, where calibration methods requiring preliminary runs to systematically explore the parameter space are impractical.
		
		\subsubsection*{Epsilon Selection and Archive Initialisation}
		Equation (\ref{Eq: Bound}) provides an upper bound for the size of an $\epsilon-$approximate Pareto set, dependent upon the choice of $\epsilon$ and the range of each objective function. Therefore, provided the range of each objective function can be ascertained, through re-arrangement of equation (\ref{Eq: Bound}), a specified maximum archive size, $\bar{N}$, can be used to determine $\epsilon$ for each objective:
		\begin{equation}\label{Eq: epsilon}
			\epsilon_i = \frac{f^{\mathrm{max}}_i - f^{\mathrm{min}}_i}{\bar{N}}
		\end{equation}
		which is equivalent to the division of each dimension of the feasible objective space into $\bar{N}$ equal intervals. In this way, the maximum duration required to complete a single generation of the algorithm can be bounded, which is an important consideration when dealing with expensive objective functions.
		
		Using this method, a decision maker can intuitively specify the maximum number of decision vectors, $\bar{N}$, that they wish to select from, with the knowledge that a maximum of 4$\bar{N}$ simulations will be required for any single generation of the algorithm. It is important to note that this approach still implicitly requires the decision maker to select a suitable value of $\epsilon$ based on their requirements. Particularly as decision-makers should be aware that as equation (\ref{Eq: Bound}) only provides an upper bound meaning that, depending on the curvature of the Pareto front, the algorithm may return $n << \bar{N}$ solutions. Therefore common sense should still be applied to ensure that the selection of $\bar{N}$ is consistent with an appropriate $\epsilon$.
		
		In order to facilitate this approach to the selection of $\epsilon$, it is therefore necessary to determine the range of each objective function. Fortunately, using domain specific knowledge, it is possible to either exactly determine or estimate the range of the objective functions.
		
		If the nadir and worst objective vectors are defined as:
		\begin{definition}[Worst Objective Vector]\label{Def: Worst Objective Vector}
			Let \(\boldsymbol{x}_p \in \Omega \) be decision vectors and \(\boldsymbol{z} \in \mathcal{Z}\) feasible objective vectors, where \(\mathcal{Z} = \{\boldsymbol{F}(\boldsymbol{x}_p) \in \mathbb{R}^m : \boldsymbol{x}_p \in \Omega\}\) is the feasible objective space and \(\boldsymbol{F}(\boldsymbol{x}_p) = [f_1(\boldsymbol{x}_p), \dots, f_m(\boldsymbol{x}_p)]^T\) the objective function. For a minimisation problem, \(\boldsymbol{z}^w = [z_1^w, \dots, z_m^w]^T\) is referred to as the worst objective vector if and only if, for all \(i \in \{1,\dots,m\}\), \(z^w_i = \max_{\boldsymbol{x}_p \in \Omega}f_i(\boldsymbol{x}_p)\).
		\end{definition}
		\begin{definition}[Nadir Objective Vector]\label{Def: Nadir Objective Vector}
			Let \(\mathcal{X}_{\mathcal{P}} \in \Omega\) be the set of decision vectors whose image under the objective function is \(\mathcal{P} = \{\boldsymbol{F}(\boldsymbol{x}_p) \in \mathbb{R}^m : \boldsymbol{x}_p \in \mathcal{X}_{\mathcal{P}}\}\), the set of Pareto-optimal vectors of \(\mathcal{Z} \subseteq \mathbb{R}^m\). For a minimisation problem, \(\boldsymbol{z}^n = [z_1^n, \dots, z_m^n]^T\) is referred to as the nadir objective vector if and only if, for all \(i \in \{1,\dots,m\}\), \(z^n_i = \max_{\boldsymbol{x}_p \in \mathcal{X}_{\mathcal{P}}}f_i(\boldsymbol{x}_p)\).
		\end{definition}
		Then for the class of bi-objective minimisation problems under consideration, it is trivial to \textit{a priori} identify the decision vectors corresponding to the worst objective vector, which may coincide with the nadir objective vector. This is predicated on design guidelines that constrain the decision space:
		\begin{enumerate}
			\item Each intervention should have at a minimum an encoded state of exclusion and an encoded state of inclusion.
			\item It should be feasible to model each encoded state for each intervention in combination with each encoded state for all other interventions.
			\item Each intervention should have a positive cost of installation and, in isolation, should not increase the flood exposure metric.
			\item No combination of interventions should result in an increase in the flood exposure metric with respect to the baseline.
		\end{enumerate}
		
		Under these assumptions, the worst objective vector can be estimated via the extreme decision vectors corresponding to the `\textit{do minimum}' decision vector, $\boldsymbol{x}_{p, \min}$, comprising of the baseline exclusion of any interventions, and the `\textit{do maximum}' decision vector, $\boldsymbol{x}_{p, \max}$, comprising of the maximal inclusion of all interventions. Furthermore, the outlined assumptions also imply that $\boldsymbol{x}_{p, \min}$ is Pareto-optimal and that $\mathbf{F}(\boldsymbol{x}_{p, \min}) = \boldsymbol{z}_{\min} = [\min_{\boldsymbol{x}_p \in \Omega} f_1(\boldsymbol{x}_p), \max_{\boldsymbol{x}_p \in \Omega} f_2(\boldsymbol{x}_p)]$. Proof by contradiction is provided by supposing that $\boldsymbol{z}_{\min} \notin \mathcal{P}$ which implies that there exists $\boldsymbol{w} \in \mathcal{Z}$ such that $\boldsymbol{w} \succ \boldsymbol{z}_{\min}$ as per definition \ref{Def: Dominance}. Let $\boldsymbol{z}_{\min} = [z_1, z_2]^T$ where by definition $z_1 = \min_{\boldsymbol{x}_p \in \Omega}f_1(\boldsymbol{x}_p)$, then $\boldsymbol{w} = [w_1, w_2]^T$ where $w_1 = \min_{\boldsymbol{x}_p \in \Omega}f_1(\boldsymbol{x}_p)$ and $w_2 < z_2$, as $w_1$ cannot be strictly smaller than the lower bound of $f_1$. However, the existence of the vector $\boldsymbol{w}_{\min} = [\min_{\boldsymbol{x}_p \in \Omega}f_1(\boldsymbol{x}_p), w_2]^T$ with $w_2 < z_2$ implies the existence of a zero cost intervention which reduces the flood exposure metric, in contradiction with design guideline 3, and therefore $\boldsymbol{w} \notin \mathcal{Z}$. Likewise, $z_2$ must equal $\max_{\boldsymbol{x}_p \in \Omega} f_2(\boldsymbol{x}_p)$, the maximal flood risk, as a consequence of design guideline 4.
		
		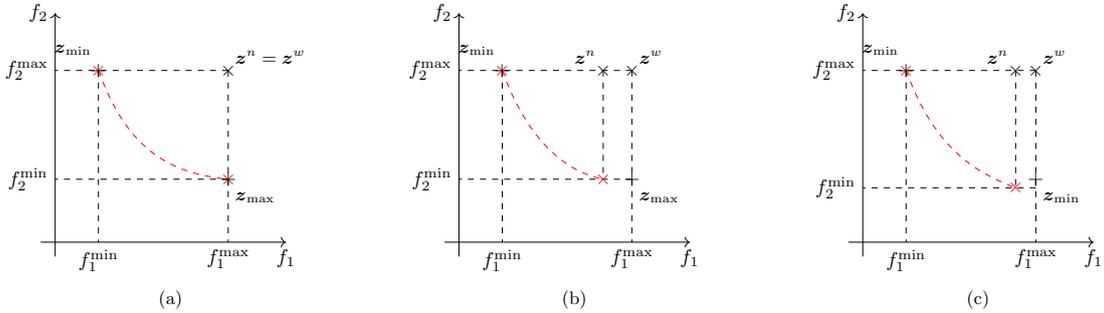
\begin{figure}[hbt!]
			\centering
			\resizebox{0.9\textwidth}{!}{
			\begin{tikzpicture}
				\draw[->] (-0.25,0) -- (4.0,0) node[anchor=north] {$f_1$};
				\draw[->] (0,-0.25) -- (0,4.0) node[anchor=east] {$f_2$};
				
				\draw[dashed] (3.00, 3.00) -- (3.00, 0.00) node[anchor=north] {$f_1^{\mathrm{max}}$};
				\draw[dashed] (3.00, 1.10) -- (0.00, 1.10) node[anchor=east] {$f_2^{\mathrm{min}}$};
				\draw[dashed] (0.75, 3.00) -- (0.75, 0.00) node[anchor=north] {$f_1^{\mathrm{min}}$};
				\draw[dashed] (3.00, 3.00) -- (0.00, 3.00) node[anchor=east] {$f_2^{\mathrm{max}}$};
				\draw[dashed, red] (0.75, 3.00) .. controls (1.25, 1.60) and (2.00, 1.20) .. (3.00, 1.10);
				\node (nadir1) [red, inner sep=0pt] at (0.75, 3.00) {$\times$};
				\node (y1) [black, inner sep=0pt] at (0.75, 3.00) {$+$};
				\node (dn1) [anchor=south east, black] at (0.75, 3.15) {\small $\boldsymbol{z}_{\min}$};
				\node (nadir2) [red, inner sep=0pt] at (3.00, 1.10) {$\times$};
				\node (x1) [black, inner sep=0pt] at (3.00, 1.10) {$+$};
				\node (dm1) [anchor=north west, black] at (3.00, 1.0) {\small $\boldsymbol{z}_{\max}$};
				\node (z1) [black, inner sep=0pt] at (3.00, 3.00) {$\times$};
				\node (dz1) [anchor=south west, black] at (3.00, 3.00) {\small $\boldsymbol{z}^n = \boldsymbol{z}^w$};
				\node at (2.00, -1.00) {\small (a)};
				
				\def\xshift{7} 
				\draw[->] (-0.25+\xshift,0) -- (4.0+\xshift,0) node[anchor=north] {$f_1$};
				\draw[->] (0+\xshift,-0.25) -- (0+\xshift,4.0) node[anchor=east] {$f_2$};
				
				\draw[dashed] (3.00+\xshift, 3.00) -- (3.00+\xshift, 0.00) node[anchor=north] {$f_1^{\mathrm{max}}$};
				\draw[dashed] (3.00+\xshift, 1.10) -- (0.00+\xshift, 1.10) node[anchor=east] {$f_2^{\mathrm{min}}$};
				\draw[dashed] (0.75+\xshift, 3.00) -- (0.75+\xshift, 0.00) node[anchor=north] {$f_1^{\mathrm{min}}$};
				\draw[dashed] (3.00+\xshift, 3.00) -- (0.00+\xshift, 3.00) node[anchor=east] {$f_2^{\mathrm{max}}$};
				\draw[dashed] (2.50+\xshift, 3.00) -- (2.50+\xshift, 1.10);
				\draw[dashed, red] (0.75+\xshift, 3.00) .. controls (1.25+\xshift, 1.60) and (2.00+\xshift, 1.20) .. (3.00+\xshift-0.5, 1.10);
				\node (nadir10) [red, inner sep=0pt] at (0.75+\xshift, 3.00) {$\times$};
				\node (y10) [black, inner sep=0pt] at (0.75+\xshift, 3.00) {$+$};
				\node (dn10) [anchor=south east, black] at (0.75+\xshift, 3.15) {\small $\boldsymbol{z}_{\min}$};
				\node (nadir20) [red, inner sep=0pt] at (3.00+\xshift-0.5, 1.10) {$\times$};
				\node (x10) [black, inner sep=0pt] at (3.00+\xshift, 1.10) {$+$};
				\node (dm10) [anchor=north west, black] at (3.00+\xshift, 1.00) {\small $\boldsymbol{z}_{\max}$};
				\node (z10) [black, inner sep=0pt] at (2.50+\xshift, 3.00) {$\times$};
				\node (dz10) [anchor=south east, black] at (2.50+\xshift, 3.00) {\small $\boldsymbol{z}^n$};
				\node (z10) [black, inner sep=0pt] at (3.00+\xshift, 3.00) {$\times$};
				\node (dz10) [anchor=south west, black] at (3.00+\xshift, 3.00) {\small $\boldsymbol{z}^w$};
				\node at (2.00+\xshift, -1.00) {\small (b)};
				
				\draw[->] (-0.25+2*\xshift,0) -- (4.0+2*\xshift,0) node[anchor=north] {$f_1$};
				\draw[->] (0+2*\xshift,-0.25) -- (0+2*\xshift,4.0) node[anchor=east] {$f_2$};
				
				\draw[dashed] (3.00+2*\xshift, 3.00) -- (3.00+2*\xshift, 0.00) node[anchor=north] {$f_1^{\mathrm{max}}$};
				\draw[dashed] (3.00+2*\xshift, 1.10-0.15) -- (0.00+2*\xshift, 1.10-0.15) node[anchor=east] {$f_2^{\mathrm{min}}$};
				\draw[dashed] (0.75+2*\xshift, 3.00) -- (0.75+2*\xshift, 0.00) node[anchor=north] {$f_1^{\mathrm{min}}$};
				\draw[dashed] (3.00+2*\xshift, 3.00) -- (0.00+2*\xshift, 3.00) node[anchor=east] {$f_2^{\mathrm{max}}$};
				\draw[dashed] (3.00-0.35+2*\xshift, 3.00) -- (3.00-0.35+2*\xshift, 1.10-0.15);
				\draw[dashed, red] (0.75+2*\xshift, 3.00) .. controls (1.25+2*\xshift, 1.60) and (2.00+2*\xshift, 1.20) .. (3.00+2*\xshift-0.35, 1.10-0.15);
				\node (nadir100) [red, inner sep=0pt] at (0.75+2*\xshift, 3.00) {$\times$};
				\node (y100) [black, inner sep=0pt] at (0.75+2*\xshift, 3.00) {$+$};
				\node (dn100) [anchor=south east, black] at (0.75+2*\xshift, 3.15) {\small $\boldsymbol{z}_{\min}$};
				\node (nadir200) [red, inner sep=0pt] at (3.00+2*\xshift-0.35, 1.10-0.15) {$\times$};
				\node (x100) [black, inner sep=0pt] at (3.00+2*\xshift, 1.10) {$+$};
				\node (dm100) [anchor=north west, black] at (3.00+2*\xshift, 1.00) {\small $\boldsymbol{z}_{\min}$};
				\node (z10) [black, inner sep=0pt] at (3.00-0.35+2*\xshift, 3.00) {$\times$};
				\node (dz10) [anchor=south east, black] at (3.00-0.35+2*\xshift, 3.00) {\small $\boldsymbol{z}^n$};
				\node (z10) [black, inner sep=0pt] at (3.00+2*\xshift, 3.00) {$\times$};
				\node (dz10) [anchor=south west, black] at (3.00+2*\xshift, 3.00) {\small $\boldsymbol{z}^w$};
				\node at (2.00+2*\xshift, -1.00) {\small (c)};
					
			\end{tikzpicture}
			} 
			\caption{The potential relative locations of the nadir objective vector ($\boldsymbol{z}^n$) and worst objective vector ($\boldsymbol{z}^w$) for the three outlined scenarios. Let $\mathcal{P} = \mathbf{F}(\mathcal{X}_p)$ denote the Pareto front: 
				In (a) $\boldsymbol{z}^n = \boldsymbol{z}^w$, $\boldsymbol{z}_{\min}, \boldsymbol{z}_{\max} \in \mathcal{P}$ with $\mathcal{P}$ spanning $[f_1(\boldsymbol{x}_{p,\min}), f_1(\boldsymbol{x}_{p,\max})] \times [f_2(\boldsymbol{x}_{p,\max}), f_2(\boldsymbol{x}_{p,\min})]$. 
				In (b) $\boldsymbol{z}^n \neq \boldsymbol{z}^w$, $\boldsymbol{z}_{\min} \in \mathcal{P}$, $\boldsymbol{z}_{\max} \notin \mathcal{P}$ where $\mathcal{P}$ is bounded by $[f_1(\boldsymbol{x}_{p,\min}), f_1^{\max}] \times [f_2(\boldsymbol{x}_{p,\max}), f_2(\boldsymbol{x}_{p,\min})]$, with $f_1^{\max} < f_1(\boldsymbol{x}_{p,\max})$. 
				(c) $\boldsymbol{z}^n \neq \boldsymbol{z}^w$, with bounds $[f_1(\boldsymbol{x}_{p,\min}), f_1^{\max}] \times [f_2^{\mathrm{lb}}, f_2(\boldsymbol{x}_{p,\min})]$, with $f_2^{\min} > f_2(\boldsymbol{x}_{p,\max})$.}
			\label{Fig: Nadir Vectors}
		\end{figure}
		
		The Pareto-optimality of $\boldsymbol{x}_{p, \max}$ is dependent upon the nature of the encoded characteristics for any local features and also the interdependency of the effectiveness of the considered interventions. The three potential scenarios are illustrated in Figure \ref{Fig: Nadir Vectors}. The Pareto front can be described by a function $p_f(x)$, defined over a closed interval $[f_1^{\min}, f_1^{\max}]$, whereby $p_f(x)$ maps the image of the Pareto set under the first objective function, $f_1$, onto the image of the Pareto set under the second objective function, $f_2$: $x \in [f_1^{\min}, f_1^{\max}] \mapsto p_f(x)$. It is straightforward to prove, as shown by Auger et al. \cite{Auger2009}, that for a minimisation problem $p_f(x)$ is a strictly monotonically decreasing function by virtue of the dominance definition \ref{Def: Dominance}. 
		
		In the first scenario, shown in Figure \ref{Fig: Nadir Vectors}a, the effectiveness of each intervention is independent of the inclusion or exclusion of each other intervention and the cost of implementation for each intervention state is proportional to its effectiveness. In this scenario the nadir objective vector and consequently the span of the Pareto set can be determined \textit{a priori} by evaluating the fitness of $\boldsymbol{x}_{p, \min}$ and $\boldsymbol{x}_{p, \max}$. Therefore via the evaluation of the extreme objective vectors upon initialisation, the bounds for each objective function can be obtained amongst other useful information regarding the feasible objective space and the Pareto front.
		
		In the second scenario, shown in Figure \ref{Fig: Nadir Vectors}b, there is some redundancy within the system and some more costly interventions, or some more costly combinations of interventions, do not contribute to a further reduction in the flood exposure metric. In this scenario the evaluation of $\mathbf{F}(\boldsymbol{x}_{p, \min})$ and $\mathbf{F}(\boldsymbol{x}_{p, \max})$ still elucidates the bounds for each objective function however, less information is available regarding the character of the Pareto front.
		
		In the final scenario, shown in Figure \ref{Fig: Nadir Vectors}c, the most expensive combination of interventions proves less effective than some cheaper combination of interventions. This may occur where an upstream intervention alters a flow path significantly, reducing the effectiveness of downstream interventions which intercept the original flow path. Or where a less expensive encoded state for a \textit{local feature} outperforms a more expensive encoded state. In this scenario, the evaluation of $\mathbf{F}(\boldsymbol{x}_{p, \min})$ and $\mathbf{F}(\boldsymbol{x}_{p, \max})$ only reveals the upper bounds for each of the objective functions. 
		
		Therefore, following the evaluation of $\mathbf{F}(\boldsymbol{x}_{p, \min})$ and $\mathbf{F}(\boldsymbol{x}_{p, \max})$, the upper bounds for each objective function are always known and only in the third scenario, are the lower bounds for each objective unknown. Consequently, the proposed strategy is to assume that $\boldsymbol{z}_{\max} = [f_1^{\max}, f_2^{\min}]^T$ to enable the use of equation \ref{Eq: epsilon}.
		
		Furthermore, since this approach necessitates the evaluation of $\mathbf{F}(\boldsymbol{x}_{p, \min})$ and $\mathbf{F}(\boldsymbol{x}_{p, \max})$ upon the initialisation of the algorithm and it is known that the at least one of the corresponding feasible objective vectors is Pareto-optimal, it is efficient to use $\boldsymbol{x}_{p, \min}$ and $\boldsymbol{x}_{p, \max}$ to seed the initial archive population. Since the update function, as per Algorithm \ref{Alg: Update}, results in an archive which contains the Pareto points of all prior generated vectors, the initial seeding process encourages the early discovery and maintenance of the boundary solutions within the archive. Specifically for the first generation of offspring, this process guarantees that a large proportion of the offspring population is derived from the recombination of one of the extreme decision vectors and one randomly generated initial population member. Over subsequent generations, the progressive addition of new archive members naturally shifts the focus away from the initial focussed exploration and exploitation of boundary solutions.
		
		From the perspective of schema theory, this initialisation process facilitates the early identification and effective propagation of highly fit schemata associated with the boundary solutions. According to schema theory \cite{Holland1975}, a schema (plural: schemata) is a pattern representing a subset of solutions in the search space. For example, the schema `$1\star0\star\star$' corresponding to all bit-strings where $b_1=1$ and $b_3=0$, with the remaining bits free to vary. Under the assumption of quasi-decomposability — where a problem can be divided into smaller, interrelated sub-problems — Goldberg's building block hypothesis \cite{Goldberg1989} posits that `\textit{Short, low order, and highly fit schemata are sampled, recombined, and resampled to form strings of potentially higher fitness}'. Consequently, the seeding process is expected to accelerate convergence while promoting diversity by preserving critical building blocks associated with the boundary solutions, identified during the early stages of evolution.
		
		\subsubsection*{Adaptive Epsilon Sizing}\label{Section: Adaptive Epsilon}
		While the initial seeding of the archive using the extreme decision vectors, $\boldsymbol{x}_{p, \min}$ and $\boldsymbol{x}_{p, \max}$, provides a robust starting point, the initial objective bounds may prove conservative. As illustrated in Figure \ref{Fig: Nadir Vectors}, the algorithm may discover solutions that either strictly dominate or box-dominate the initial boundary solutions, which is possible even when $\boldsymbol{x}_{p, \min}$ and $\boldsymbol{x}_{p, \max}$ are both Pareto optimal since $\mathcal{P}_{\epsilon} \subseteq \mathcal{P}$. Such a scenario renders the initial selection of $\epsilon$, based upon a prescribed maximum front size, invalid. 
		
		Consequently, following the archive update procedure, a check is performed to determine whether a newly archived individual strictly dominates, box-dominates, or extends beyond the current boundary solutions. To maintain the desired resolution, the algorithm then dynamically adapts $\epsilon$ based upon the updated boundary solutions. The procedure used to dynamically recalculate $\epsilon$ is outlined in Algorithm \ref{Alg: update epsilon}.
		
		\begin{algorithm}[H]\label{Alg: update epsilon}
			\caption{Dynamically update $\epsilon$ based on the current archive extents.}
			\KwIn{An updated archive population $\bar{P}_{t+1} = \{\bar{p}_1,\dots,\bar{p}_N\}$, with a corresponding set of feasible objective vectors $\{\mathbf{z}_1,\dots,\mathbf{z}_N\}$ and box index vectors $\{\mathbf{b}(\mathbf{z}_i),\dots,\mathbf{b}(\mathbf{z}_N)\}$ plus objective bounds given by $\mathbf{z}_{\min} = [f_1^{\min}, f_2^{\max}]^T$ and $\mathbf{z}_{\max} = [f_1^{\max}, f_2^{\min}]^T$. A user specified maximum front size $\bar{N}$ and a small tolerance $tol = 10^{-12}$.}
			\KwOut{An updated value for $\epsilon$ based on the updated archive extents.}
			\SetKwComment{Comment}{$\triangleright$\ }{}
			\For{$\bar{p} \in \bar{P}_{t+1}$ with objective vector $\bar{\mathbf{z}} = [\bar{z}_1, \bar{z}_2]^T$}{
				\If{$\bar{\mathbf{z}} \succ_{\mathit{box}} \mathbf{z}_{\min}$}{
					$\mathbf{z}_{\min} \gets [f_1^{\min}, \bar{z}_2 + tol]^T$ \quad \Comment{Left boundary is box-dominated, update $f_2^{\max}$ only}
				}
				\If{$\bar{\mathbf{z}} \succ \mathbf{z}_{\max}$}{
					$\mathbf{z}_{\max} \gets [\bar{z}_1 + tol, \bar{z}_2 - tol]^T$ \quad \Comment{Right boundary is strictly dominated, update $\mathbf{z}_{\max}$}
				}
				\ElseIf{$\bar{\mathbf{z}} \succ_{\mathit{box}} \mathbf{z}_{\max}$}{
					$\mathbf{z}_{\max} \gets [f_1^{\max}, \bar{z}_2 - tol]^T$ \quad \Comment{Right boundary is box-dominated, update $f_2^{\min}$ only}
				}
				\ElseIf{$\bar{z}_1 > f_1^{\max} \lor \bar{z}_2 < f_2^{\min}$}{
					$\mathbf{z}_{\max} \gets [\max{(\bar{z}_1 + tol, f_1^{\max})}, \min{(\bar{z}_2 - tol, f_2^{\min})}]^T$ \quad \Comment{Expand extents}
				}
			} 
			$\mathbf{\epsilon} \gets \left[\frac{f^{\mathrm{max}}_1 - f^{\mathrm{min}}_1}{\bar{N}}, \frac{f^{\mathrm{max}}_2 - f^{\mathrm{min}}_2}{\bar{N}}\right]^T$ \quad \Comment{Update $\epsilon$ using equation (\ref{Eq: epsilon})}
			\Return $\mathbf{\epsilon}$\;
		\end{algorithm}
		
		This ensures that the user-defined maximum front size, $\bar{N}$, remains optimally distributed across the current known extent of the archive. However, a recalculation of $\epsilon$ intrinsically shifts the dimensional boundaries of the $\epsilon$-boxes which delineate the objective space. Consequently, the existing archive must be pruned to maintain the $\epsilon$-approximate Pareto set properties, as outlined in Definition \ref{def: eps-approx set}, under the new grid configuration. The procedure used to prune the archive is outlined as follows:
		\begin{enumerate}
			\item Re-mapping: All existing individuals are re-evaluated and assigned to a new box index vector based on the updated $\epsilon_i$ values.
			\item Inter-box-dominance: All archive members are compared and any individuals which are box-dominated are removed from the archive.
			\item Intra-box-dominance: Where multiple non-dominated archive members share the same box index vector, strict dominance and tie-breaker checks are performed to select a single dominant solution to be retained, as in Algorithm \ref{Alg: Update}.
		\end{enumerate}
		
		\subsubsection*{Adaptive Population Sizing}
		Following the initialisation of the archive population, a random initial population, $P_0$ of size $N=5$ is initialised and an intra-run criterion is introduced to trigger a change in the population size following failure of the archive to adequately converge. Once the intra-run criterion is triggered, the population $P_t$ is discarded and replaced with a new population of size:
		\begin{equation}\label{eq: pop size}
			N = \min{\left(2N, \, \frac{1}{i}|\bar{P_t}|\right)}
		\end{equation}
		where $i$ is the injection rate set to $25\%$ on the recommendation of Kollat and Reed \cite{Kollat2005}. There are two conditions to consider:
		\begin{enumerate}[label=Case \Roman*:, leftmargin=5em, rightmargin=2em, topsep=0em, partopsep=0em]
			\item The new population size is smaller than the subset of the archive to be injected, ${i}|\bar{P_t}| > N$. In this case, the injection population is truncated via random selection to fit.
			\item The injected population is insufficient to fill the new population, ${i}|\bar{P_t}| \leq N$. In this case, the population is filled with randomly generated offspring.
		\end{enumerate}
		The intra-run criterion therefore monitors the convergence of the archive to adjust the population size adaptively in an attempt to balance exploration and exploitation. Full details for the intra-run criterion are provided in Algorithm \ref{Alg: Intra-Run}. Using equation (\ref{eq: pop size}) to update the population size ensures that the maximum size of the population is bounded by four times the size of the archive, which itself is bounded as per equation (\ref{Eq: Bound}). This is particularly important for MOOPs with expensive objective functions as it also ensures that the time required to evaluate any single generation is bounded. The limit behaviour of the connected runs is equivalent to time continuation \cite{Goldberg2000}, with the $25\%$ injection rate assisting the algorithm in escaping from local non-dominated fronts \cite{Kollat2005}.
		
		\begin{algorithm}[H]\label{Alg: Intra-Run}
			\caption{Intra-run criterion for adaptively updating the population size in an \(\epsilon\)-MOEA.}
			\KwIn{Population $P_{t}$ of size $N_{t}$, archive population $\bar{P}_t$ of size $\bar{N}_t$, the size of the archive at the previous generation, $\bar{N}_{t-1}$, the number of archive members replaced during advancement from $t-1$ to $t$, denoted $K$, and a convergence threshold $\Delta p$.}
			\KwOut{Updated population $P_{t}$ of size $N_{t}$.}
			
			\SetKw{KwSet}{Set}
			\SetKwComment{Comment}{$\triangleright$\ }{} 
			
			$\Delta p \gets 0.1$\quad \Comment{Set the convergence threshold to $0.1$, as per \cite{Reed2003}}
			$i \gets 0.25$\quad \Comment{Set the injection rate to $0.25$, as per \cite{Kollat2005}}
			
			\If{$\Delta p > \left(\frac{|\bar{N}_{t}-\bar{N}_{t-1}|}{\bar{N}_{t-1}} \right)$}{
				\Comment{Inadequate change in archive size, check for the number of archive members replaced by superior solutions.}
				\If{$\Delta p > \left(\frac{|K-\bar{N}_{t-1}|}{\bar{N}_{t-1}} \right)$}{
					\Comment{Inadequate number of archive member replacements signalling a lack of convergence.}
					$N_{t} \gets \min{\left(2N_{t-1}, \, \frac{1}{i}|\bar{P}_{t}|\right)}$\quad \Comment{Updated population size}
					$N_{\mathrm{inj}} \gets i|\bar{P}_t|$ \quad \Comment{Injected archive population size}
					\If{$N_{\mathrm{inj}} > N_{t}$}{
						$P_t \gets \{\mathbf{x}_i \in \bar{P}_t\}_{i=1}^{N_{t}}$ \quad \Comment{Pre-condition the population with randomly selected archive members}
					}
					\Else{
						$P_t \gets \{\mathbf{x}_i \in \bar{P}_t\}_{i=1}^{N_{\mathrm{inj}}} \cup \{\mathbf{y}_j \in \{0,1\}^n\}_{j=1}^{N_t-N_{\mathrm{inj}}}$ \quad \Comment{Inject archive members and fill remainder with random offspring}
					}
				}
			}
			\Return $P_t$\; 
		\end{algorithm}
				
		\subsubsection*{Self-termination Criterion}\label{Section: Self-Termination}
		The hyper-volume indicator, also referred to as the S-metric \cite{Zitzler2000}, is used as a general performance indicator capturing the effect of both convergence and distribution \cite{Miqing2019, Audet2021}. The S-metric describes the volume of the objective subspace dominated by the non-dominated Pareto front approximation, denoted $\mathcal{P}^*$, with respect to a specified reference objective vector, denoted $\mathbf{z}_{\mathrm{ref}}$, as shown in Figure \ref{Fig: S-Metric}. The location of the reference objective vector is arbitrary; however, in order to ensure that the maximum hyper-volume is achieved when $\mathcal{P}^* = \mathcal{P}$ for a discontinuous Pareto front of finite size, $\mathbf{z}_{\mathrm{ref}}$ must be selected such that it is strictly dominated by the nadir objective vector.
		
		The hyper-volume indicator is given as:
		\begin{equation}\label{Eq: Hyper-volume indicator}
			HV(\mathcal{P}^*; \mathbf{z}_{\mathrm{ref}}) = \lambda_m \left(\bigcup_{\mathbf{p}^* \in \mathcal{P}^*} [\mathbf{p}^*, \mathbf{z}_{\mathrm{ref}}] \right)
		\end{equation}	
		where $\lambda_m$ is the $m$-dimensional Lebesgue measure. Where the exact Pareto set, $\mathcal{P}$, is known, the hyper-area ratio can be defined as:
		\begin{equation}\label{Eq: Hyper-volume ratio}
			HR(\mathcal{P}^*, \mathcal{P}; \mathbf{z}_{\mathrm{ref}}) = \frac{HV(\mathcal{P}^*; \mathbf{z}_{\mathrm{ref}})}{HV(\mathcal{P}; \mathbf{z}_{\mathrm{ref}})}
		\end{equation}
		where $HR(\mathcal{P}^*, \mathcal{P}; \mathbf{z}_{\mathrm{ref}}) = 1$, or 100\% if expressed as a percentage, indicates convergence to the exact Pareto set ($\mathcal{P}^* = \mathcal{P}$). 
		
		\begin{figure}[hbt!]
			\centering
			\resizebox{0.25\textwidth}{!}{
				\begin{tikzpicture}
					\draw[->] (-0.25,0) -- (3.5,0) node[anchor=north] {$f_1$};
					\draw[->] (0,-0.25) -- (0,3.5) node[anchor=east] {$f_2$};
					
					\foreach \x/\y [count=\n] in {0.75/2.85, 1.25/2, 1.5/1.68, 2/1.24, 2.75/0.79} {
						\node[red] at (\x,\y) {$\times$};
						\node[anchor=north east, red] at (\x,\y) {$p_{\n}^*$};
					}
					\node[black] at (3.5,3.5) {$\times$};
					\node[anchor=south west, black] at (3.5,3.5) {$z_{\mathrm{ref}}$};
					\fill[pattern=north east lines, pattern color=gray!80, draw=gray!50] (0.75,3.5) -- (0.75,2.85) -- (1.25,2.85) -- (1.25,2) -- (1.5,2) -- (1.5,1.68) -- (2,1.68) -- (2,1.24) -- (2.75,1.24) -- (2.75,0.79) -- (3.5,0.79) -- (3.5,3.5) -- (0.75,3.5) -- cycle;
				\end{tikzpicture}
			} 
			\caption{Illustration of the S-metric (hyper-volume indicator) for a Pareto front approximation, represented by the grey-shaded region dominated by the approximation $\mathcal{P}^* = \{ \mathbf{p}^*_1, \dots, \mathbf{p}^*_5 \}$, with respect to the reference objective vector $\mathbf{z}_{\mathrm{ref}}$.}
			\label{Fig: S-Metric}
		\end{figure}
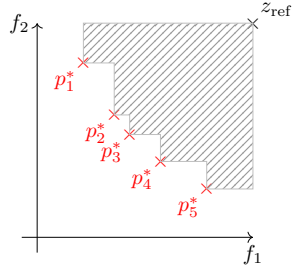
		
		Within the context of the proposed algorithm, the hyper-volume indicator is used to monitor the convergence of the archive, forming the basis of the inter-run self termination criterion. To prevent biasing of the metric due to the differences in scales between the objectives, the hyper-volume metric is calculated using a normalised objective space $\mathcal{Z} \in [0, 1] \times [0, 1]$, based upon the dynamic archive bounds. Furthermore, to ensure that the boundary solutions make a positive non-zero contribution to the calculated hyper-volume, the normalised reference objective vector is offset beyond the strict maximum bounds of the archive: $\mathbf{z}_{\mathrm{ref}} = [1.25, 1.25]^T$. Within the context of MOOPs with expensive objective functions, the computational expense involved in the computation of the S-metric for the archive is considered to be negligible, especially since the user controls the maximum size of the archive, $\bar{N}$. 

		The self-termination criterion specifically measures the change in the magnitude of the hyper-volume metric across a number of generations, terminating the algorithm if, over the specified window, the hyper-volume metric changes by less than the hyper-volume of a single box. In this way, the self-termination criterion is tied to the initial indirect specification of $\epsilon$. This is in contrast to the inter-run criterion proposed by Kollat and Reed \cite{Kollat2005}, which is based upon the same convergence measure used by the intra-run criterion (Algorithm \ref{Alg: Intra-Run}). The more accurate but computationally expensive hyper-volume based measure is used in this instance as measuring convergence based upon archive additions and replacement is agnostic of the magnitude of convergence; when evaluating expensive objective functions every evaluation is precious and it is wasteful to continue when convergence is continual but incremental. 
		
		The proposed inter-run criterion is determined based on convergence across $w$ generations, where for expensive objective functions, a smaller value of $w$ is recommended to maximise efficiency in terms of convergence rate per fitness evaluation. Generally for MOEAs, termination is handled via manual termination, based on a decision-maker's judgement, or based on fixed criteria such as elapsed time or a maximum number of fitness evaluation or generations. The proposed inter-run criterion complements these methods to ensure that the algorithm is adaptively terminated when convergence sufficiently halts. 
		
		\subsubsection{Offspring Generation}\label{Section: Offspring Generation}
		A simple selection process is employed to generate an offspring population, $P_{t+1}$. During selection, two candidates are randomly drawn from the offspring population, and the individual with the lowest overall SPEA-2 fitness is selected. Ties are broken by favouring the solution situated in the least crowded region of the objective space. The second mating partner is selected uniformly from the archive. Since the offspring population is preconditioned using the archive following the invocation of the intra-run criterion (Algorithm \ref{Alg: Intra-Run}), this strategy balances the robust exploitation of existing elite solution with a highly selective, fitness-driven exploration of the search space. The introduction of genetic diversity following the filling of the population with randomly generated individuals is crucial to ensure that the algorithm can escape local non-dominated fronts \cite{Kollat2005}. Through these two mechanisms, the algorithm seeks to balance the exploration of the search space with the exploitation of existing individuals.
		
		The use of SPEA-2 environmental selection mechanics (Algorithm \ref{Alg: SPEA-2}) to select between the randomly drawn offspring is also crucial, as simple selection based upon strict dominance lacks the capacity to differentiate between mutually non-dominated individuals. This is particularly important as archive preconditioning introduces a significant number of offspring which are guaranteed to be mutually non-dominating, resulting in a lack of selection pressure. Alternatively, NSGA-II environmental selection mechanics (Algorithm \ref{Alg: NSGA-II}) could also be used; however, the more sophisticated fitness and distance metrics of SPEA-2 are preferred in this instance as the additional computational expense is negligible with respect to the evaluation of the objective function. 
		
		\section{Validation}
		When evaluating the capability of an evolutionary algorithm using standard test suites, it is necessary to consider the implications of the No Free Lunch theorem No Free Lunch theorem \cite{Wolpert1997}. The theorem posits that, averaged across all possible objective functions, no single optimisation algorithm universally outperforms any other. Consequently, without incorporating domain-specific knowledge, there are no formal guarantees of an algorithm's general effectiveness. Wile embedding problem-specific knowledge specialises an algorithm and enhances its performance on a target class of problems, it inherently reduces its efficacy on unrelated domains. Therefore, test suites are typically designed to isolate major problem domain characteristics, such that an algorithm performing well in the test environment can be expected to be broadly applicable to similar real-world problems \cite{Veldhuizen1999b}. The underlying $\epsilon$-MOEA architecture upon which this work is based has already been rigorously tested against standard mathematical benchmarks \cite{Kollat2005}, demonstrating the broader baseline applicability of the algorithm's mechanics. 
		
		Given that the proposed optimisation tool is intended specifically for the placement and design of BGI in urban environments, domain-specific validation test cases are utilise the validate its performance. The validation is split into two phases. The first test case involves the exhaustive study of a tractable search space, enabling a rigorous, exact validation of the algorithm's performance with respect to a known, true Pareto set. The second test case involves a complex, realistic modelling setup for which an exhaustive search is intractable. Because the true Pareto set for this scenario is unknown, inter-algorithm comparisons form the basis of the performance analysis. 
		
		In all test cases the following settings are used for all algorithms:
		\begin{description}[style=nextline]
			\item Mutation probability = $L^{-1}$, where $L$ is the length of the representation.
			\item Mutation operator = bit-flip mutation.
			\item Recombination probability = 1.0.
			\item Recombination operator = one-point crossover.
			\item Random sampling technique = stochastic universal sampling.
		\end{description}
		For the proposed algorithm local features are encoded using Gray coding, for the benchmark algorithms standard binary is used.
		
		\subsection{Study Area and Initial Model Setup}
		The city centre of Newcastle upon Tyne, UK, provides an ideal case study due to its significant flood risk, particularly during severe storm events. One of the most notable recent storm events occurred on June 28$^{\mathrm{th}}$, 2012, and is locally referred as the ‘\textit{Toon Monsoon}’. Over the past decade, the area has been the focus of numerous research studies \cite{Bertsch2022, Dawson2020, Fenner2019, Glenis2018, Iliadis2023b, Kutija2014, Vercruysse2019}. As shown in Figure \ref{Fig: Study Area}, the city centre is characterised by a mix of historic and commercial buildings, as well as green spaces, most notably the Town Moor and Leazes Park, which together cover a substantial area (almost 50\% of the study area).
		
		\begin{figure}[hbt!]
			\centering
			\includegraphics[width=0.85\linewidth]{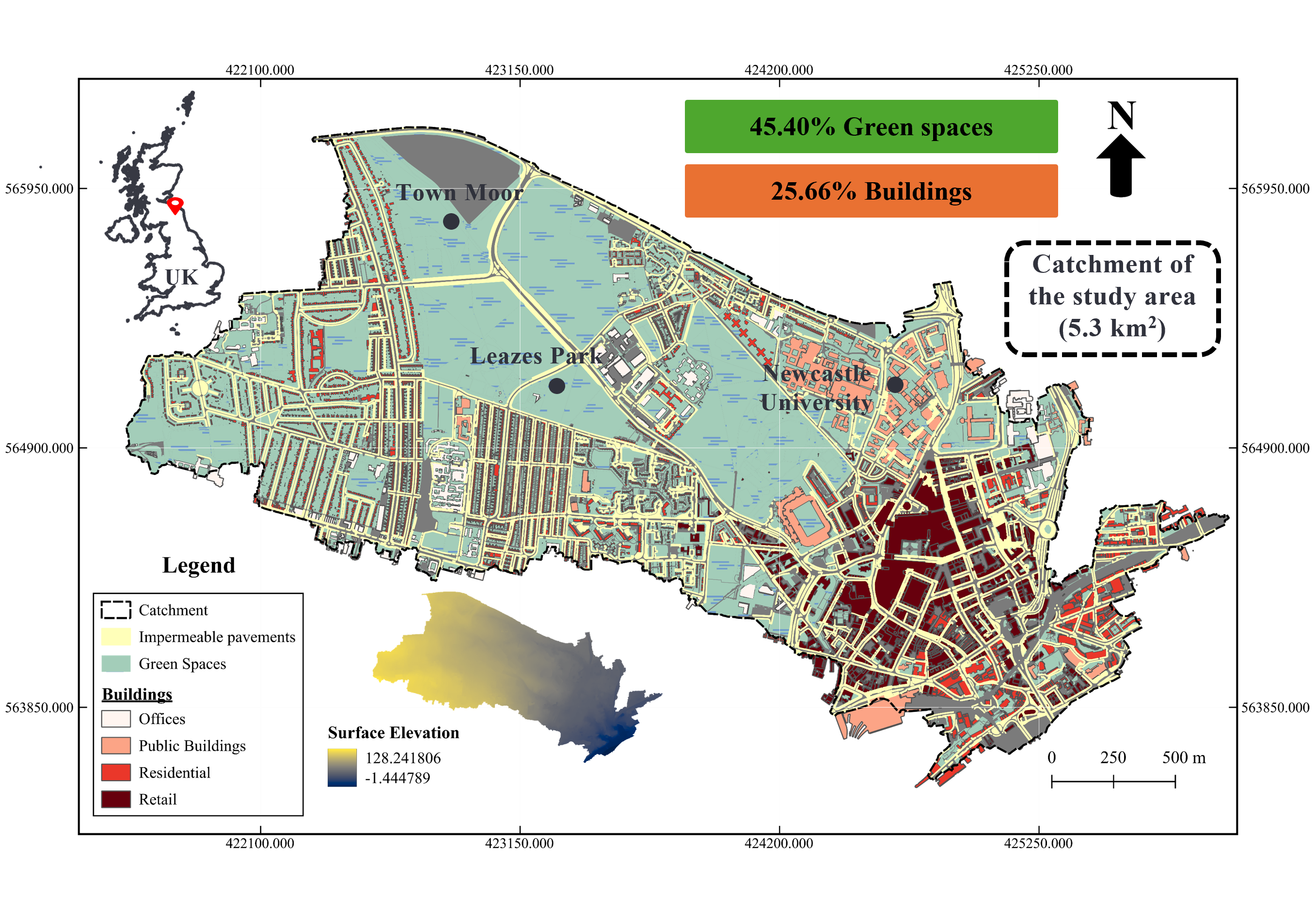}
			\caption{Overview of the study area.}
			\label{Fig: Study Area}
		\end{figure} 

		The catchment area is simulated using CityCAT to assess storm events with return periods of 1 in 30 years (for designs including permeable pavements) and 1 in 50-years (for designs including detention basins), each with a 60-min duration. Figure \ref{Fig: Study Area} presents the catchment area, with buildings categorised by use, land cover, and the proportion of green space. A Digital Terrain Model (DTM) with a resolution of four metres (each grid square covering 16m$^2$) is employed, derived from LiDAR (2022). The computational grid comprised 249,865 cells, encompassing a total area of 5.30 km$^2$. The computational grid is generated using the ‘Building Hole’ approach, with friction coefficients set at 0.02 for impermeable surfaces and 0.035 for permeable ones, consistent with typical values for urban environments \cite{Chow1988}. For the exposure analysis tool (Section \ref{Section: Exposure Analysis}), the buffer zone extent is set equal to a single cell (a 4m buffer zone).
		
		\subsection{Test Case I}
		The modelling setup for Test Case I delineates the study area into twelve arbitrary zones within which the installation of permeable paving, represented as \textit{zonal features}, is considered. The genotype used is therefore of length twelve, with each of the permeable paving zones being modelled as \textit{zonal features}. Through exhaustive evaluation of all 4,096 feasible objective vectors, a test environment is produced for which the exact Pareto set is known (Figure \ref{Fig: 12 Zone Feasible Objective Space}). 
		
		Within CityCAT, the permeable paving is modelled by converting impermeable surfaces into green spaces and modifying their infiltration properties correspondingly. Within the model, the infiltration process is modelled as a function of the soils hydraulic conductivity, porosity and suction head using the Green-Ampt method \cite{Glenis2018}. Impermeable areas are assigned the following properties: hydraulic conductivity = 1.09cm/hr, wetting front suction head = 11.01cm, effective porosity = 0.412, effective saturation = 0.30. For permeable paving, the properties are modified to be: hydraulic conductivity = 4.421cm/hr, wetting front suction head = 19.123cm, effective porosity = 0.345, effective saturation = 0.30.
		
		To estimate the whole-life cost of the proposed permeable paving, the financial cost of the intervention is estimated based on Present Value (PV) discounting and all sourced costs have been adjusted to present day costs accounting for inflation. The design lifespan of permeable paving is assumed to be $T = 50$ years. The unit Capital Expenditure (CAPEX) is calculated by applying an assumed $12\%$ professional fee rate to a base installation cost of £90m$^{-2}$ \cite{CIRIA2007, Stovin2007, GordonWalker2007}. The Operational Expenditure (OPEX) over the lifespan includes annual routine maintenance, a major refurbishment, and end of life decommissioning. Annual maintenance is calculated volumetrically at £1.50m$^{-3}$ \cite{Keating2015, HRWallingford2004} based on an assumed attenuation volume of 0.15m$^3$m$^{-2}$ \cite{CIRIA2015}. Major maintenance (e.g. removal of paviours, jet washing, removal and washing of 5 mm aggregate, removal and replacement of geotextile, replacement of blocks and aggregate and disposal of waste) incurs a cost of £10m$^{-2}$ at year 25 \cite{Pratt2002}, while decommissioning at year 50 is estimated at $38.5\%$ of the base installation cost \cite{Keating2015}. 
		
		To determine the total unit Whole-Life Cost ($C_{\mathrm{WLC}}$) in present terms, future operational costs are discounted at a standard rate of $r=3.5\%$ (in accordance with the HM Treasury Green Book guidelines \cite{HMTreasury2022}) and summed with the CAPEX \cite{HRWallingford2004}:
		\begin{equation}\label{Eq: Whole Life Cost}
			C_{\mathrm{WLC}} = C_{\mathrm{CAPEX}} + \sum_{t=1}^{T} \frac{C_{\mathrm{OPEX}, t}}{(1+r)^t}
		\end{equation}
		where $C_{\mathrm{OPEX}, t}$ represents the sum of all operational and maintenance costs occurring in year $t$. The total cost for a specific objective vector is then obtained by multiplying $C_{\mathrm{WLC}}$ by the active permeable paving area dictated by the evaluated genotype.
		
		\begin{figure}[hbt!]
			\centering
			\includegraphics[width=0.65\linewidth]{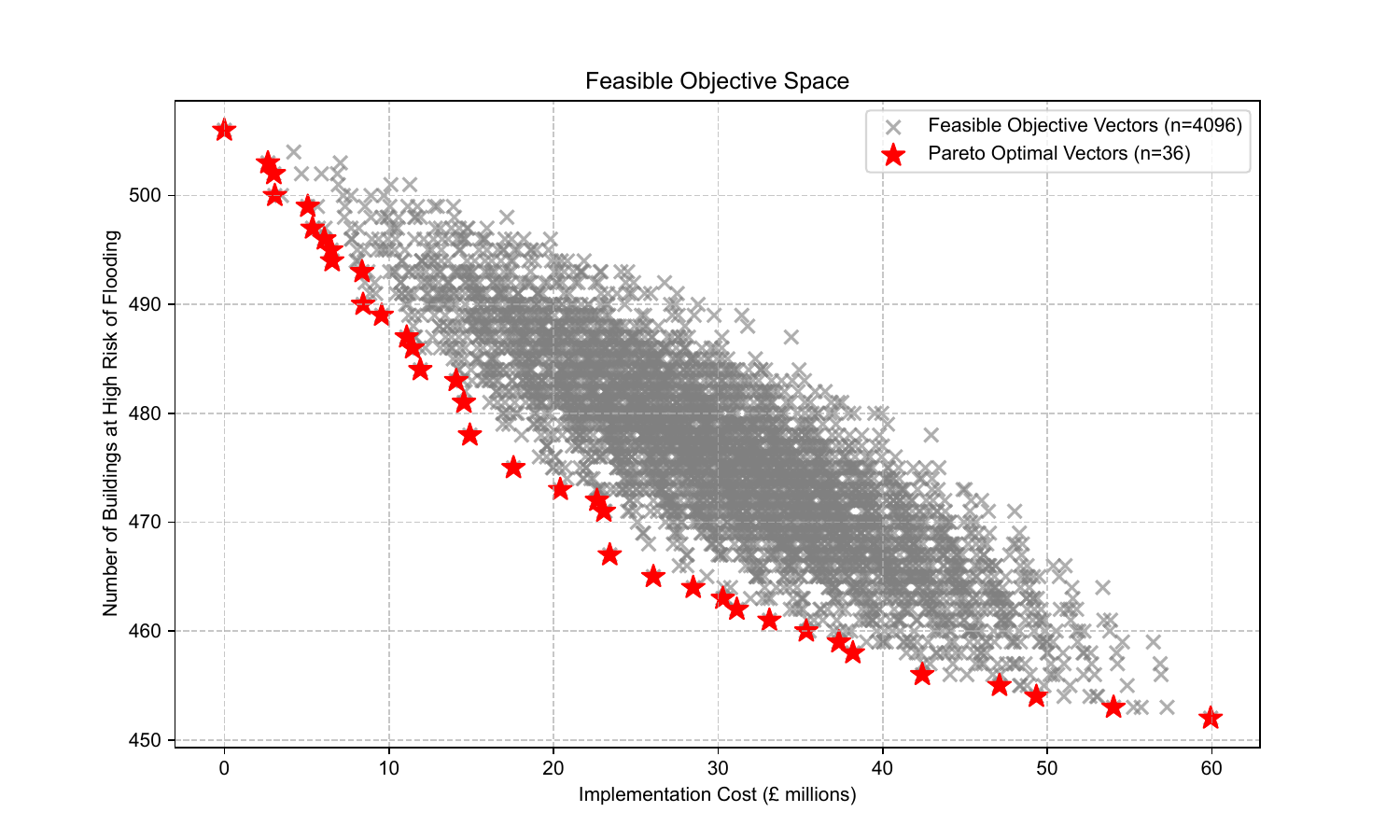}
			\caption{A plot showing all 4,096 feasible objective vectors for the twelve zonal feature test scenario. Pareto optimal vectors are highlighted in red with dominated vectors shown in grey.}
			\label{Fig: 12 Zone Feasible Objective Space}
		\end{figure}
		
		\subsubsection{Results}
		Figures \ref{Fig: Individual Convergence} and \ref{Fig: Combined Convergence} show the hyper-area ratio, as defined by equation (\ref{Eq: Hyper-volume ratio}), for SPEA-2, NSGA-II and the proposed $\epsilon-MOEA$ for Test Case I. The algorithms were tested with various population sizes, $\bar{N} \in \{ L, 2L, |\mathcal{P}|, 2|\mathcal{P}| \}$, where $L=12$ is the length of the encoding and $|\mathcal{P}|=36$ is the cardinality of the Pareto set. For NSGA-II and SPEA-2, $\bar{N}$ corresponds to the fixed archive size, with $\bar{N} = N$, $|\bar{P}|  = N$ and $|P|=N$ (see Algorithms \ref{Alg: SPEA-2} and \ref{Alg: NSGA-II}). For the proposed $\epsilon-$MOEA, $\bar{N}$ represents the maximum archive size, as per equation (\ref{Eq: Bound}). Although the inherent structural differences between the algorithms preclude a perfectly equivalent comparison, this setup provides a robust practical baseline for evaluating performance. Comparisons are made based on the average hyper-volume ratio per unique fitness evaluation (referred to as unique simulations) across 100 independent runs for each algorithm.
		
		\begin{figure}[hbt!]
			\includegraphics[width=\linewidth]{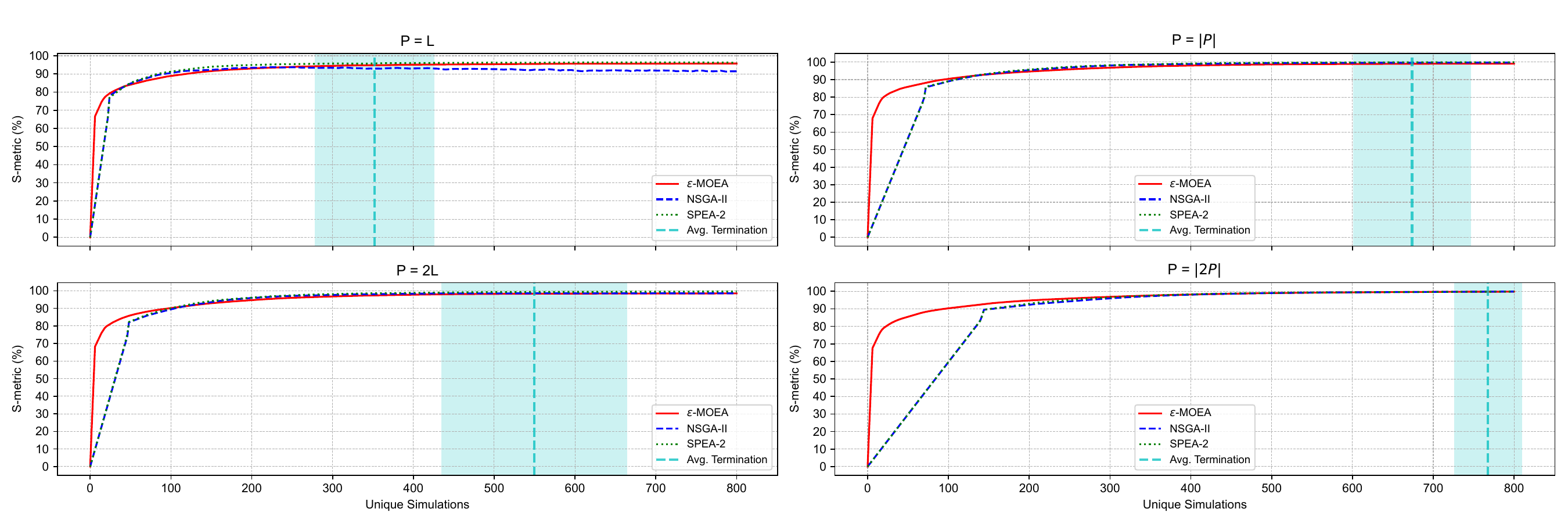}
			\caption{Comparison between the hyper-volume ratio (S-metric (\%)) versus the number of fitness evaluations for the three studied MOEAs with different maximum archive sizes ($L=12$ and $|\mathcal{P}|=36$). The average time of self-termination for the $\epsilon-$MOEA is illustrated via the dashed vertical blue line, with the shaded area showing the variance (one standard deviation).}
			\label{Fig: Individual Convergence}
		\end{figure}
		
		\begin{figure}[hbt!]
			\includegraphics[width=\linewidth]{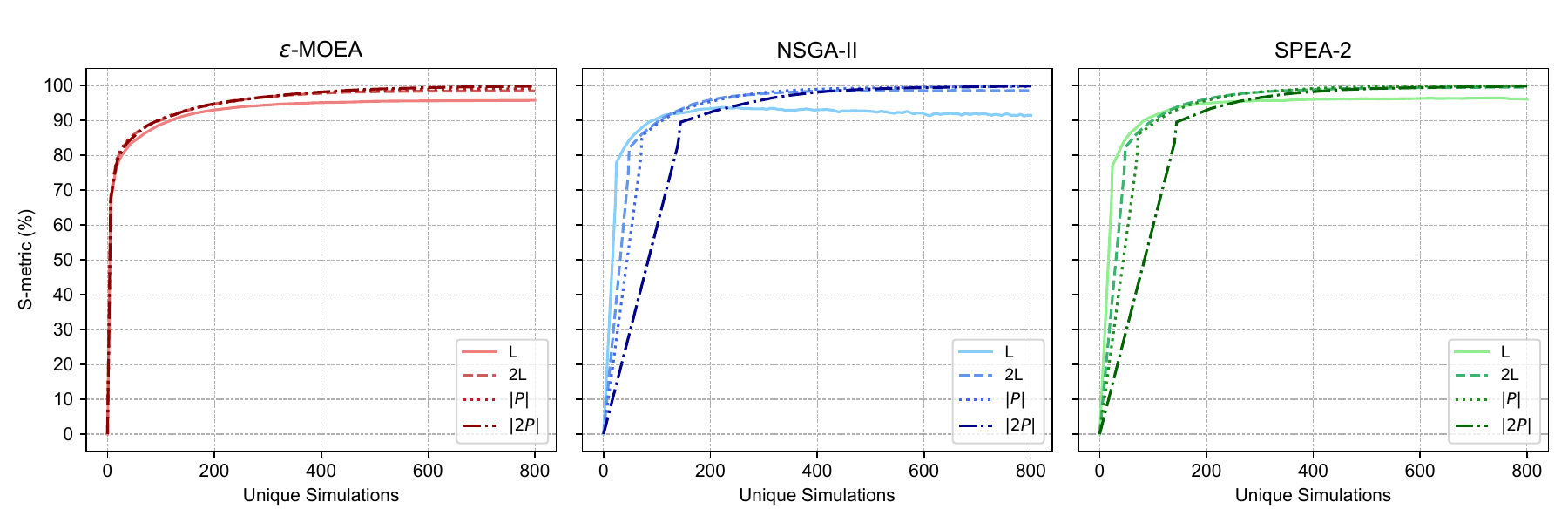}
			\caption{Performance of the algorithms for varying maximum archive sizes ($L=12$ and $|\mathcal{P}|=36$), shown through plots of the hyper-volume ratio against the number of fitness evaluations.}
			\label{Fig: Combined Convergence}
		\end{figure}
		
		The results demonstrate the capability of all three MOEAs in approximating the Pareto set in an efficient manner, requiring far fewer fitness evaluations than an exhaustive or random search. On a per unique simulation basis, the quality of the approximation of the Pareto set is comparable across all three algorithms, particularly as the number of fitness evaluations increases. As expected, when the archive population is smaller than the cardinality of the Pareto set, a hyper-area ratio of $1$ is unattainable. 
		
		When considering the effectiveness of the algorithms on a per unique simulation basis, it is also important to examine, within the context of expensive objective functions, the average time taken to reach an appropriate quality threshold and the corresponding first opportunity to terminate the algorithm. As shown in Figure \ref{Fig: Combined Convergence} and \ref{Fig: Individual Convergence}, there is a delay of one generation prior to the first convergence results, which is equal to $\approx 2\bar{N}$ unique simulations for the second generation algorithms and only $\bar{N} + N \approx 7$ unique simulations for the proposed $\epsilon-$MOEA. Therefore, when the number of unique simulations, $N_f$, is in the range $\bar{N} + N \lesssim N_f \lesssim 2\bar{N}$ the $\epsilon-$MOEA offers an incomparable quality of approximation. 
		
		Furthermore, opportunities to terminate the second generation algorithms, following evaluation of the first generation, occur at intervals of $\approx \bar{N}$ unique fitness evaluations. Whereas, for the proposed $\epsilon-$MOEA opportunities to terminate the algorithm occur at intervals of $\approx 5 - 4|\mathcal{P}^*|$ fitness evaluations, with the maximum value of $|\mathcal{P}^*|$ determined by the selection of $\epsilon$. Hence, for MOOPs involving expensive objective functions, the proposed $\epsilon-$MOEA offers significant advantages in terms of maximising the quality of the approximation of the Pareto set per fitness evaluation and minimising the total number of fitness evaluations required. This is particularly the case for MOOPs involving large search spaces, whereby $L$ is large and $\bar{N}=L$ is often employed as a heuristic baseline population size for the second generation MOEAs. 
		
		Figure \ref{Fig: Individual Convergence} also demonstrates the effectiveness of the hyper-volume based self-termination criterion. In each case the algorithm self-terminated once convergence slowed, with smaller $\bar{N}$ and larger $\epsilon$ resulting in earlier termination since, as shown in Figure \ref{Fig: Combined Convergence}, the performance of the algorithm is less affected by the choice of $\bar{N}$ due to the adaptive population sizing. 
		
		\subsection{Test Case II}
				
		\begin{figure}[hbt!]\centering
			\begin{subfigure}{.75\linewidth}
				\centering
				\includegraphics[width=\linewidth]{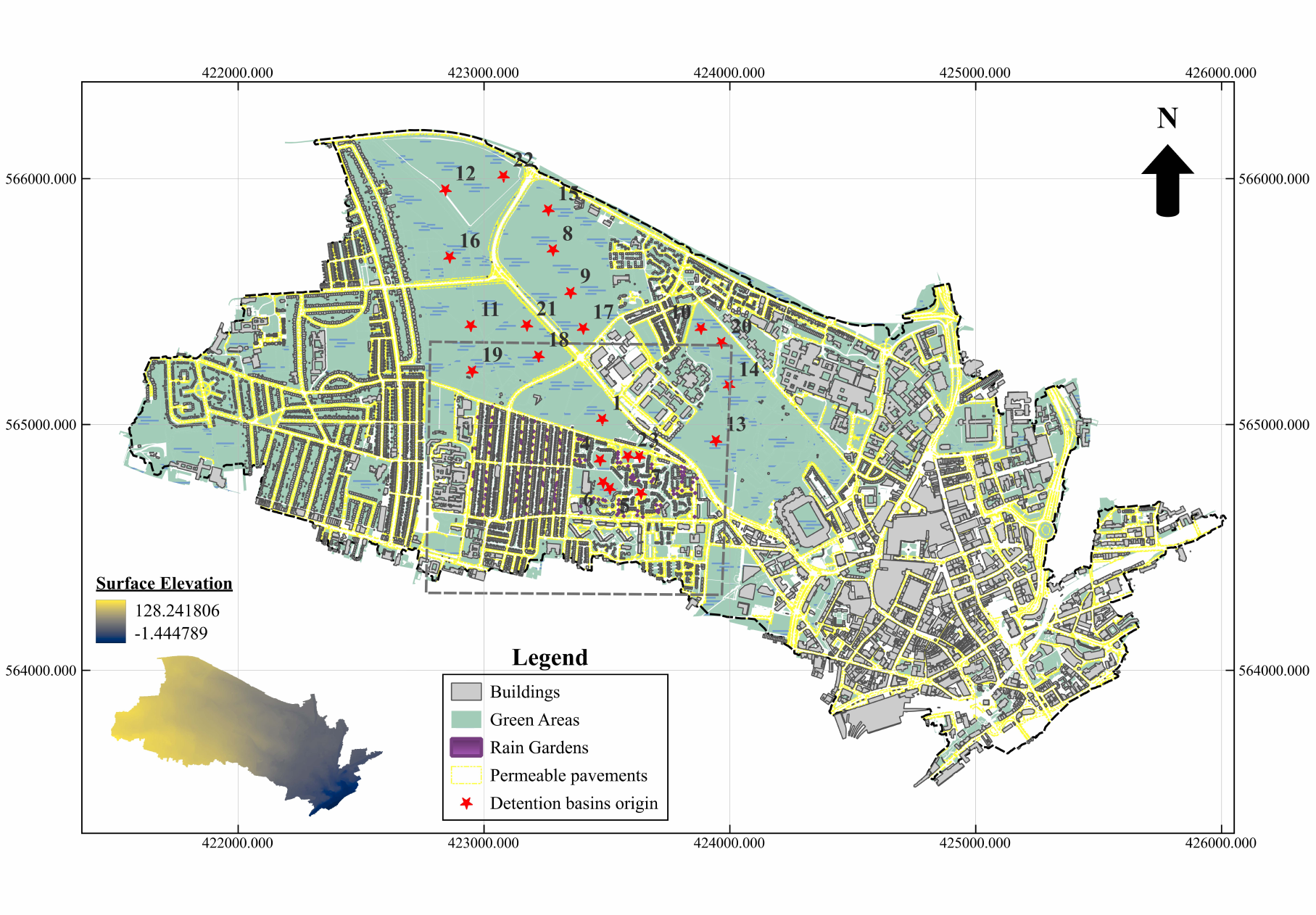}
				\caption{Catchment area and BGI features.}\label{Fig: Domain}
			\end{subfigure}
			\begin{subfigure}{.6\linewidth}
				\centering
				\includegraphics[width=\textwidth]{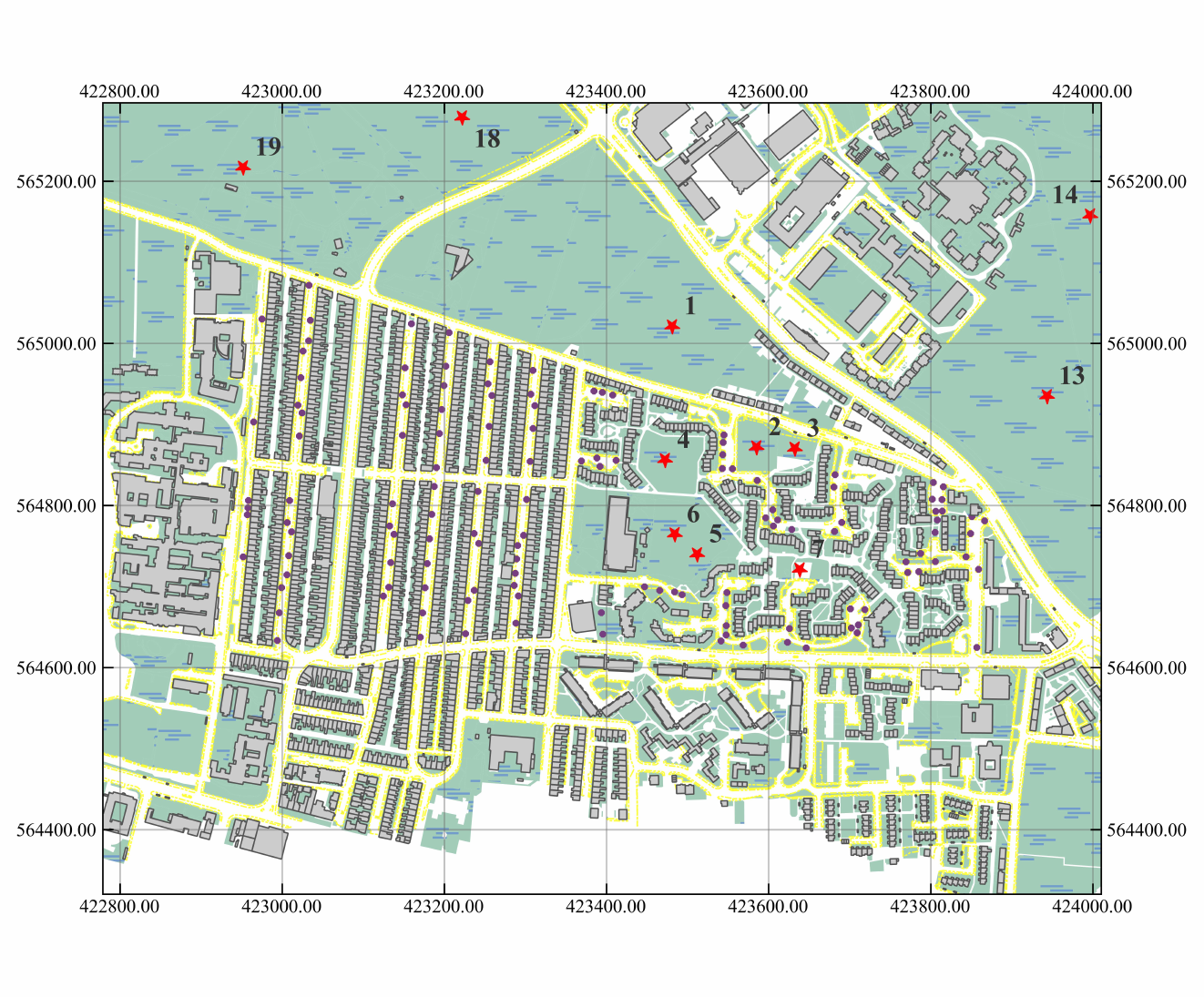}
				\caption{Rain garden locations (purple circles). Extent of the plot is illustrated above via the dashed rectangle.}\label{Fig: Zoom}
			\end{subfigure}
			\caption{The spatial domain for Test Case II.}
			\label{Fig: Realistic Case Study Setup}
		\end{figure}
		
		Although Test Case I provides an opportunity for a rigorous validation of the capabilities of the proposed $\epsilon-$MOEA, the search space is, by necessity, small and of limited practical relevance. Hence, Test Case II is also presented to demonstrate the real world applicability of the optimisation tool. Particularly as the time to convergence is expected to grow super-linearly with the size of the search space \cite{Laumanns2004, Giel2010, Osuna2020, Doerr2022, Zheng2022, Doerr2023}, the use of a much larger search space enables for a clearer understanding of the performance differences which are mostly indistinguishable for Test Case I. However, by using a large search space, an exhaustive search of the search space is rendered intractable. As a result, the Pareto set is unknown, and inter-algorithm comparisons provide the only meaningful basis for analysis. As such, the hyper-volume indicator (equation \ref{Eq: Hyper-volume indicator}) is used instead of the hyper-volume ratio (equation \ref{Eq: Hyper-volume ratio}) as the performance indicator.
		
		The modelling setup includes the delineation of the catchment into 124 zones in which permeable paving is considered for installation, each modelled as a \textit{zonal feature}, the installation of 131 10m$^2$ rain gardens, each modelled as a \textit{zonal feature}, and the construction of 22 detention basins, each modelled as a \textit{local feature}. The proposed locations for the rain gardens and detention basins can be seen in Figure \ref{Fig: Realistic Case Study Setup}. For each detention basin, an origin is specified and the following characteristics are optimised: x-shift, y-shift, radius (surface area) and depth. 
		
		The algorithm generates each basin by translating the origin by the $x$-shift and $y$-shift and constructing a circle of radius $r$. The surrounding elevation is then estimated by taking the 10th percentile elevation of the perimeter cells from the DEM. The elevation of the cells contained within the perimeter of the constructed circle are then modified to be equal to the 10th percentile perimeter elevation minus the specified depth. An exclusion zone, generated using a 5m buffer around the buildings and pavements prevents the modification of cells which are too close to existing infrastructure. The selection of these characteristics enables the algorithm to optimise the position and sizing of each basin. 
		
		The rain gardens are modelled using the following parameters: hydraulic conductivity = 10.8cm/hr, wetting front suction head = 4.55cm, effective porosity = 0.34, effective saturation = 0.20. The total length of the genotype for each candidate solution is 507 bits, with the first 124 bits corresponding to the permeable paving zones, the next 131 bits corresponding to the rain gardens and the final 252 bits corresponding to the detention basins.
		
		Full details for the encoding of each detention basin is provided in Table \ref{Table: Pond Data}. It is worth noting that the use of $0$m as the minimum depth for each basin encodes the exclusion of a basin, in compliance with design guideline 1 (Section \ref{Section: Guidelines}). This does however introduce some phenotypic redundancy, whereby the synonymous genotypes form a connected region in the Hamming space. Precisely, this means that multiple unique genetic representations correspond to a storage volume of $0$m$^3$. The level of redundancy is however minimised by ensuring that only the depth, and not the surface area, has a minimum value of zero.
		
		In accordance with the work of Rothlauf and Goldberg \cite{Rothlauf2003}, the over-representation of the exclusion of the detention basins in the genetic representation is expected to accelerate the evolutionary search where the exclusion of basins is optimal. This can be a useful tool to take advantage of if a degree of redundancy is incorporated into the system, such that it is expected that the inclusion of only a subset of the potential detention basins is optimal. Hence, the biasing of the search through synonymous redundancy is considered to be appropriate for this scenario, where an infeasible number of detention basins are optimised. Where this is not the case, it is preferable to encode the volume directly, varying the surface area and depth of the basin based on a pre-defined depth-surface area relationship for a given storage volume such that each phenotype is unique. As always, selecting an appropriate genetic representation for a MOOP requires careful consideration of the specific problem context and objectives.
		
		\begin{table}[ht]
			\centering
			\resizebox{\textwidth}{!}{%
				\footnotesize
				\renewcommand{\arraystretch}{1.2}
				\setlength{\tabcolsep}{3pt}
				\begin{tabular}{|c|c|c|c|c|c|c|c|c|c|c|c|c|c|c|}
					\hline
					\textbf{Basin} & \multicolumn{2}{c|}{\textbf{Centre Point (m)}} & \multicolumn{3}{c|}{\textbf{Depth (m)}} & \multicolumn{3}{c|}{\textbf{Surface Area (m$^2$)}} & \multicolumn{3}{c|}{\textbf{x-Shift (m)}} & \multicolumn{3}{c|}{\textbf{y-Shift (m)}} \\ \cline{2-15}
					& \textbf{E} & \textbf{N} & \textbf{Min} & \textbf{Max} & $\mathbf{\epsilon}_d$ & \textbf{Min} & \textbf{Max} & $\mathbf{\epsilon}_a$ & \textbf{Min} & \textbf{Max} & $\mathbf{\epsilon}_x$ & \textbf{Min} & \textbf{Max} & $\mathbf{\epsilon}_y$ \\ \hline
					1 & 423481 & 565021 & 0 & 1.5 & 0.5 & 314 & 18051 & 1182 & -5.00 & 5.00 & 3.33 & -10.00 & 10.00 & 6.67 \\
					2 & 423585 & 564872 & 0 & 1.5 & 0.5 & 314 & 1948 & 545 & -5.00 & 5.00 & 3.33 & -10.00 & 10.00 & 6.67 \\
					3 & 423632 & 564870 & 0 & 1.5 & 0.5 & 177 & 707 & 177 & -10.00 & 10.00 & 6.67 & -5.00 & 5.00 & 3.33 \\
					4 & 423472 & 564856 & 0 & 1.5 & 0.5 & 314 & 1576 & 421 & -10.00 & 10.00 & 6.67 & -10.00 & 10.00 & 6.67 \\
					5 & 423512 & 564740 & 0 & 1.5 & 0.5 & 314 & 2341 & 676 & -10.00 & 10.00 & 6.67 & -10.00 & 10.00 & 6.67 \\
					6 & 423484 & 564765 & 0 & 1.5 & 0.5 & 314 & 2341 & 676 & -10.00 & 10.00 & 6.67 & -10.00 & 10.00 & 6.67 \\
					7 & 423638 & 564721 & 0 & 1.5 & 0.5 & 177 & 531 & 118 & -5.00 & 5.00 & 3.33 & -5.00 & 5.00 & 3.33 \\
					8 & 423280 & 565711 & 0 & 1.5 & 0.5 & 314 & 6121 & 830 & -163.86 & 156.75 & 10.34 & -30.89 & 29.42 & 4.02 \\
					9 & 423352 & 565537 & 0 & 1.5 & 0.5 & 314 & 7205 & 984 & -102.74 & 103.93 & 6.67 & -31.92 & 33.01 & 4.33 \\
					10 & 423883 & 565391 & 0 & 1.5 & 0.5 & 314 & 1470 & 385 & -16.62 & 14.42 & 4.43 & -70.73 & 73.60 & 4.65 \\
					11 & 422946 & 565403 & 0 & 1.5 & 0.5 & 314 & 18909 & 1240 & -96.28 & 103.18 & 6.43 & -51.71 & 53.95 & 7.04 \\
					12 & 422843 & 565955 & 0 & 1.5 & 0.5 & 314 & 31858 & 2103 & -67.64 & 67.13 & 4.35 & -85.16 & 79.13 & 5.30 \\ 
					13 & 423943 & 564935 & 0 & 1.5 & 0.5 & 314 & 5690 & 768 & -73.84 & 73.94 & 4.77 & -28.37 & 28.78 & 3.81 \\ 
					14 & 423997 & 565158 & 0 & 1.5 & 0.5 & 314 & 6052 & 820 & -29.26 & 29.44 & 3.91 & -95.75 & 96.79 & 6.21 \\ 
					15 & 423262 & 565874 & 0 & 1.5 & 0.5 & 314 & 6404 & 870 & -32.22 & 32.40 & 4.31 & -30.29 & 30.11 & 4.03 \\ 
					16 & 422860 & 565682 & 0 & 1.5 & 0.5 & 314 & 6743 & 918 & -105.03 & 102.65 & 6.70 & -30.88 & 34.54 & 4.36 \\ 
					17 & 423403 & 565392 & 0 & 1.5 & 0.5 & 314 & 3587 & 1091 & -28.45 & 28.85 & 3.82 & -24.23 & 22.53 & 6.68 \\ 
					18 & 423222 & 565279 & 0 & 1.5 & 0.5 & 314 & 6651 & 905 & -42.76 & 43.58 & 5.76 & -34.85 & 30.68 & 4.37 \\ 
					19 & 422952 & 565217 & 0 & 1.5 & 0.5 & 314 & 3056 & 914 & -157.26 & 147.89 & 9.84 & -26.69 & 20.79 & 6.78 \\ 
					20 & 423966 & 565334 & 0 & 1.5 & 0.5 & 314 & 1012 & 233 & -26.51 & 26.51 & 3.53 & -11.97 & 12.59 & 8.19 \\ 
					21 & 423174 & 565405 & 0 & 1.5 & 0.5 & 314 & 2309 & 665 & -18.07 & 18.60 & 5.24 & -18.86 & 19.39 & 5.46 \\ 
					22 & 423079 & 566011 & 0 & 1.5 & 0.5 & 314 & 4329 & 574 & -28.51 & 29.64 & 3.88 & -24.75 & 25.49 & 7.18 \\ \hline
				\end{tabular}
			}
			\caption{Pond Data Summary.}
			\label{Table: Pond Data}
		\end{table}
		
		The Whole-Life Cost for each basin is calculated using equation (\ref{Eq: Whole Life Cost}). The design lifespan is assumed to be $T = 50$ years \cite{CIRIA2015, Keating2015}. The unit CAPEX is calculated by applying a $12\%$ professional fee rate to a base installation cost of £33.6m$^{-3}$ \cite{Keating2015} excavated volume. The OPEX over the lifespan is discounted annually at a rate of $3.5\%$ in accordance with the Green Book \cite{HMTreasury2022}. The OPEX includes:
		\begin{enumerate}
			\item Annual Maintenance: Regular upkeep calculated at £1.14m$^{-3}$ \cite{Keating2015, Ellis2003}.
			\item Minor Maintenance: Periodic maintenance costing £840 per basin \cite{Keating2015, Ellis2003}, applied every three years (excluding years when major maintenance takes place).
			\item Major Maintenance: Significant upkeep costing £$8,400$ per basin \cite{Keating2015, Ellis2003}, applied every ten years.
			\item Decommissioning: Applied in the final year of the lifespan, estimated at $38.5\%$ of the initial base construction cost \cite{Keating2015}.
		\end{enumerate}
		
		For the rain gardens, the design lifespan is assumed to be $T = 50$ years \cite{CIRIA2015, Keating2015}. The unit CAPEX is calculated by applying a $12\%$ professional fee rate to a base installation cost of £250m$^2$. The OPEX over the lifespan is also discounted annually at $3.5\%$ and includes:
		\begin{enumerate}
			\item Annual Maintenance: Regular gardening and upkeep at a rate of £2m$^{-2}$ \cite{Keating2015}.
			\item Major Maintenance: Periodic replanting costing £50m$^{-2}$ \cite{Keating2015}, applied every ten years.
			\item Decommissioning: Applied in the final year of the lifespan, estimated at $38.5\%$ of the initial base construction cost \cite{Keating2015}.
		\end{enumerate}
		
		For this test case, the maximum front size for the $\epsilon-$MOEA was selected as $\bar{N}=16$, based on an assumed manageable number of solutions for a decision-maker, and the algorithm was run until self-termination. For the benchmark algorithms, the population size was set equal to the representation length, $\bar{N}=507$, and the algorithms were run for $\approx$3000 simulations.
		
		\subsubsection{Results}
		
		\begin{figure}[hbt!]
			\centering
			\includegraphics[width=0.75\linewidth]{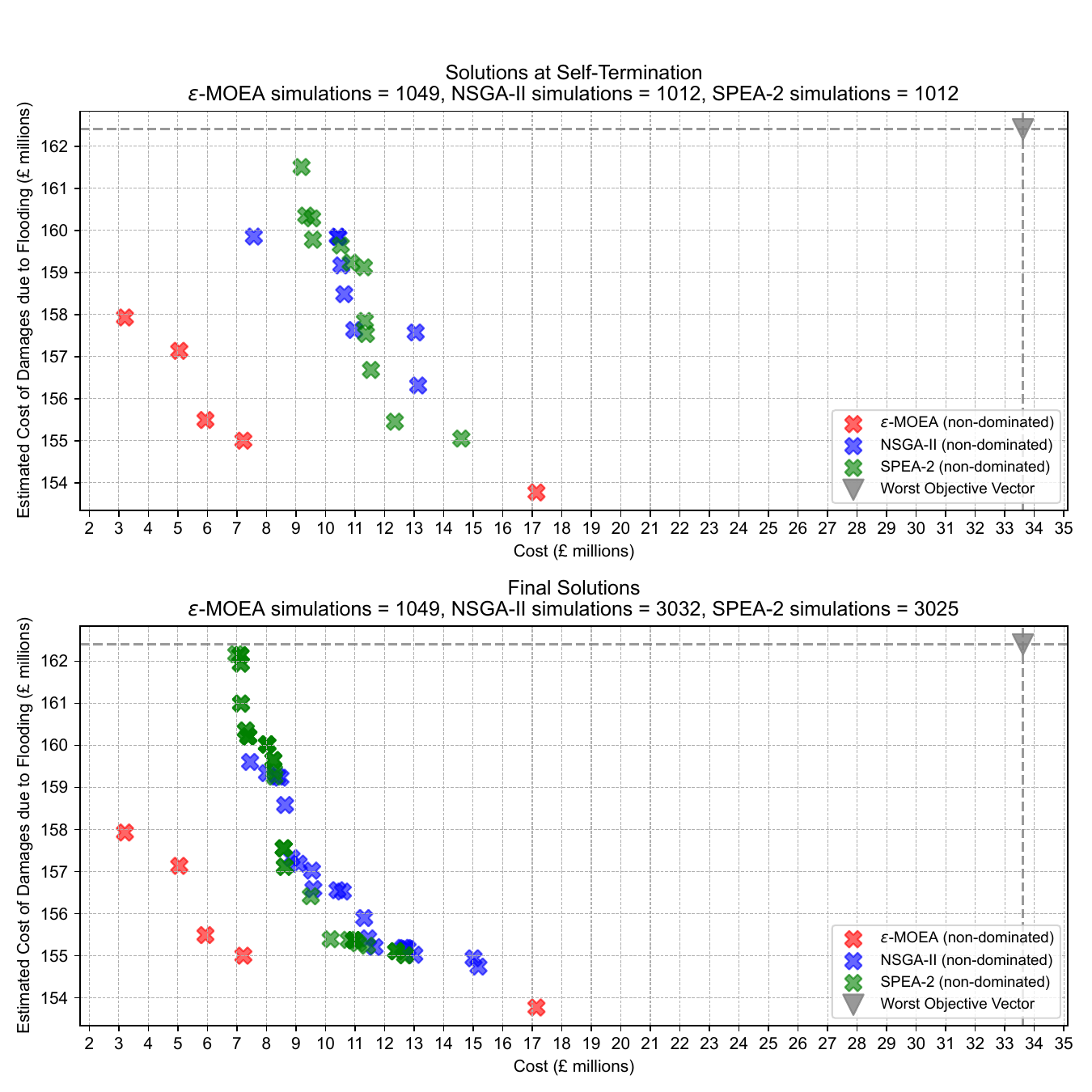}
			\caption{Comparison between the Pareto front approximations for the three MOEAs after $\approx$1000 simulations (top) and then after the benchmark algorithms have performed an additional $\approx$2000 simulations (bottom). The worst objective vector is used as the fixed reference point for the calculation of the performance metric.}
			\label{Fig: Front Comparison}
		\end{figure}
		
		\begin{figure}[hbt!]
			\centering
			\includegraphics[width=0.75\linewidth]{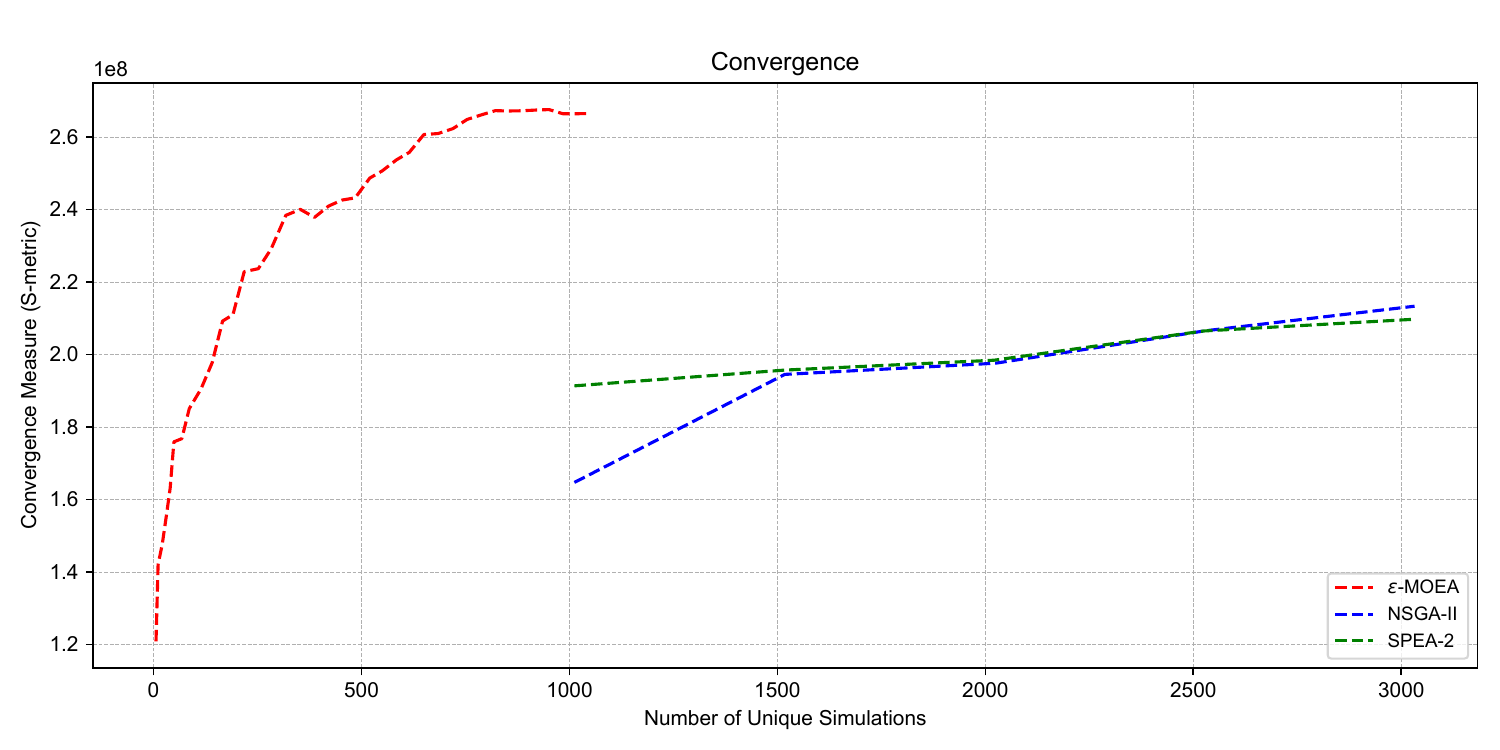}
			\caption{S-metric per unique simulation for the three MOEAs.}
			\label{Fig: Convergence Comparison}
		\end{figure}
		
		From Figures \ref{Fig: Front Comparison} and \ref{Fig: Convergence Comparison} the relative superiority of the performance of the proposed $\epsilon-$MOEA with respect to the second generation benchmark MOEAs is clear. After an equivalent number of simulations ($\approx 1000$) all of the $\epsilon-$MOEA solutions clearly dominate the benchmark solutions and there is a large difference in the performance as measured by the hyper-volume indicator. Notably, even after the benchmark algorithms perform a further $\approx2000$ simulations, the Pareto front approximation remains notably inferior which is also reflected in the difference in the hyper-volume indicator. Furthermore, it is clear from Figure \ref{Fig: Convergence Comparison} that the self-termination criteria has successfully terminated the algorithm as the convergence rate has begun to plateau. This demonstrates the suitability of the proposed $\epsilon-$MOEA for the purpose of automating the optioneering of BGI in urban environments in a computationally practical manner. The flood risk maps for each of the five optimal solutions discovered by the $\epsilon-$MOEA are presented in the Appendices.
		
		\section{Conclusion}
		This study presents the development and validation of a novel multi-objective optimisation tool designed to support the automated placement and design of BGI in urban environments. By coupling a state-of-the-art hydrodynamic model (CityCAT) with a bespoke evolutionary algorithm ($\epsilon-$MOEA), the proposed tool overcomes the limitations of traditional design approaches. Unlike other optimisation approaches which employ simplified hydrodynamic models, this framework utilises fully dynamic modelling to accurately evaluate flood risk at the property level, enabling the rigorous optimisation of complex intervention strategies.
		
		The validation results demonstrate that the $\epsilon$-MOEA significantly outperforms standard second-generation algorithms (NSGA-II and SPEA-2). Specifically, the proposed tool efficiently explores realistic search spaces, providing high-quality approximations of the Pareto-optimal front with a fraction of the computational effort required by the benchmark algorithms. This efficiency renders the tool computationally practical for real-world applications where hydrodynamic simulations are computationally burdensome, especially for large catchments and intense rainfall events.
		
		While the case studies only considered the implementation of permeable paving, rain gardens and detention basins within Newcastle upon Tyne, the framework is inherently flexible and capable of representing a diverse range of BGI features across complex urban environments. A dockerised version of the tool has been deployed on the Data and Analytics Facility for National Infrastructure (DAFNI) \cite{Matthews2023}, facilitating its use by practitioners and researchers to enhance urban flood resilience. The deployment of the tool on the DAFNI platform is highly beneficial as it enables decision-makers who otherwise lack access to high end computational resources to access a platform independent high-throughput computational resource and data repository, without requiring significant and unrealistic investment in computational hardware.
		
		Future work will focus on expanding the library of BGI features and incorporating true multi-objective optimisation to capture the wider co-benefits of BGI, such as amenity value and biodiversity. Although the \textit{curse of dimensionality} presents a significant challenge when exploring higher-dimensional search spaces, the efficiency gains demonstrated by this tool provide a strong foundation for these advancements. However, further significant improvements in algorithmic efficiency, as well as the development of robust tools to quantify the wider co-benefits, is likely required to achieve this goal. Ultimately, this research offers a robust, automated pathway for decision-makers to identify cost-effective, resilient nature-based solutions in an increasingly uncertain climate.
		
		\section*{Declaration of Competing Interest}
		The authors declare that they have no known competing financial interests or personal relationships that could have appeared to influence the work reported in this paper.
		
		\section*{Acknowledgements}
		This work was supported by the DAFNI Centre of Excellence for Resilient Infrastructure Analysis within the Building a Secure and Resilient World programme (grant ST/Y003837/1), and by the European Union’s Horizon 2020 research and innovation programme under grant agreement No 101081555 (IMPETUS4CHANGE).
		
		\appendix
		\section{Appendix 1: Flood Maps}
		\begin{figure}[hbt!]
			\centering
			\includegraphics[width=0.975\linewidth]{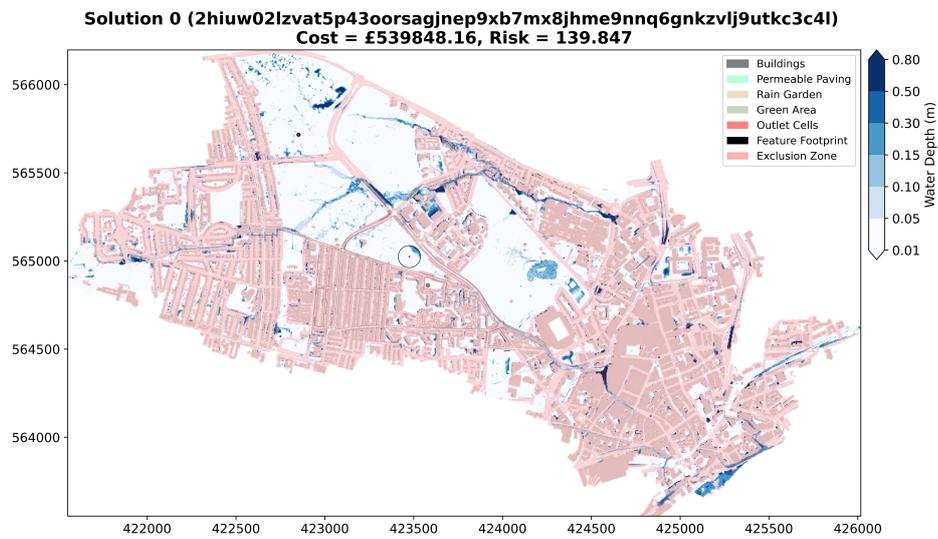}
			\caption{Flood map for a solution discovered by the $\epsilon-$MOEA algorithm with a cost of £0.5 million and an expected cost of damages due to flooding of £140 million. See Figure \ref{Fig: Front Comparison} for a visualisation of the solution within the objective space.}
			\label{Fig: Appendix 1}
		\end{figure}
		\begin{figure}[hbt!]
			\centering
			\includegraphics[width=0.975\linewidth]{Appendix_2.jpg}
			\caption{Flood map for a solution discovered by the $\epsilon-$MOEA algorithm with a cost of £2.5 million and an expected cost of damages due to flooding of £138 million. See Figure \ref{Fig: Front Comparison} for a visualisation of the solution within the objective space.}
			\label{Fig: Appendix 2}
		\end{figure}
		\begin{figure}[hbt!]
			\centering
			\includegraphics[width=0.975\linewidth]{Appendix_3.jpg}
			\caption{Flood map for the solution discovered by the $\epsilon-$MOEA algorithm with a cost of £0.3 million and an expected cost of damages due to flooding of £142 million. See Figure \ref{Fig: Front Comparison} for a visualisation of the solution within the objective space.}
			\label{Fig: Appendix 3}
		\end{figure}
		\begin{figure}[hbt!]
			\centering
			\includegraphics[width=0.975\linewidth]{Appendix_4.jpg}
			\caption{Flood map for a solution discovered by the $\epsilon-$MOEA algorithm with a cost of £1.3 million and an expected cost of damages due to flooding of £139 million. See Figure \ref{Fig: Front Comparison} for a visualisation of the solution within the objective space.}
			\label{Fig: Appendix 4}
		\end{figure}
		\begin{figure}[hbt!]
			\centering
			\includegraphics[width=0.975\linewidth]{Appendix_5.jpg}
			\caption{Flood map for a solution discovered by the $\epsilon-$MOEA algorithm with a cost of £1.5 million and an expected cost of damages due to flooding of £138 million. See Figure \ref{Fig: Front Comparison} for a visualisation of the solution within the objective space.}
			\label{Fig: Appendix 5}
		\end{figure}
	}
	\clearpage
	\bibliography{References}
\end{document}